\documentclass[preprints,article,accept,moreauthors,pdftex]{Definitions/mdpi}

\firstpage{1} 
\pubvolume{xx}
\issuenum{1}
\articlenumber{5}
\pubyear{2020}
\copyrightyear{2020}
\history{Received: date; Accepted: date; Published: date}
\pdfoutput=1

\usepackage{float}
\usepackage{subfig}
\usepackage{graphicx}

\definecolor{BLACK}{rgb}{0,0,0}
\definecolor{WHITE}{rgb}{1,1,1}
\definecolor{RED}{rgb}{1,0.0,0.0}
\definecolor{BLUE2}{rgb}{0.0,0.3,0.6}
\definecolor{DARKSILVER}{rgb}{0.5,0.5,0.5}
\definecolor{CYAN}{rgb}{0.0,1.0,1.0}
\definecolor{GREEN2}{rgb}{0.2,1.0,0.0}
\definecolor{YELLOW}{rgb}{1.0,0.88,0.21}
\definecolor{DARKRED}{rgb}{0.5,0.0,0.13}
\definecolor{LIGHTRED}{rgb}{0.8,0.0,0.0}
\definecolor{LIGHTPURPLE}{rgb}{0.75,0.58,0.89}
\definecolor{PURPLE}{rgb}{0.54,0.17,0.89}
\definecolor{CYAN}{rgb}{0.3,0.91,0.87}
\definecolor{BLUEGREEN}{rgb}{0.11,0.67,0.84}
\definecolor{DARKGREEN}{rgb}{0.0,0.5,0.0}
\definecolor{LIGHTGREEN}{rgb}{0.4,1.0,0.0}
\definecolor{YELLOW}{rgb}{1.0,0.88,0.21}
\definecolor{SILVER}{rgb}{0.75,0.75,0.75}
\definecolor{DARKSILVER}{rgb}{0.5,0.5,0.5}
\definecolor{BLUEBROWN}{rgb}{0.52,0.4,0.27}

\usepackage{array} % for '\newcolumntype' and '\extrarowheight' macros
\newcolumntype{C}[1]{>{\centering\arraybackslash}p{#1}} % centered version of 'p' column type
\usepackage{arydshln}

\usepackage[T1]{fontenc}
\usepackage{lmodern}

\floatstyle{plaintop}
\restylefloat{table}
\usepackage{alphalph}

\usepackage{pgfplots} 
\pgfplotsset{compat=newest}
\usepackage{tikz}
\usetikzlibrary{shapes,arrows}
\pgfplotsset{plot coordinates/math parser=false}
\pgfplotsset{scaled y ticks=base 10:-3}
\newlength\figureheight
\newlength\figurewidth 

\Title{Ice Monitoring in Swiss Lakes from Optical Satellites and Webcams using Machine Learning}

\Author{Manu Tom $^\text{1}$, Rajanie Prabha $^\text{2}$, Tianyu Wu $^\text{1}$, Emmanuel Baltsavias $^\text{1}$, Laura Leal-Taixe $^\text{2}$, Konrad Schindler $^\text{1}$}

\AuthorNames{Manu Tom, Rajanie Prabha, Tianyu Wu, Emmanuel Baltsavias, Laura Leal-Taixe and Konrad Schindler}

\address{%
$^\text{1}$ \quad Institute of Geodesy and Photogrammetry, ETH Zurich, 8093 Zurich, Switzerland;\\%~~~~~~~manu.tom@geod.baug.ethz.ch (M.T.); wuti@student.ethz.ch (T.W.);\\ ~~~~~~~emmanuel.baltsavias@geod.baug.ethz.ch (E.B.); konrad.schindler@geod.baug.ethz.ch (K.S.)\\
$^\text{2}$ \quad Dynamic Vision and Learning Group, TU Munich, 85748 Munich, Germany;\\
}

\abstract{Continuous observation of climate indicators, such as trends in lake freezing, is important to understand the dynamics of the local and global climate system. Consequently, lake ice has been included among the Essential Climate Variables (ECVs) of the Global Climate Observing System (GCOS), and there is a need to set up operational monitoring capabilities.
Multi-temporal satellite images and publicly available webcam streams are among the viable data sources to monitor lake ice. In this work we investigate machine learning-based image analysis as a tool to determine the spatio-temporal extent of ice on Swiss Alpine lakes as well as the ice-on and ice-off dates, from both multispectral optical satellite images (VIIRS and MODIS) and RGB webcam images. We model lake ice monitoring as a pixel-wise semantic segmentation problem, i.e., each pixel on the lake surface is classified to obtain a spatially explicit map of ice cover. We show experimentally that the proposed system produces consistently good results when tested on data from multiple winters and lakes. Our satellite-based method obtains mean Intersection-over-Union (mIoU) scores $>$93\%, for both sensors. It also generalises well across lakes and winters with mIoU scores $>$78\% and $>$80\% respectively. On average, our webcam approach achieves mIoU values of $\approx$87\% and generalisation scores of $\approx$71\% and $\approx$69\% across different cameras and winters respectively. Additionally, we put forward a new benchmark dataset of webcam images (\textit{Photi-LakeIce}) which includes data from from two winters and three cameras.}

\begin{document}
%%%%%%%%%%%%%%%%%%%%%%%%%%%%%%%%%%%%%%%%%%

%%%%%%%%%%%%%%%%%%%%%%%%%%%%%%%%%%%%%%%%%%
%\setcounter{section}{-1} %% Remove this when starting to work on the template.
%\section{How to Use this Template}
%The template details the sections that can be used in a manuscript. Note that the order and names of article sections may differ from the requirements of the journal (e.g., the positioning of the Materials and Methods section). Please check the instructions for authors page of the journal to verify the correct order and names. For any questions, please contact the editorial office of the journal or support@mdpi.com. For LaTeX related questions please contact latex@mdpi.com.
%The order of the section titles is: Introduction, Materials and Methods, Results, Discussion, Conclusions for these journals: aerospace,algorithms,antibodies,antioxidants,atmosphere,axioms,biomedicines,carbon,crystals,designs,diagnostics,environments,fermentation,fluids,forests,fractalfract,informatics,information,inventions,jfmk,jrfm,lubricants,neonatalscreening,neuroglia,particles,pharmaceutics,polymers,processes,technologies,viruses,vision
%\input{section1_introduction}
\section{Introduction}
%Climate change is no more a myth, it is a reality. In the words of Stephen Haddrill (2014), "Climate change is a reality that is happening now, and that we can see its impact across the world". The Arctic sea ice is melting, sea levels are rising, snow cover is decreasing, ice sheets are shrinking, glaciers are retreating, oceans are warming, extreme events such as floods, droughts, forest fires etc. have increased significantly in the past few years and the global temperature is on the rise.
Climate change is one of the main challenges for humanity today, and there is a great necessity to observe and understand the climate dynamics and quantify its past, present and future state \cite{rolnick2019arXiv,Sharma2019_Journal_Nature_ClimateChange}. Lake observables such as ice duration, freeze-up and break-up dynamics etc. play an important role in understanding climate change and provide a good opportunity for long-term monitoring. \textit{Lake ice} (as part of \textit{Lakes}) is therefore considered an Essential Climate Variable (ECV) \cite{WMO2020} of the Global Climate Observing System (GCOS). In addition, European Space Agency (ESA) encourages climate research and long-term trend analysis through the Climate Change Initiative (CCI \cite{CCI}, CCI+ \cite{CCIplus}). This consortium recently addressed the following variables: \textit{lake water level, lake water extent, lake surface water temperature, lake ice and lake water reflectance}. Recent research also emphasises the socio-economic and biological role of lake ice \cite{Knoll2019}. Moreover, according to an analysis of data from 513 lakes, winter ice in lakes is depleting at a record pace due to global warming~\cite{Sharma2019_Journal_Nature_ClimateChange}. That study also underlined the importance of lake ice monitoring, observing that a comprehensive, large-scale assessment of lake ice loss is still missing. The vanishing ice affects winter tourism, cold-water ecosystems, hydroelectric power generation, water transportation, freshwater fishing etc., which further emphasises the need to monitor lake ice in an efficient and repeatable manner \cite{Schindler1990_science_lakeice}.
%\textcolor{blue}{A recent investigation  \cite{Sharma2019_Journal_Nature_ClimateChange} expects freshwater ice to significantly deplete in the coming years. 
Interestingly, an investigation of the long-term ice phenological patterns in Northern Hemisphere lakes~\cite{Magnuson1743_science_lakeice_trends} observed trends towards later freeze-up (average shift of 5.8 days per 100 years) and earlier break-up (average shift of 6.5 days per 100 years), which was also confirmed by another overview \cite{Duguay_lake_river_ice2015}. However, a previous 50 year (1951–2000) study \cite{Duguay2006journal} based on Canadian lakes confirmed the earlier break‐up trend but reported less of a clear trend for freeze‐up dates.
\par
The idea of monitoring lake ice for climate studies is not new in the cryosphere research community. A main requirement for monitoring lake ice is high temporal resolution (daily) with an accuracy of $\pm$2 days for phenological events such as ice-on/off dates (according to GCOS). Among the data sources that fulfil this requirement are optical satellite images such as MODIS and VIIRS. In the following, we delve deeper into the literature on using optical satellite images and webcams for monitoring lake ice. 
\subsection{Optical satellite images for lake ice monitoring}
At present, satellite images are the only means for systematic, dense, large-scale monitoring applications. Satellite observations with good temporal as well as spatial resolution are ideal for remote sensing of lake ice phenology. Optical satellite imagery such as MODIS and VIIRS offer very good temporal resolution and satisfactory spatial resolution which makes them a good choice. On the other hand, although sensors such as Landsat-8 and Sentinel-2 have good spatial resolution, the insufficient temporal resolution rules them out as main sources for monitoring lake ice. There exists quite some literature which uses optical satellite data for lake ice analysis. Inter-annual changes in the temporal extent and intensity of lake ice and snow cover in Alaska region have been studied using MODIS imagery~\cite{Spencer2008}. In addition, studies by Brown and Duguay \cite{brown2012} and Kropáček et al.~\cite{lakeice_MODIS_Tibet2013} demonstrated that MODIS data is effective for surveying lake ice. The former approach used MODIS and Interactive Multi-sensor Snow and Ice Mapping System (IMS) snow products to monitor daily ice cover changes. The latter derived ice phenology of 59 lakes (area larger than 100$\,$km$^\text{2}$) on the Tibetan Plateau from MODIS 8-day composite data for the period 2001-2010. The estimated area of open water was compared against the area extracted from high-resolution satellite images (Landsat, Envisat/ASAR, TerraSAR-X and SPOT) and achieved an RMS error of 9.6 days. Recently, Qiu et al. \cite{lakeice_MODIS_Tibet2019} derived the daily lake ice extent from MODIS data by employing the snowmap algorithm \cite{snowmap}. The results of this approach were consistent with the reference observations from passive microwave data (AMSR-E and AMSR2, average correlation coefficient of 0.91). Additionally, the MODIS daily snow product was used to derive the lake ice phenology of more than 20 lakes in China (Xinjiang territory) using a threshold-based method \cite{lakeice_MODIS_timeseries_China2020}. On average, the estimated freeze-up start and break-up end dates were respectively 7.33 and 4.73 days different (mean absolute error) compared to the reference dates derived from passive microwave data (AMSR-E and AMSR2). Very recently, another threshold-based technique \cite{Zhang2019_Journal_MODIS} was also proposed using MODIS data which achieved mean absolute error of 5.54 days and 7.31 days for break-up and freeze-up dates respectively.
\par
Lake Ice Cover (LIC), a sub-product of the newly released CCI Lakes \cite{CCI_lakes} product, provides the spatial cover (spatial resolution of 250$\,$m) of lake ice and the associated uncertainty at a daily temporal resolution. At present, LIC is available only for 250 lakes spread across the globe. However, none of the target lakes in Switzerland are included in this list. Hence, a direct comparison with our results is not feasible. Lake Ice Extent (LIE) \cite{LIE} is one of the Copernicus Global Land Service near-real-time products derived by thresholding the Top-of-Atmosphere reflectances from Level 1B calibrated radiances of Terra MODIS (Collection 6) for snow-covered ice, snow-free ice and water. The 250m resolution product have been validated against ice break-up observations over 34 Finnish lakes spanning four years (2010-2013). However, the LIE product has high uncertainty during the lake freezing period due to low light conditions in the higher latitudes as well as uncertainty in cloud cover detection. In addition, the LIE differs by an average of 3.3 days compared to the in-situ ground truth, not quite meeting the GCOS specification. MODIS snow product \cite{snowmap,snowmap6.1} maps snow and ice cover (including ice on large, inland lakes) at a relatively coarse spatial resolution of 500$\,$m and daily temporal resolution using Earth Observation System (EOS) MODIS data. Comparison of specifications of our machine learning-based product with the operational products is shown in Table \ref{table:operational_products_comparison}.
\begin{table}[ht!]
\small
	    \centering
	    \vspace{-0.5em}
	    \begin{tabular}{ cccccc } 
		\hline
        	   &  &  &  &  &\\ 
	     & \textbf{CCI LIC} \cite{CCI_lakes} & \textbf{LIE} \cite{LIE} & \textbf{MODIS SP} \cite{snowmap} & \textbf{VIIRS SP} \cite{VIIRS_product} & \textbf{Ours}\\ 
		   &  &  &  &  &\\ 
		Temporal resolution & 1 day & 1 day & 1 day & 1 day & 1 day\\ 
		Spatial resolution (GSD)   & 250$\,$m & 250$\,$m & 500$\,$m & 375$\,$m & 250$\,$m \\ 
		Input data & MODIS & MODIS & MODIS & VIIRS & MODIS, VIIRS \\ 
		   &  &  & & &\\ 
		\hline
	    \end{tabular}
	    \vspace{-0.5em}
	    \caption{Comparison of specifications of our machine learning-based product with the operational products such as CCI Lake Ice Cover (CCI LIC), Lake Ice Extent (LIE), MODIS Snow Product (MODIS SP) and VIIRS Snow Product (VIIRS SP).}
	    \label{table:operational_products_comparison}
	\end{table}
\normalsize
\par
Though in many aspects VIIRS and MODIS imagery are similar \cite{Trishchenko2019_MODIS_VIIRS,Trishchenko2017_2}, the former has not been leveraged much to study lake ice. Previously, Sütterlin et al. \cite{Suetterlin2017} proposed to use VIIRS data to retrieve lake ice phenology in Swiss lakes using a threshold approach. Another algorithm  \cite{Liu2016_VIIRS} used VIIRS to detect inland lake ice in Canada. Using VIIRS as well as MODIS, Trishchenko and Ungureanu \cite{Trishchenko2018_IGARSS} constructed a long time series over Canada and neighbouring regions. They also developed ice probability maps using both sensors. Various approaches have been proposed using the Landsat-8 and/or Sentinel-2 optical satellite images \cite{Barbieux_2018_Journal,Williamson2018_Journal_L8S2,Miles2018Journal_OLI_SAR,lakesize_lakeice_SAR2018}. However, we do not go into the details since our work is focused on sensors with at least daily coverage.
\subsection{Webcams for lake ice monitoring}
To some extent, satellite remote sensing can be substituted by close-range webcams \cite{Jacobs2009_Webcams}, especially in cloudy scenarios. As far as we know, the \textit{FC-DenseNet (Tiramisu)} model \cite{Jegou2016_CVPRW} of Xiao et al. \cite{muyan_lakeice_2018} used terrestrial webcam images for the first time for lake ice monitoring application, followed by a joint approach \cite{LIP1_final_report_2019} which used in-situ temperature and pressure observations, and a satellite-based technique in addition to webcams. We note that these two works presented results only on cameras that capture a single lake (St.~Moritz) and the generalisation performance was poor, especially the cross-camera predictions. In this work we achieve better prediction performance using webcams compared to these approaches. In addition, we report results on data from two lakes (St.~Moritz, Sihl) and two winters (2016-17, 2017-18).
\subsection{Machine (deep) learning approaches for lake ice monitoring}
The literature on lake ice monitoring is vast. However, most works make use of elementary threshold-based or index-based methodologies. While, machine learning approaches have been successfully leveraged for various remote sensing and environment monitoring applications, their use for lake ice detection remains under explored. We intend to fill this research gap in our paper. To our knowledge, the previous version of our satellite-based method \cite{tom_lakeice_2018} and Xiao et al. \cite{muyan_lakeice_2018} applied machine learning techniques for the first time to detect ice in lakes. Very recently, we also proposed a preliminary version of our webcam-based methodology \cite{rajanie_tom_2020}. In this paper, we extend our works \cite{tom_lakeice_2018,LIP1_final_report_2019,rajanie_tom_2020} and perform thorough experimentation, targeting an operational system for lake ice monitoring. For completeness, we mention that, very recently, we have also explored the possibility to detect lake ice using Sentinel-1 SAR data with deep learning \cite{tom_aguilar_2020}.
\subsection{Motivation and Contributions}
Existing observations and data on lake ice from local authorities, fishermen, ice-skaters, police, internet media, publications etc. are not well documented. Additionally, there has been a significant decrease in the number of such field observations in the past two decades \cite{Lenormand2002,Duguay_lake_river_ice2015}. At the same time, the potential of different remote sensing sensors has been demonstrated to measure the occurrence of lake ice.
In this context we note that, for our target region of Switzerland, the database at the National Snow and Ice Data Centre (NSIDC) currently includes only the ice-on/off dates of a single lake (St.~Moritz), and only until 2012. Given the need for automated, continuous monitoring of lake ice, we propose to explore the potential of artificial intelligence to support an operational system. In this paper, we put forward a system which monitors selected Alpine lakes in Switzerland and detects the spatio-temporal extent of ice and in particular the \textit{ice-on/off} dates. Though satellite data is the best operational input for global coverage, close-range webcam data can be very valuable in regions with large enough camera networks (including Switzerland). Firstly, we use low spatial resolution (250-1000$\,$m) but high temporal resolution (1 day) multispectral satellite images from two optical satellite sensors (Suomi NPP VIIRS~\cite{SuomiNPP}, Terra MODIS~\cite{TerraMODIS}). Here, we tackle lake ice detection using XGBoost \cite{xgboost} and Support Vector Machines (SVM) \cite{SVM}. Secondly, we investigate the potential of images from freely available webcams using Convolutional Neural Networks (CNN), for independent estimation of lake ice. Given an input webcam image, such networks are designed to predict pixel-wise class probabilities. Additionally, we use webcam data for validation of results from satellite data.
\section{Materials and Methods}
\subsection{Definitions used}
By definition, \textit{ice-on} date is the first day a lake is totally or in great majority frozen ($\approx$90\%), with a similar day after it (same definition as in Franssen and Scherrer \cite{franssen_scherrer2008}, i.e. end of \textit{freeze-up}). \textit{Ice-off} is used here as the symmetric of ice-on, i.e., the first day after having all or almost all lake frozen, when any significant amount of clear water appears and in the subsequent days this water area increases. We point out that our ice-off date marks the start of melting (\textit{break-up}), such that the two dates symmetrically delimit the fully frozen period. As far as we know there is no universally accepted definition of ice-on/off dates. Hence, in the scientific cooperation we had with MeteoSwiss we adopted the above definition which is consistent throughout this work. Ice thickness plays no role in this definition. In very rare cases, in Switzerland, there may be more than one such date. In those cases we use the latest ice-on and the earliest ice-off dates. Some researchers, especially in North America and in the NSIDC database, define ice-off as the end of break-up, when almost everything is water \cite{Duguay_lake_river_ice2015}. That date can also be retrieved with our scheme, without any changes to the methodology. \textit{Clean pixels} are those that are totally within the lake outline. In all subsequent investigations with satellite image data, only the cloud-free clean pixels are used. Additionally, \textit{non-transition} dates are the days when a lake is mostly ($\approx$90\% or above) frozen (ice, snow) or non-frozen (water) while the partially frozen days are termed as \textit{transition} dates.
\subsection{Target lakes and winters}
\begin{figure}[h!]
\centering
	\begin{tabular}{@{}cc@{}}
	    \includegraphics[width=0.99\linewidth]{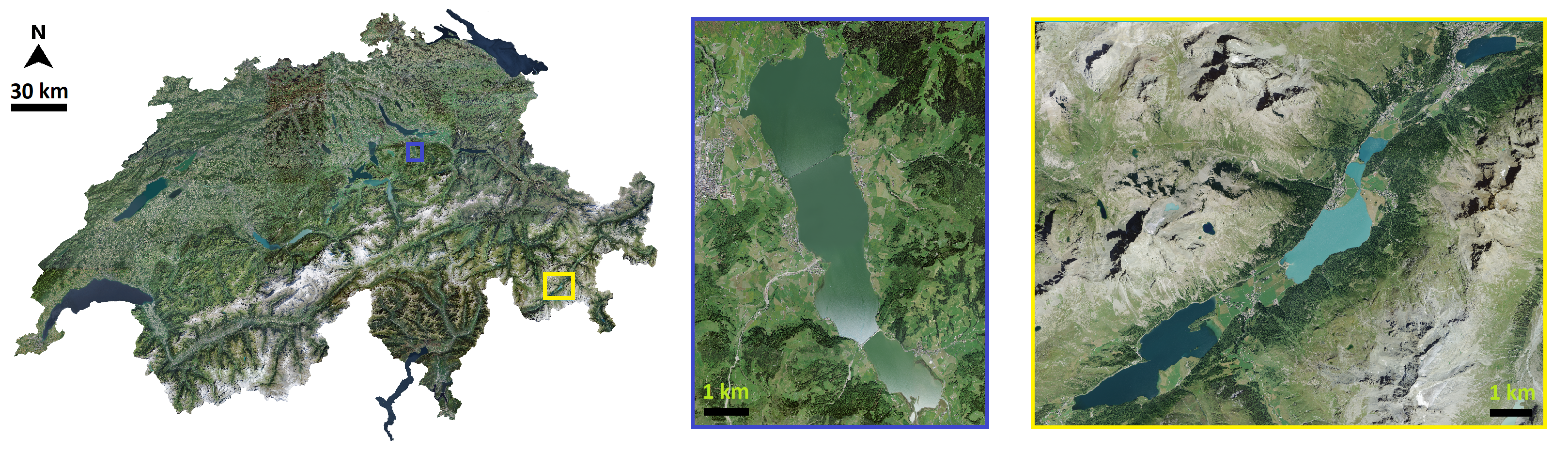}\\
	    \includegraphics[width=0.99\linewidth]{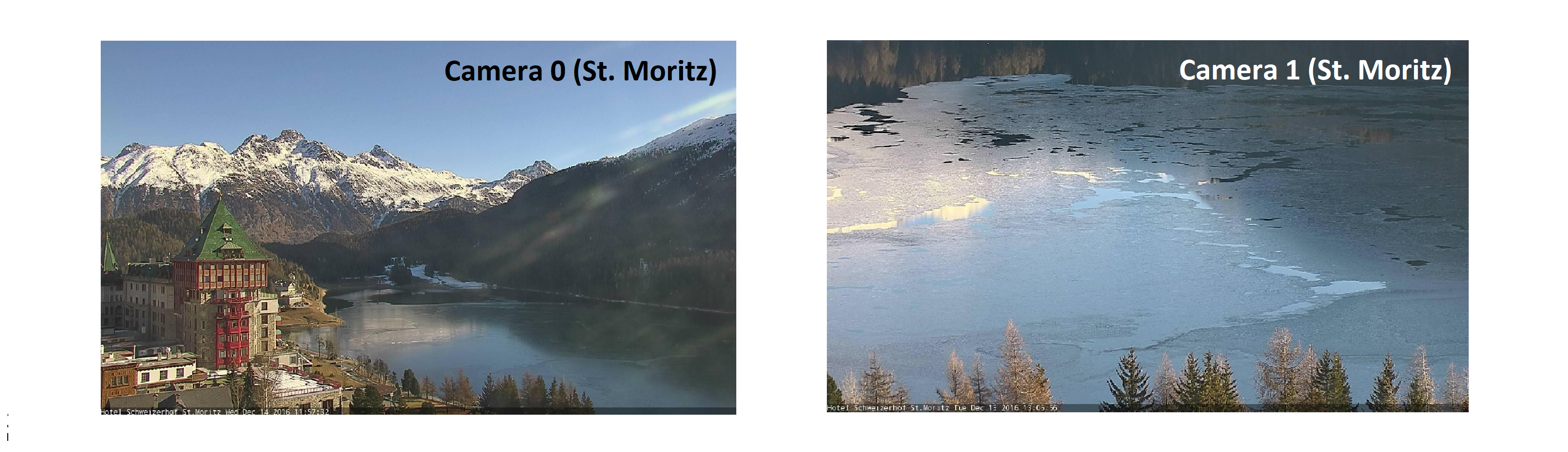}
	\end{tabular}
	%\vspace{-1.5mm}
	\caption{On the first row, left image shows the orthophoto map of Switzerland (source: Swisstopo \cite{swisstopo}). Regions around the four target lakes (shown as blue and yellow rectangles on the map) are zoomed in and shown on the right side of the map (lake Sihl on the left, region around lakes Sils, Silvaplana, St.~Moritz on the right). On the second row, the image footprints of two webcams monitoring lake St. Moritz are displayed (Camera 0 and 1 images were captured on 14.12.2016 and 13.12.2016  respectively when the lake was partially frozen). Best if viewed on screen.}
	\label{fig:target_lakes}
\end{figure}

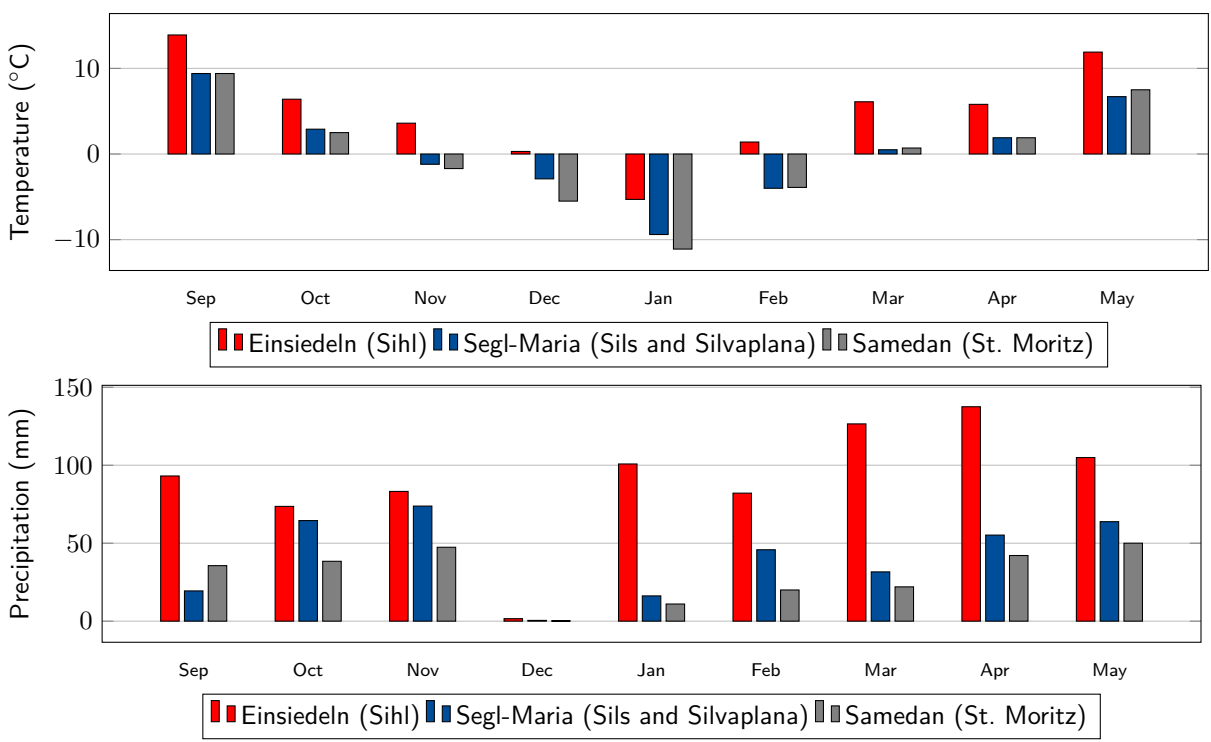
\begin{figure}[h]
    \begin{tikzpicture}
        \centering
        \begin{axis}
        [
            width  = 1.0\linewidth,
            height = 5cm,
            major x tick style = transparent,
            ybar,
            bar width=7pt,
            ymajorgrids = true,
            ylabel = {Temperature ($^{\circ}$C)},
            symbolic x coords={Sep, Oct,Nov,Dec,Jan,Feb,Mar,Apr,May},
            xtick = data,
            scaled y ticks = false,
            x tick label style={font=\scriptsize,text width=2.5cm,align=center},
            enlarge x limits=0.1,
            legend style={at={(0.5,-0.200)}, anchor=north,legend columns=-1},
        ]
            \addplot[style={BLACK,fill=RED,mark=none}]
                coordinates {(Sep,13.9) (Oct,6.4) (Nov,3.6) (Dec,0.3) (Jan,-5.3) (Feb,1.4) (Mar,6.1) (Apr,5.8) (May,11.9) };
            \addplot[style={BLACK,fill=BLUE2,mark=none}]
                coordinates {(Sep,9.4) (Oct,2.9) (Nov,-1.2) (Dec,-2.9) (Jan,-9.4) (Feb,-4.0) (Mar,0.5) (Apr,1.9) (May,6.7) };
            \addplot[style={BLACK,fill=DARKSILVER,mark=none}]
                coordinates {(Sep,9.4) (Oct,2.5) (Nov,-1.7) (Dec,-5.5) (Jan,-11.1) (Feb,-3.9) (Mar,0.7) (Apr,1.9) (May,7.5) };
           
            \legend{Einsiedeln (Sihl), Segl-Maria (Sils and Silvaplana), Samedan (St.~Moritz)}
        \end{axis}
    \end{tikzpicture}\\
    \begin{tikzpicture}
        \centering
        \begin{axis}
        [
            width  = 1.0\linewidth,
            height = 5cm,
            major x tick style = transparent,
            ybar,
            bar width=7pt,
            ymajorgrids = true,
            ylabel = {Precipitation (mm)},
            symbolic x coords={Sep, Oct,Nov,Dec,Jan,Feb,Mar,Apr,May},
            xtick = data,
            scaled y ticks = false,
            x tick label style={font=\scriptsize,text width=2.5cm,align=center},
            enlarge x limits=0.1,
            legend style={at={(0.5,-0.200)}, anchor=north,legend columns=-1},
        ]
            \addplot[style={BLACK,fill=RED,mark=none}]
                coordinates {(Sep,93.1) (Oct,73.6) (Nov,83.2) (Dec,1.6) (Jan,100.8) (Feb,82.1) (Mar,126.5) (Apr,137.5) (May,104.9) };
            \addplot[style={BLACK,fill=BLUE2,mark=none}]
                coordinates {(Sep,19.4) (Oct,64.5) (Nov,73.8) (Dec,0.5) (Jan,16.2) (Feb,45.8) (Mar,31.6) (Apr,55.2) (May,63.8) };
            \addplot[style={BLACK,fill=DARKSILVER,mark=none}]
               coordinates {(Sep,35.6) (Oct,38.4) (Nov,47.4) (Dec,0.2) (Jan,11.0) (Feb,20.0) (Mar,22.0) (Apr,42.1) (May,50.0) };

            \legend{Einsiedeln (Sihl), Segl-Maria (Sils and Silvaplana), Samedan (St.~Moritz)}
        \end{axis}
    \end{tikzpicture}
    \caption{Bar graphs of mean monthly air temperature 2$\,$m above ground (top) and total monthly  precipitation (bottom) in winter 2016-17, recorded at the meteorological stations closest to the respective lakes. Data courtesy of MeteoSwiss.}
    \label{fig:temperature_and_precipitation_bargraph}
\end{figure}

Using satellite images, we process the Swiss Alpine lakes: \textit{Sihl}, \textit{Sils}, \textit{Silvaplana}, and \textit{St.~Moritz}, see Fig.~\ref{fig:target_lakes}. To assess the performance, the data from two full winters (16-17 and 17-18) are used, including the relatively short but challenging freeze-up and break-up periods. In each winter, we process all the dates available from beginning of September till end of May. The target lakes  exhibit moderate to high difficulty, with variable area (very small to mid-sized), altitude (low to high), and surrounding topography (flat/hilly to mountainous), and they freeze more or less often. More details of the target lakes are shown in Table \ref{table:target_lakes}, including the details of the nearest meteorological stations. For completeness, we also plot the temperature and precipitation data near the observed lakes (see Fig.~\ref{fig:temperature_and_precipitation_bargraph} for 2016-17 winter months). Additionally, we process three different webcams monitoring lakes St.~Moritz and Sihl from the same two winters. For satellite images, the lake outlines are digitised from Open Street Map (OSM) and have an accuracy of $\approx$25-50$\,$m. For webcams, our algorithm automatically determines the lake outline. 
\begin{table}[h!]
\small
	    \centering
	    \begin{tabular}{|ccccc|c|} 
		\hline
		 &  &  &  &  & \\ 
		\textbf{Lake} & \textbf{Lat, Lon, Alt}  & \textbf{Area, Vol}  & \textbf{Depth (M, A)} & \textbf{Remarks}& \textbf{MS, Lat, Lon, Alt} \\ 
		 &  &  &  &  &  \\ 
		Sihl & 47.14, 8.78, 889 & 11.3, 96 & 23, 17  & frozen most years & Einsiedeln, 47.13, 8.75, 910 \\ 
		Sils & 46.42, 9.74, 1797 & 4.1, 137 & 71, 35 & freezes every year & Segl-Maria, 46.43, 9.77, 1804\\ 
		Silvaplana & 46.45, 9.79, 1791 & 2.7, 140  & 77, 48 & freezes every year & Segl-Maria, 46.43, 9.77, 1804 \\
	    St.~Moritz & 46.49, 9.85, 1768 & 0.78, 20 & 42, 26 & freezes every year & Samedan, 46.53, 9.88, 1708 \\
		 &  &  &  &  &  \\ 
		\hline
	    \end{tabular}
	    \caption{Characteristics of the target lakes (data mainly from Wikipedia). Latitude (lat, $^{\circ}$North), longitude (lon, $^{\circ}$East), altitude (alt, m), area (km$^\text{2}$), volume (vol, Mm$^\text{3}$), maximum and average depth [depth(M,A)] in m are shown. Additionally, for each lake, the nearest meteorological station (MS) is shown together with the corresponding latitude, longitude and altitude.}
	    \label{table:target_lakes}
	\end{table}%
\normalsize%
\subsection{Data}
We use data from three different type of sensors for lake ice monitoring. Parameters of all these data types are shown in Table \ref{table:data_params}. 
\begin{table}[ht!]
\small
	    \centering
	    \vspace{-0.5em}
	    \begin{tabular}{ cccc } 
		\hline
        	   &  &  & \\ 
	    &   \textbf{MODIS} &
		                     \textbf{VIIRS} & \textbf{Webcams}\\ 
		   &  &  & \\ 
		Temporal resolution   & 1 day & 1 day & 1 hour (typically)\\ 
		Spatial resolution (GSD)   & 250-1000$\,$m & 375-750$\,$m & ca. 4$\,$mm to 4$\,$m \\ 
		Spectral resolution   & 36 bands (0.41-14.24 $\mu m$) & 22 bands (0.41-
12.01$\mu m$) & RGB\\ 
	    Radiometric resolution & 12 bits & 12 bits & 8 bits\\ 
	    Costs & free & free & free\\ 
	    Availability & very good & very good &  depending on location\\ 
	    Cloud mask issues& slight & slight & NA\\
	    Cloud problems& severe & severe & negligible\\
		   &  &  & \\ 
		\hline
	    \end{tabular}
	    \vspace{-0.5em}
	    \caption{Parameters of the used data (GSD = Ground Sampling Distance).}
	    \label{table:data_params}
	\end{table}
\normalsize
\subsubsection{Optical satellite images}
\label{sec:optical_sat_data}
\begin{figure}[h!]
\centering
  \subfloat[]{\includegraphics[width=.66\linewidth]{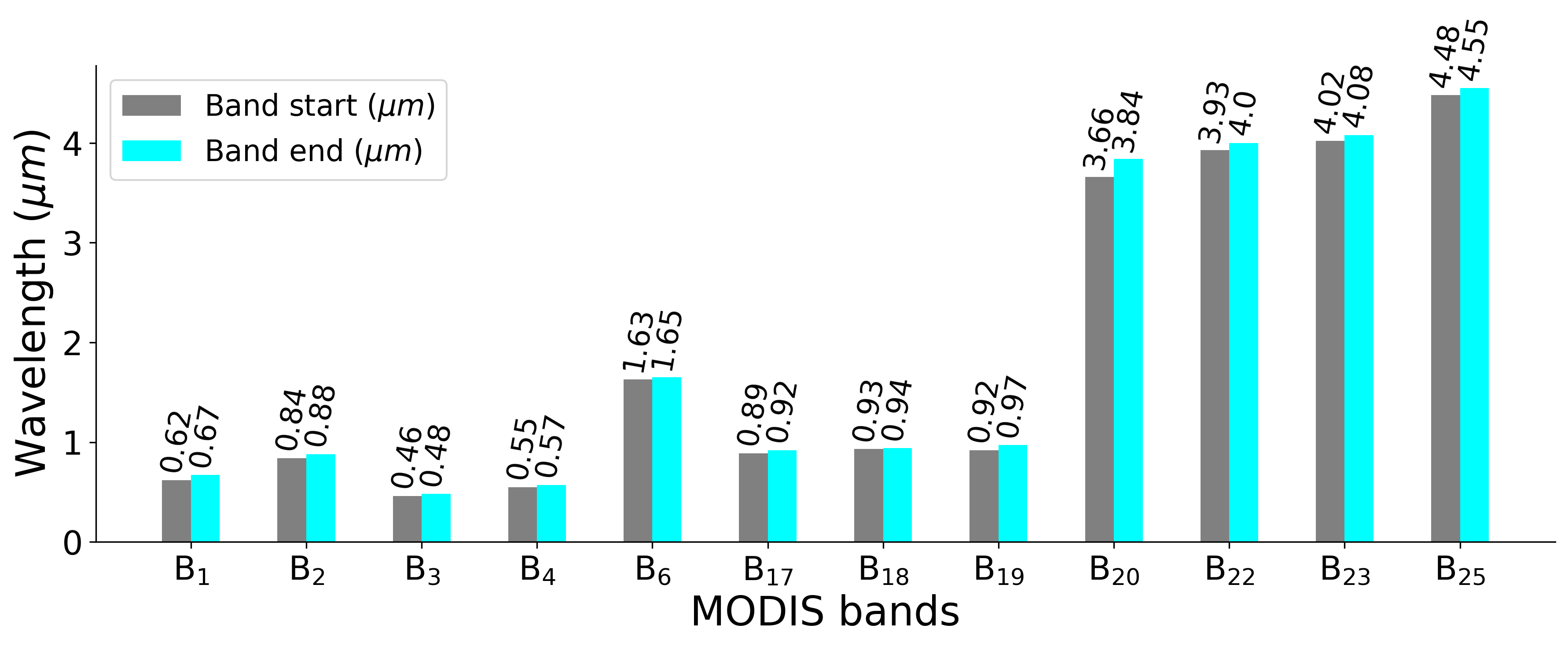}}\hspace{0.1em}
  \subfloat[]{\includegraphics[width=.33\linewidth]{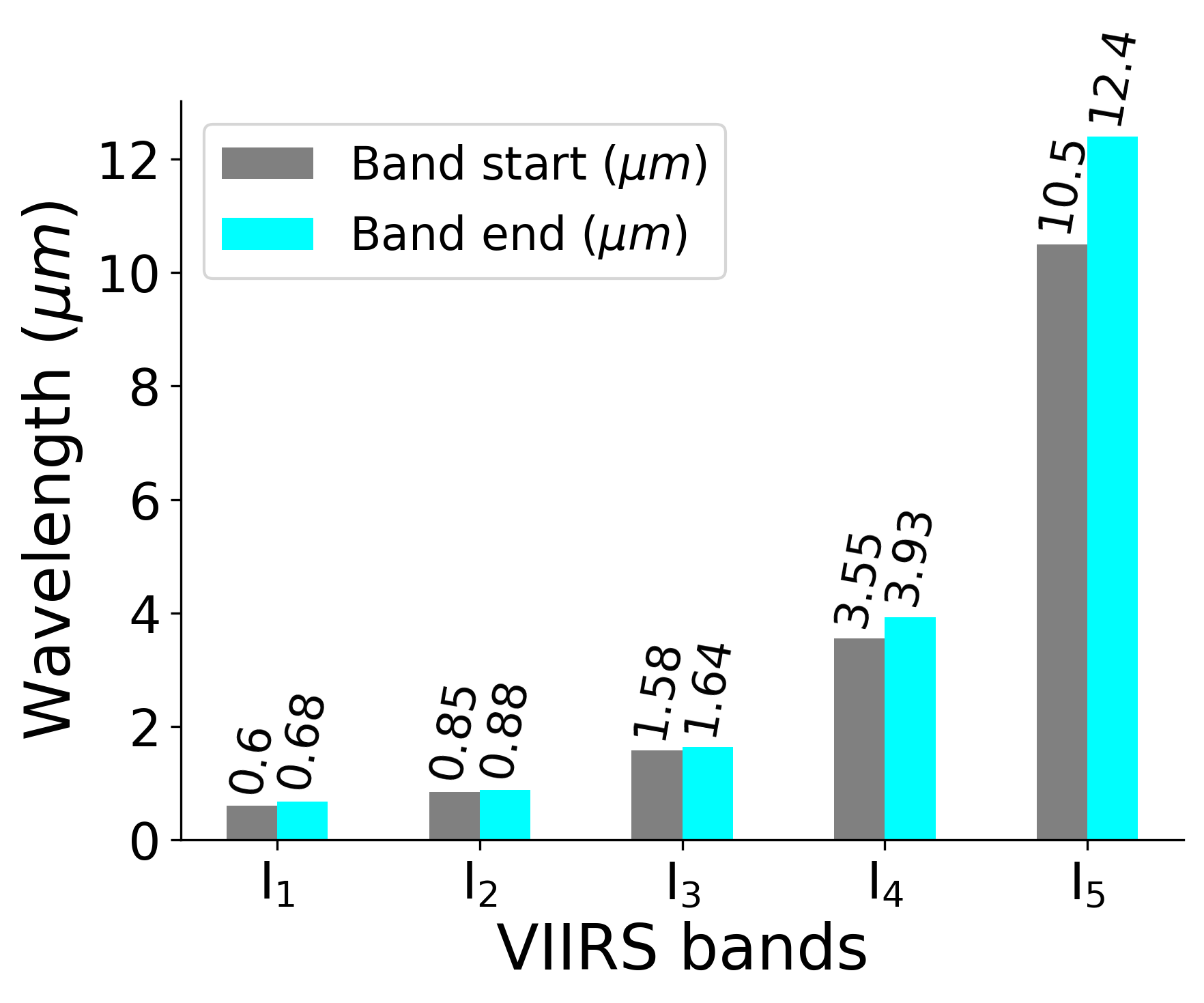}}
  \vspace{-1em}
  \caption{Spectral range of MODIS (left) and VIIRS (right) bands used in our analysis. The start and end wavelengths are shown for each band.} 
\label{fig:spectral_range}
\end{figure}

Both MODIS (aboard Terra \cite{TerraMODIS} and Aqua \cite{AquaMODIS} satellites) and VIIRS (Suomi NPP~\cite{SuomiNPP} satellite) images are freely available and have high temporal resolution. Because of lower quality of Aqua imagery we just use Terra images in our analysis. Additionally, following the approach of Tom et al. \cite{tom_lakeice_2018}, we use only 12 MODIS bands and discard the rest. For our MODIS analysis, we downloaded the following products: MOD02 (calibrated and geolocated radiance, level 1B), MOD03 (geolocation) and MOD35 (48-bit cloud mask) from the LAADS DAAC (Level-1 and Atmosphere Archive \& Distribution System Distributed Active Archive Center) archive \cite{LAADS_DAAC}. Note that, only the I-bands are used in our VIIRS analysis. See Fig.~\ref{fig:spectral_range} for the spectral range of MODIS and VIIRS bands used in our approach.
\par
Technicalities about the satellite data that we process are shown in Tables \ref{table:sat_data_stats} and \ref{table:dataset_stats}. It can be seen that we analyse relatively less data in winter 17-18 as opposed to the previous winter, due to the fact that winter 17-18 was more cloudy than 16-17 in the regions of interest. We process only the dates when a lake is at least 30\% cloud-free, which effectively lowered the temporal resolution from 1 day to approximately 2 days. The effective temporal resolution varies across sensors and winters (see Table \ref{table:dataset_stats}). Additionally, for lakes Sihl and St.~Moritz, there were more transition days in winter 17-18. Throughout, we just use the non-transition dates for training our SVM model, refer Section \ref{experiments:satellites}. This factor along with class-imbalance explains why the decrease in data is more evident for the class \textit{frozen}. Note also that the transition dates are more likely to occur near the freezing and thawing periods. One can note class imbalance in the dataset of both winters. In each winter, we process all the available acquisitions during the period from September till May, while the lakes are typically fully (or mostly) frozen only during a small subset of these dates. Moreover, the class imbalance is alarmingly high for lake Sihl. This is because Sihl has a moderate freezing frequency compared to the other three lakes, because of its lower altitude and larger area, see Table \ref{table:target_lakes}. Note also that the presence of a dam near the northern part of lake Sihl makes its freezing pattern relatively less natural.
\begin{table}
    \centering
    \renewcommand{\arraystretch}{1}
{\small
    \begin{tabular}{|c|c|cc|cc|cc|cc|cc|}
	\hline
	\multirow{2}{*}{} &\multirow{2}{*}{\textbf{Winter}} & \multicolumn{2}{c|}{\textbf{Sihl}} & %
    \multicolumn{2}{c|}{\textbf{Sils}}  & \multicolumn{2}{c|}{\textbf{Silvaplana}} &  \multicolumn{2}{c|}{\textbf{St.~Moritz}} &  \multicolumn{2}{c|}{\textbf{Total}}\\
	\cline{3-12}
	 & & \textbf{M} & \textbf{V} & \textbf{M} & \textbf{V} & \textbf{M} & \textbf{V} & \textbf{M} & \textbf{V} & \textbf{M} & \textbf{V}  \\
	\textbf{Frozen} & 16-17  & 4137 & 1919 & 2345 & 894 & 1736 & 739 & 157 &--- & 8375 & 3552\\
    \textbf{Non-Frozen} & 16-17  & 13568 & 4598 & 3019 & 1051 & 1965 & 765  & 191 &--- & 18743 & 6414\\
     &  &  &  &  &  &  &  &  &  &  &\\
    \textbf{Frozen} & 17-18  & 1005 & 198 & 1858 & 722 & 1169 & 591  & 124 &--- & 4156 & 1511\\
    \textbf{Non-Frozen} & 17-18  & 11804 & 4311 & 2435 & 784 & 1574 & 621  & 140 &--- & 15953 & 5716\\
      &  &  &  &  &  &  &  &  &  &  &\\
     \hline
     \multicolumn{2}{|c|}{\textbf{Total}}  & 30514 & 11026 & 9657 & 3451 & 6444 & 2716 & 612 & --- & 47227  & 17193\\
    \hline
    \end{tabular}
    }%
  \caption{Total number of \textit{clean, cloud-free} pixels on \textit{non-transition dates} from MODIS (M) and VIIRS (V) sensors (at least 30\% cloud-free days) used in our experiments.}
  \label{table:sat_data_stats}
\end{table}
\normalsize
\begin{table}
\centering
\small
\vspace{-0.5em}
\begin{tabular}{cccccccc}
    \hline
                 &  &  &  &  &  & \\
  \textbf{Lake}&\textbf{\#Pixels}&\textbf{Winter}&\multicolumn{2}{c} {\textbf{Non-Transition days}}&\textbf{Transition}&\textbf{Resolution}&\textbf{Trans fraction}\\
    \cline{4-5}   & \textbf{(clean)} &
    & \textbf{Non-Frozen} & \textbf{Frozen} & \textbf{days} & \textbf{(days)} & \\
                 &  &  &  &  &  & \\
    \multirow{2}{*}{Sihl} & \multirow{2}{*}{115 / 45}       & 16-17 & 98 / 87 & 32 / 33 & 12 / 11  & 1.9 / 2.1 & 0.09 / 0.08\\
                       &         & 17-18 & 90 / 88 & 8 / 6 & 24 / 22  & 2.2 / 2.4 & 0.20 / 0.19\\
    \multirow{2}{*}{Sils} & \multirow{2}{*}{33 / 11}      & 16-17 & 70 / 73 & 57 / 59 & 33 / 30 & 1.7 / 1.7 & 0.21 / 0.19\\
                        &        & 17-18 & 60 / 57 & 49 / 48 & 25 / 32 & 2.0 / 2.0 & 0.19 / 0.23 \\ 
    \multirow{2}{*}{Silvaplana} & \multirow{2}{*}{21 / 9} & 16-17 & 66 / 66 & 63 / 59 & 33 / 34 & 1.7 / 1.7 & 0.20 / 0.21\\
                         &       & 17-18 & 58 / 58 & 43 / 54 & 27 / 31 & 2.1 / 1.9 & 0.21 / 0.22\\
    \multirow{2}{*}{St.~Moritz}& \multirow{2}{*}{4 / 0}  & 16-17 & 79 / --- & 65 / --- & 14 / --- & 1.7 / --- & 0.09 / ---\\
                          &      & 17-18 & 64 / --- & 58 / --- & 16 / --- & 2.0 / --- & 0.12 / ---\\ 
                &  &  &  &  &  & \\
                \hline
  \end{tabular}
  \vspace{-0.5em}
  \caption{Dataset statistics. M/V format displays the respective numbers of MODIS/VIIRS. The effective temporal resolution (shown as 'resolution') and fraction of transition dates (Trans fraction) are also shown. \#Pixels (clean) displays the number of clean pixels per acquisition.}
  \label{table:dataset_stats}
 \end{table} 
\normalsize
\par
Even after collecting the data from a full winter, very few pixels are available to train a machine learning-based system using MODIS imagery, due to the low spatial resolution. For instance, in every acquisition, there exist only four clean pixels for lake St.~Moritz, refer Table \ref{table:dataset_stats}. This problem is even worse for VIIRS where there is no clean pixel at all for the same lake (see Table \ref{table:dataset_stats}). Note from Table \ref{table:sat_data_stats} that the total number of VIIRS (clean) pixels we process is significantly less compared to MODIS mainly due to the lower spatial resolution (see also Table \ref{table:dataset_stats}). A challenge for machine learning is the scarcity of lake pixels. Note also that the small number of pixels per lake makes correction of the lake outlines' absolute geolocation a necessity (refer Section \ref{sec:MainBody}). Furthermore, it is highly probable during the transition periods that both frozen and non-frozen classes coexist within a single clean pixel (mixels). For this reason, we also generate the probability for each pixel to be frozen as an end result, especially targeting such mixels during the transition periods. Note also that data hungry deep learning approaches cannot be deployed, as they cannot be reliably trained with such small datasets.
\subsubsection{Webcam images}
\label{sec:webcam_data}
\begin{figure}[h!]
  \centering
  \subfloat{\fbox{\includegraphics[height=2cm,width=3cm]{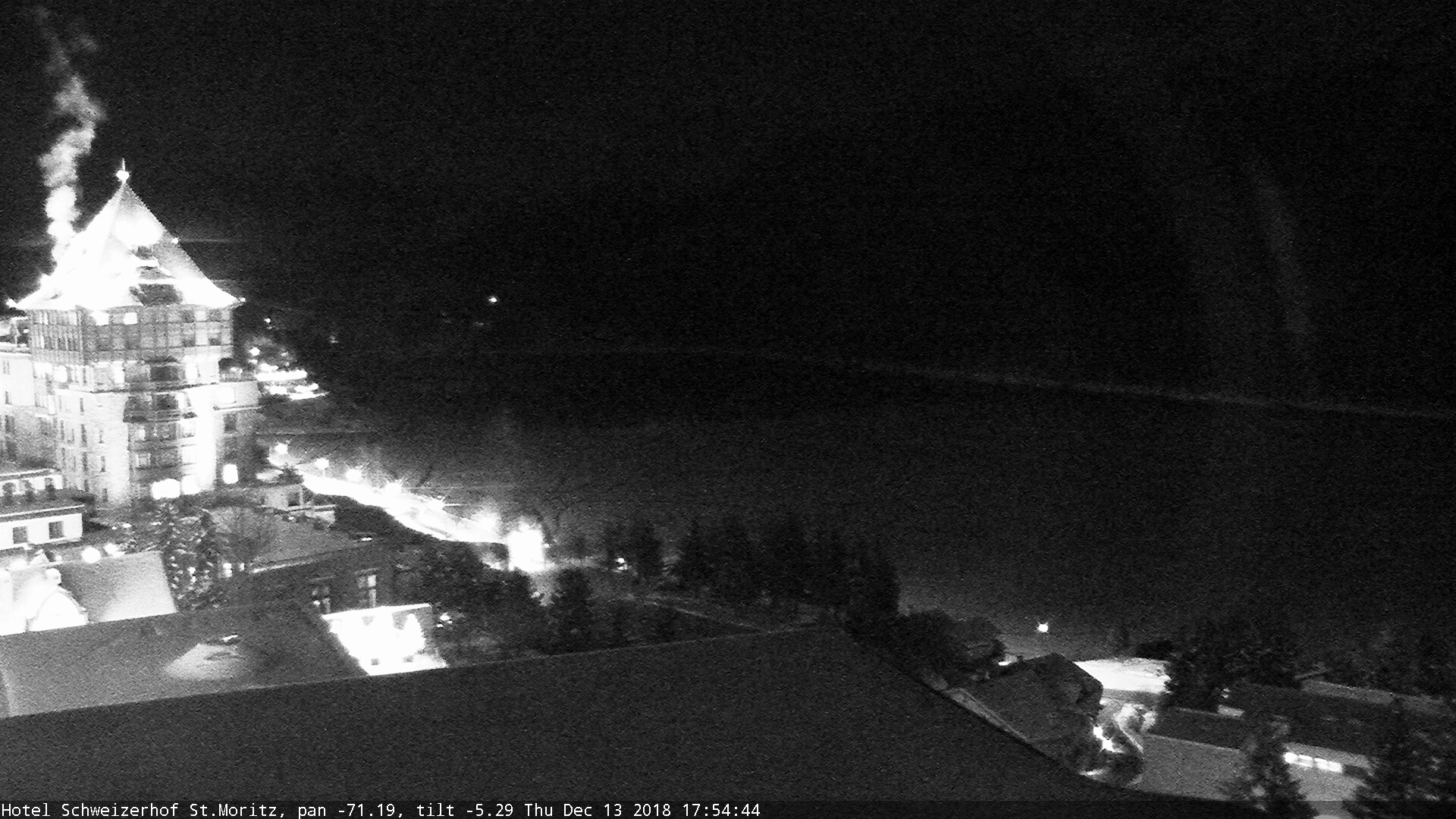}}}\hspace{0.1em}
  \subfloat{\fbox{\includegraphics[height=2cm,width=3cm]{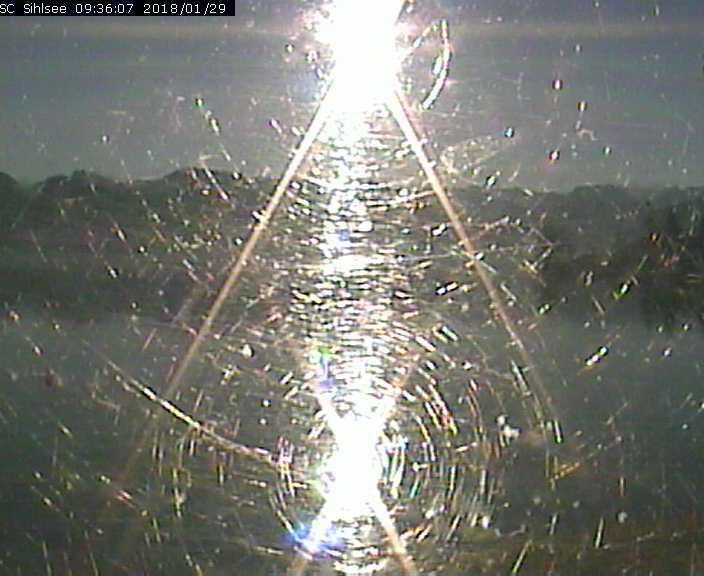}}}\hspace{0.1em}
  \subfloat{\fbox{\includegraphics[height=2cm,width=3cm]{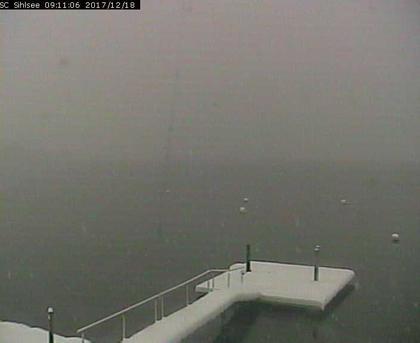}}}
  \caption{Example images that were discarded from the dataset due to bad illumination (left), over sun exposure (mid), and thick fog (right).}
  \label{fig:baddata}
\end{figure}

We report our results on various cameras with different intrinsic and extrinsic parameters, which are freely available from internet. For the experiments, we manually removed some unusable images, examples are shown in Fig.~\ref{fig:baddata}. We point out that for oblique webcam viewpoints, the GSD varies greatly between nearby and distant parts of a lake, as does the angle between the viewing rays and the lake surface. As a consequence, webcam images are hard to interpret in the far field, in practice usable distances tend to be up to $\approx$1$\,$km. We note that the usable distance also depends on the surface material, e.g., snow on ice can be detected at further distances where it is already impossible (for humans as well as machines) to distinguish black ice from water.
\par
\begin{figure}[t!]
  \footnotesize
  \centering
  \subfloat[Camera 0(w)]{\fbox{\includegraphics[height=2.25cm,width=3.56cm]{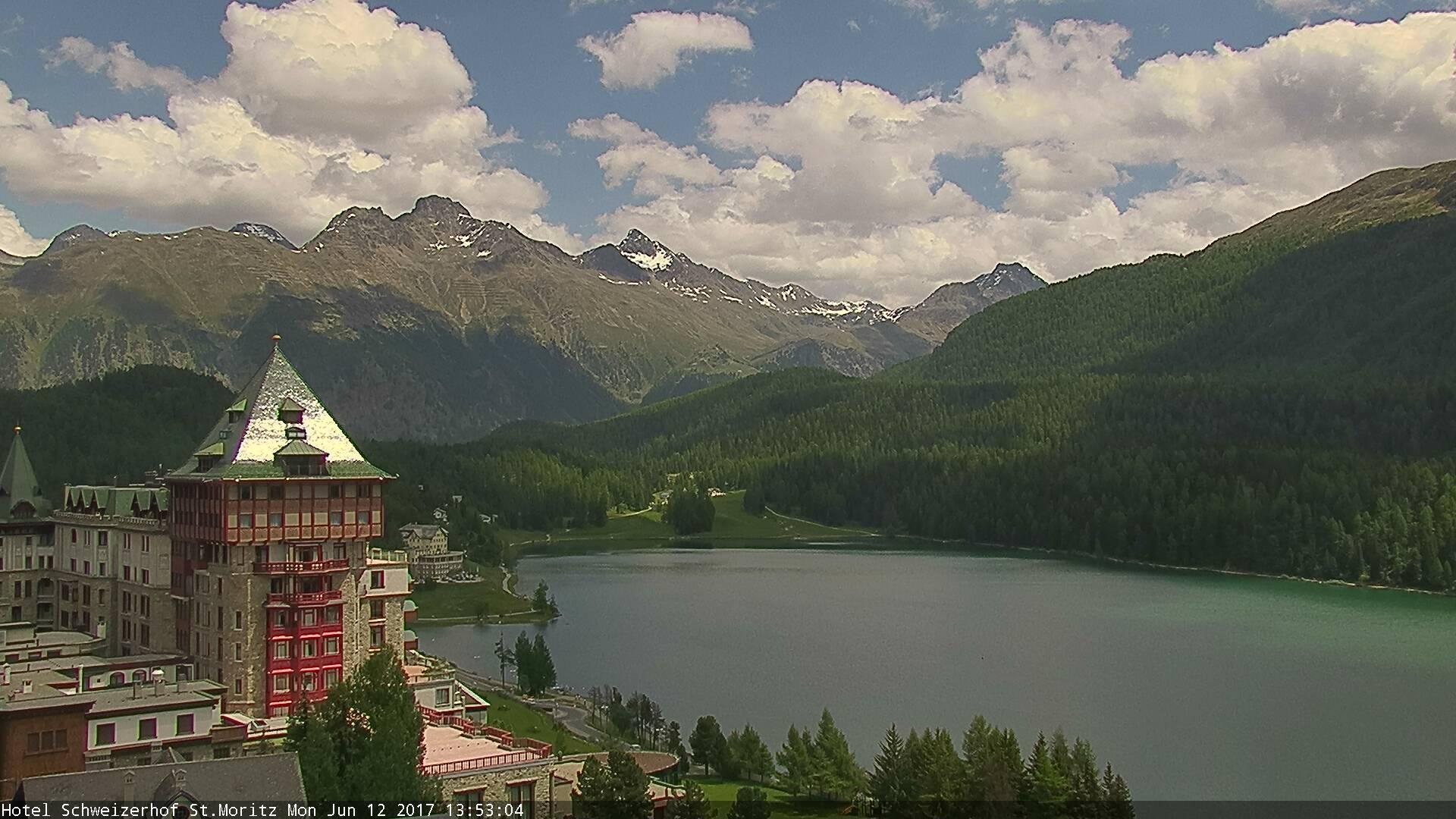}}}\hspace{0.001cm}
  \subfloat[Camera 0(w + i)]{\fbox{\includegraphics[height=2.25cm,width=3.56cm]{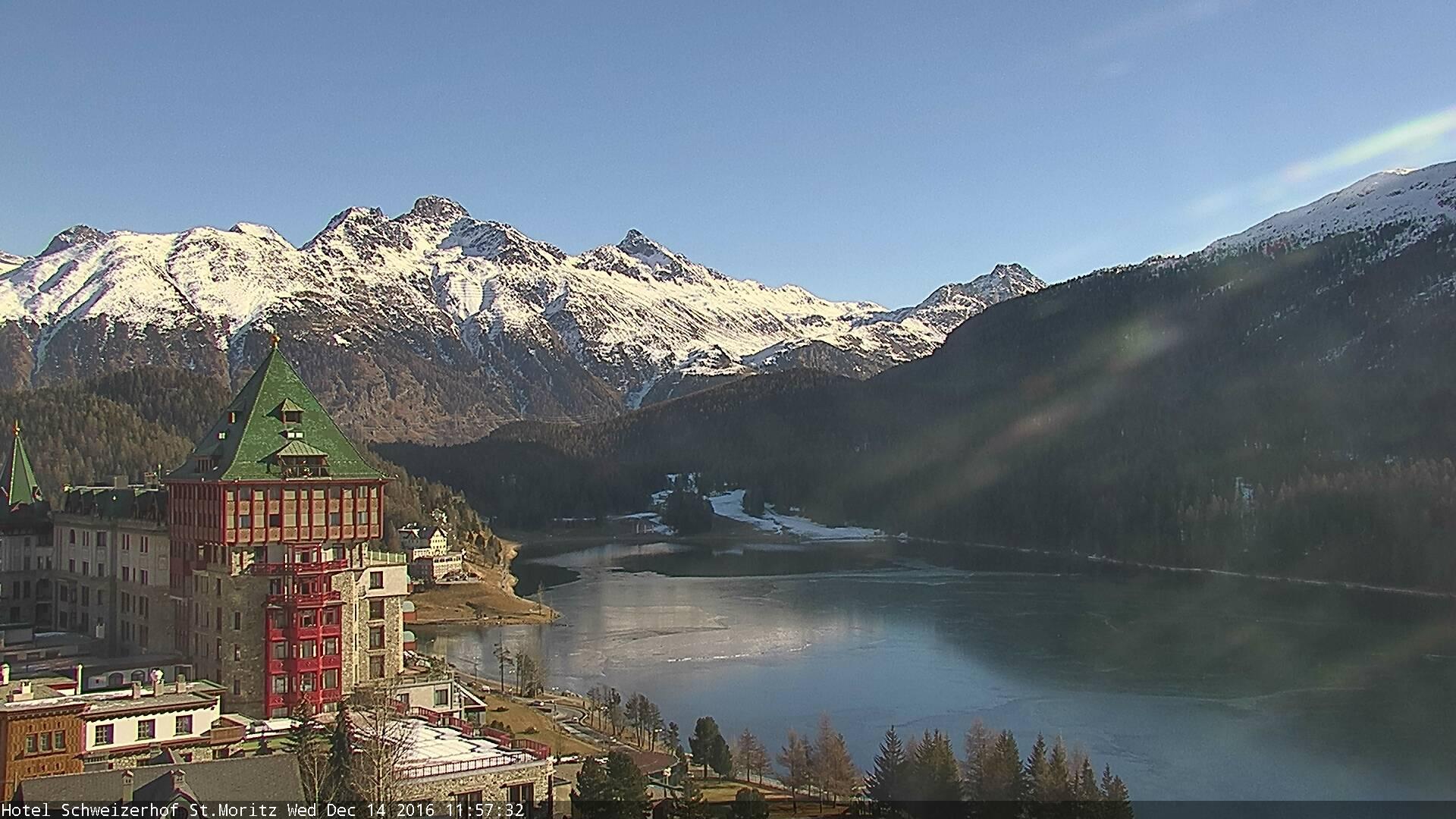}}}\hspace{0.001cm}
  \subfloat[Camera 0(s + c)]{\fbox{\includegraphics[height=2.25cm,width=3.56cm]{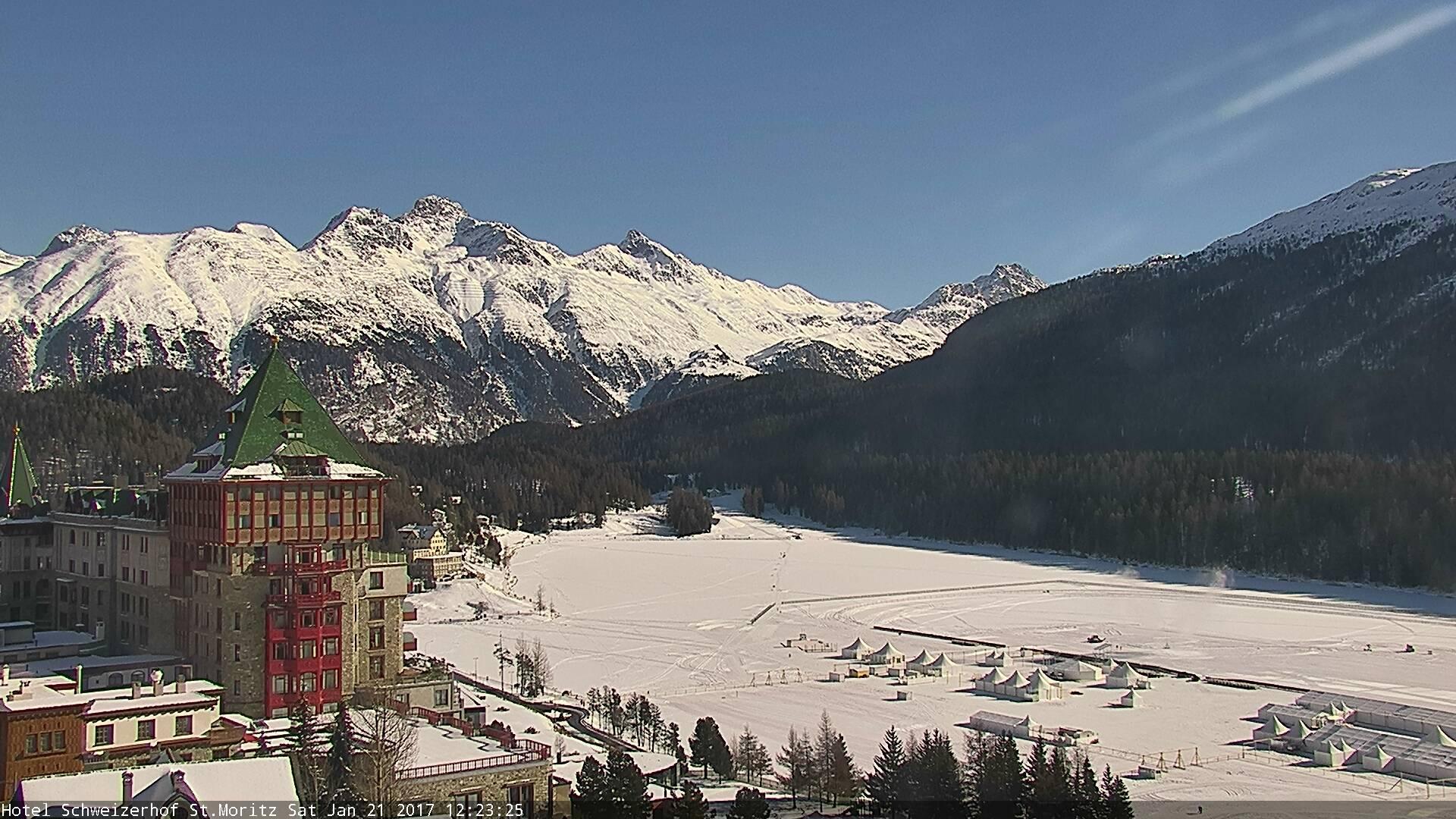}}}\hspace{0.001cm}
  \subfloat[Camera 0(s)]{\fbox{\includegraphics[height=2.25cm,width=3.56cm]{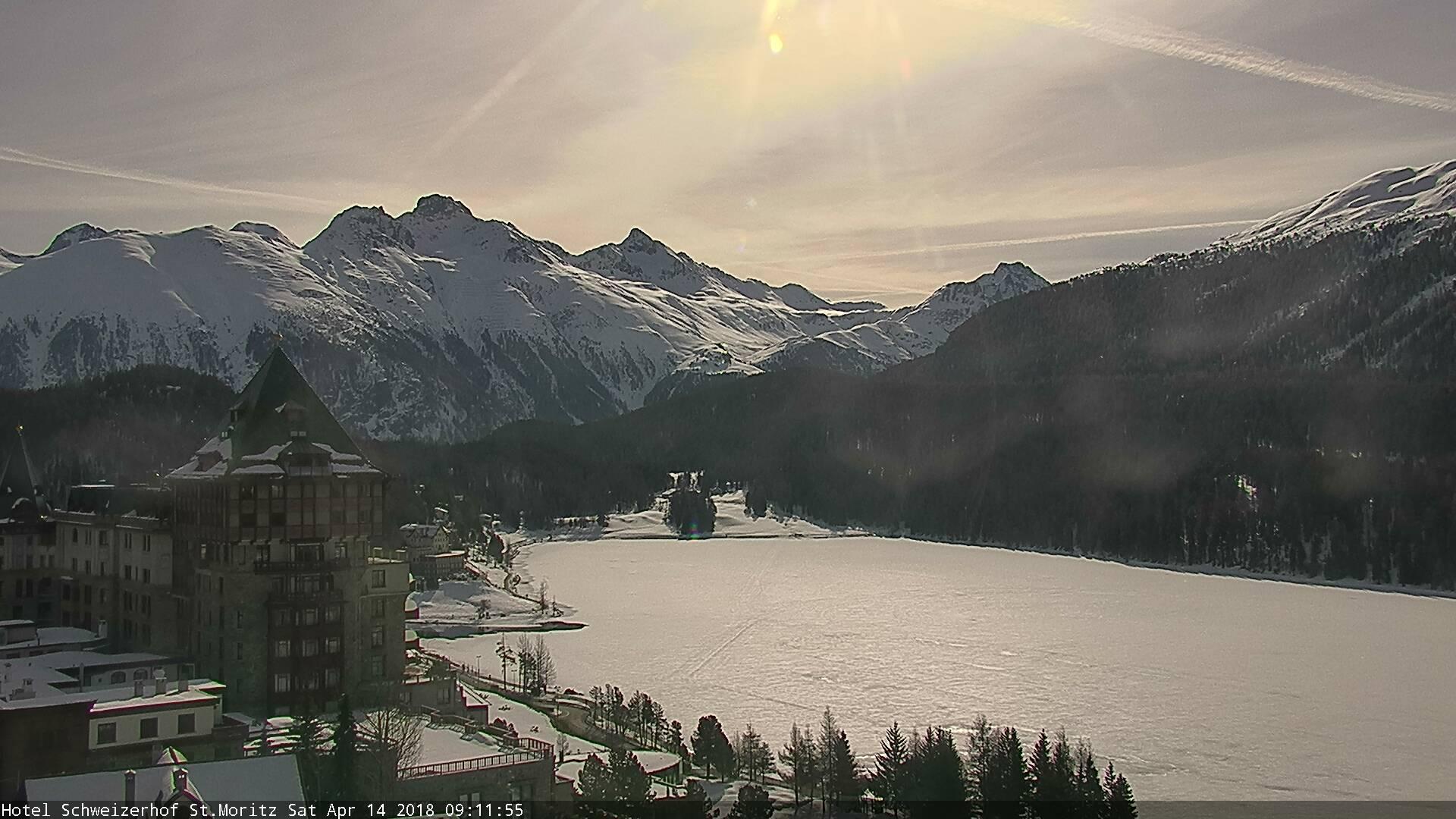}}}\\
  \vspace{-0.65em}
  \subfloat[Camera 1(w)]{\fbox{\includegraphics[height=2.25cm,width=3.56cm]{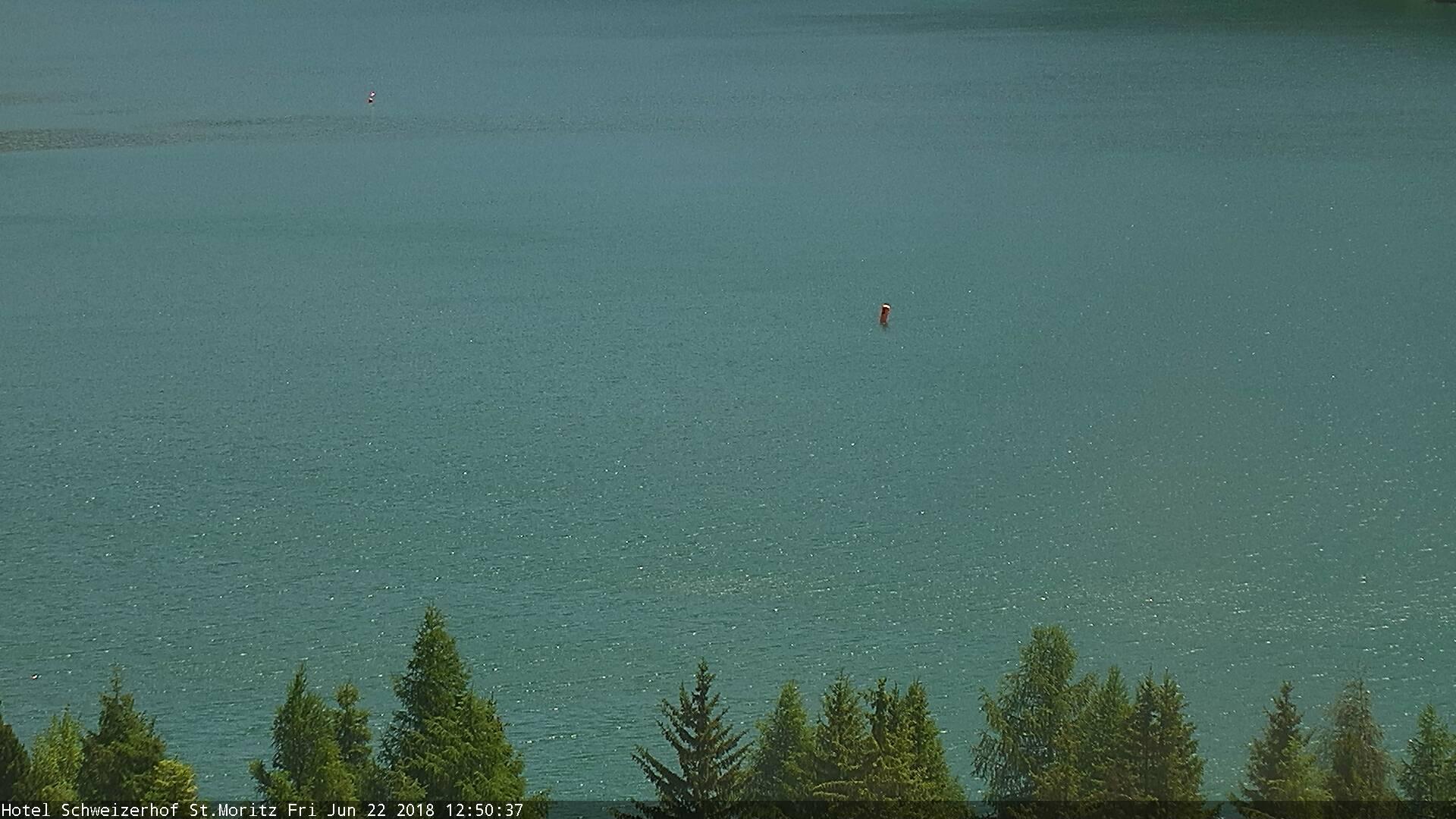}}}\hspace{0.001cm}
  \subfloat[Camera 1(w + i)]{\fbox{\includegraphics[height=2.25cm,width=3.56cm]{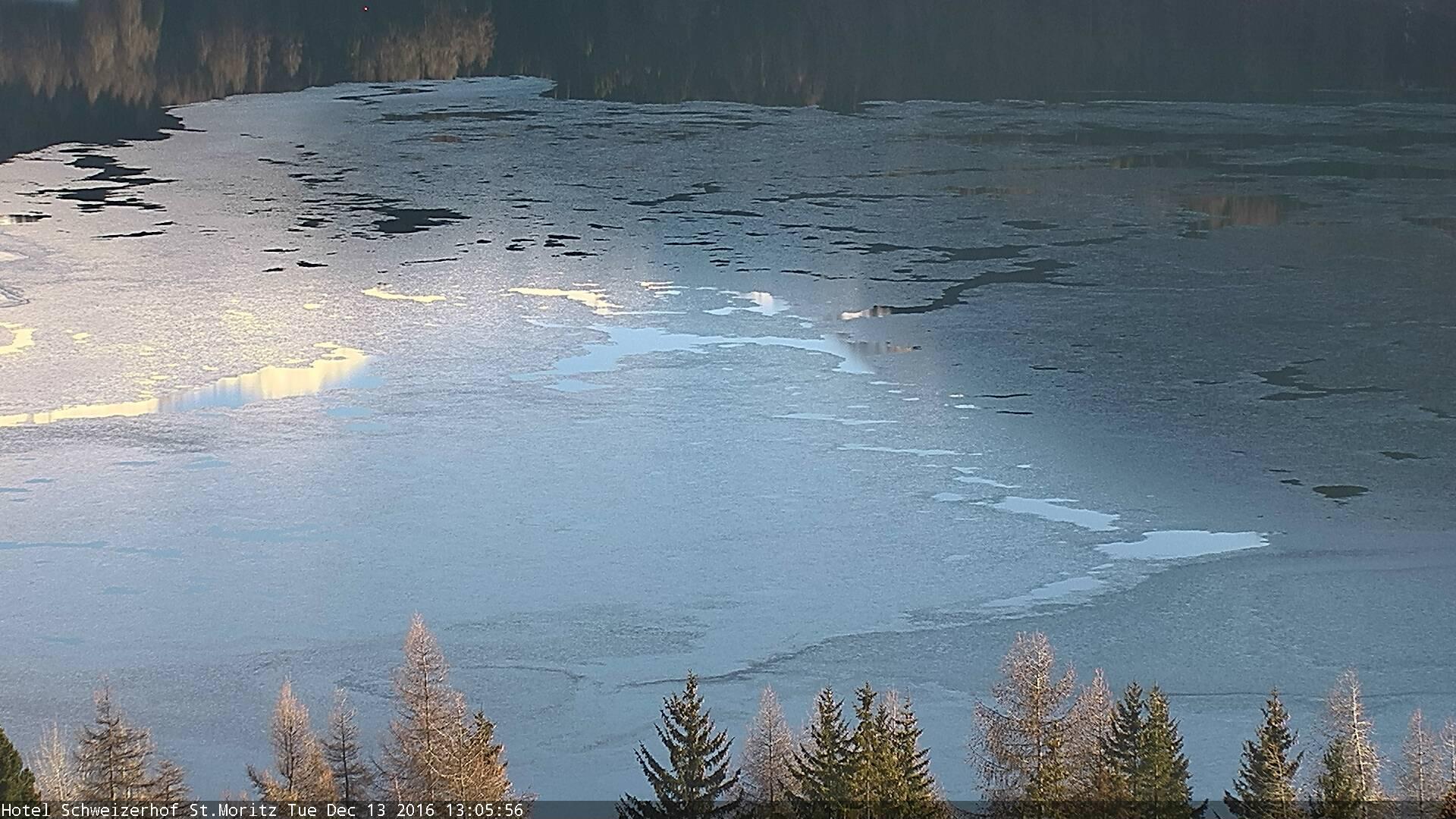}}}\hspace{0.001cm}
  \subfloat[Camera 1(s + c)]{\fbox{\includegraphics[height=2.25cm,width=3.56cm]{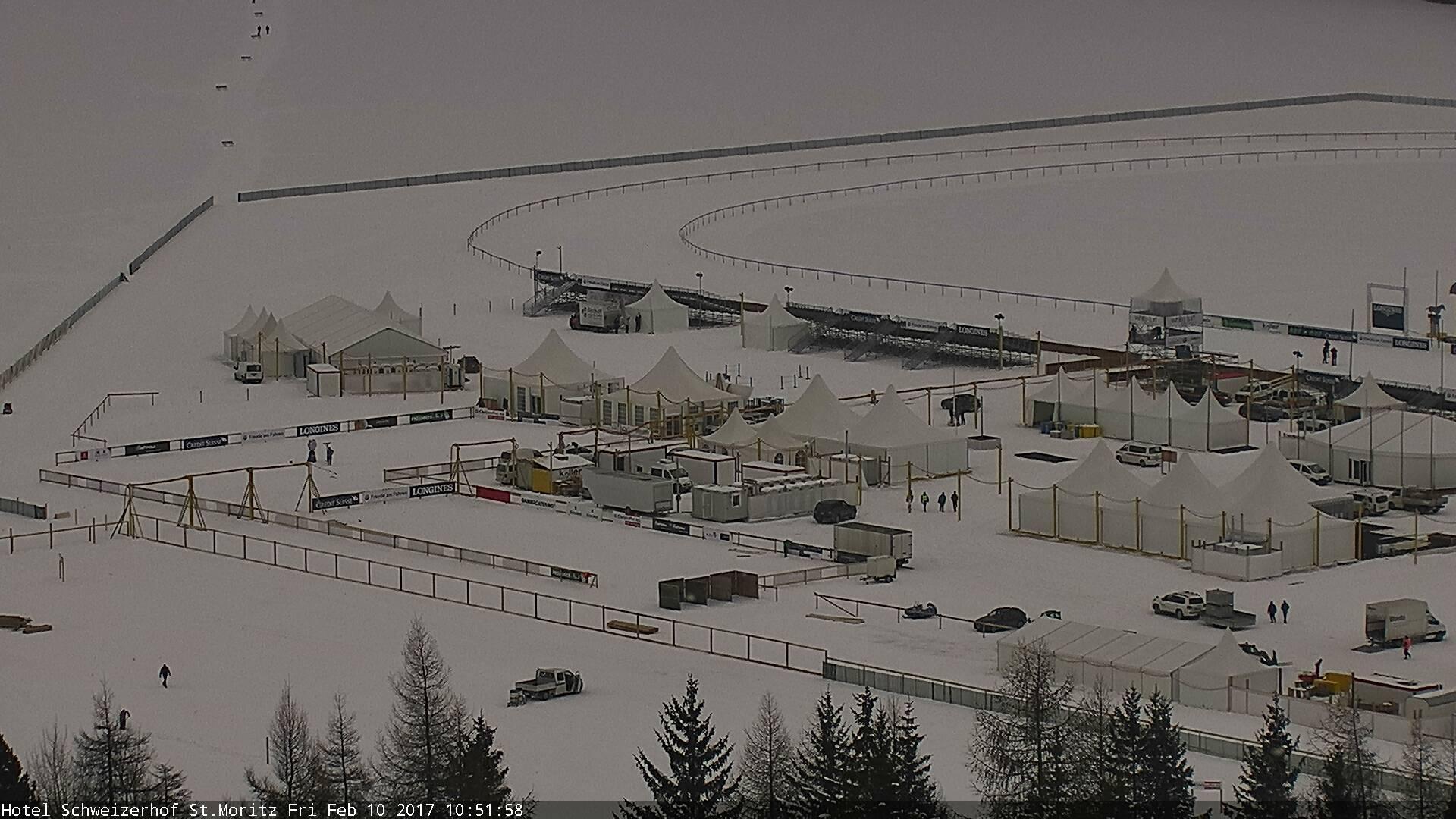}}}\hspace{0.001cm}
  \subfloat[Camera 1(s)]{\fbox{\includegraphics[height=2.25cm,width=3.56cm]{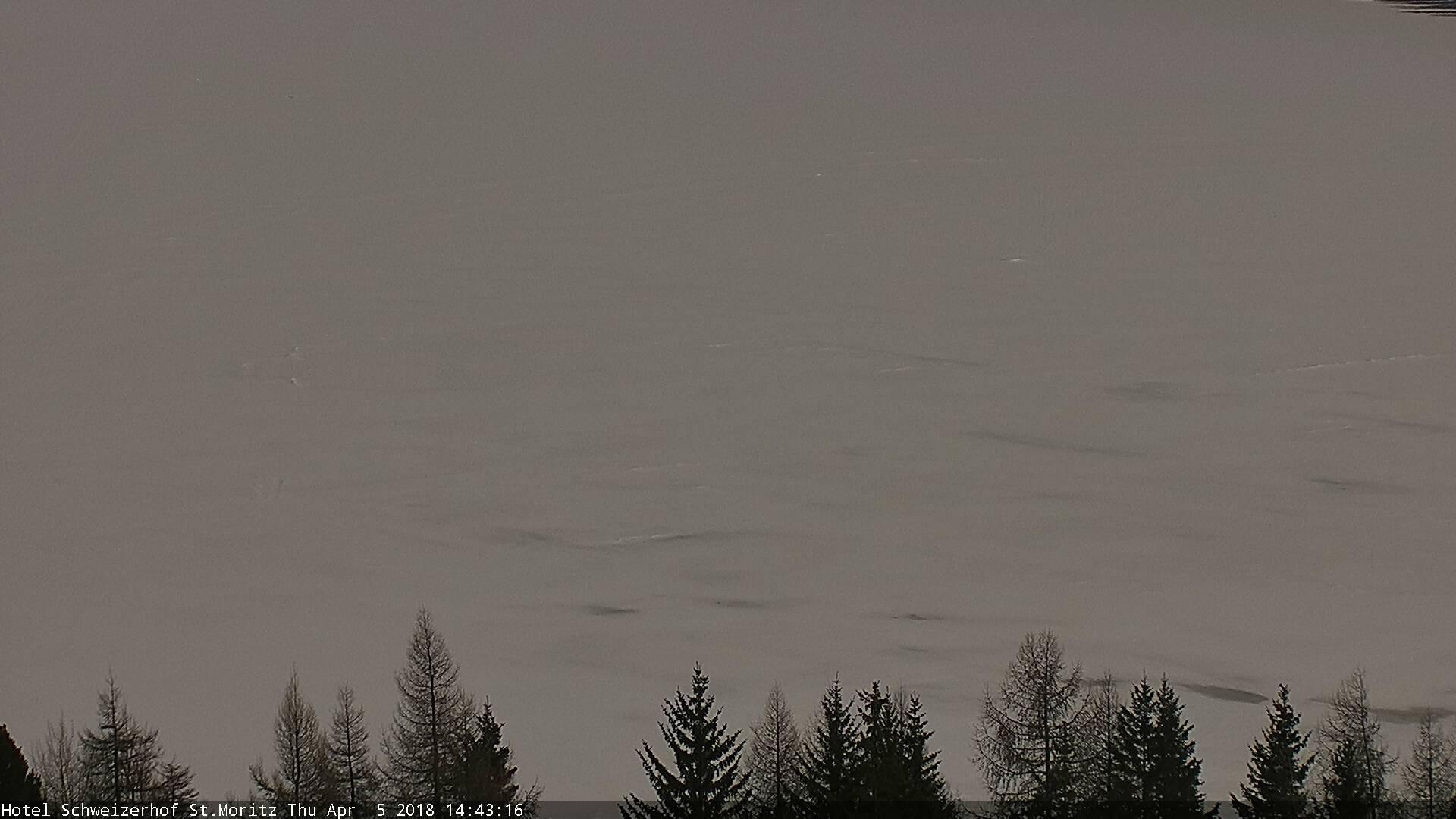}}}\\
  \vspace{-0.65em}
  \subfloat[Camera 2($R_{1}$, w)]{\fbox{\includegraphics[height=2.25cm,width=3.56cm]{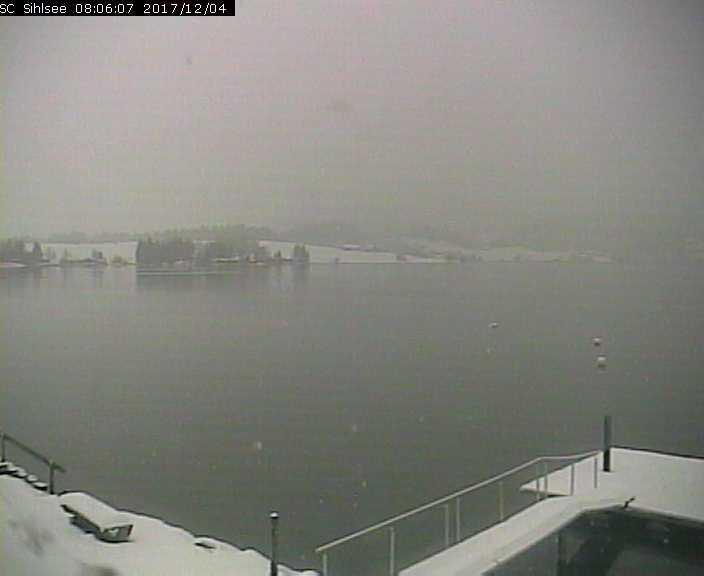}}}\hspace{0.001cm}
  \subfloat[Camera 2($R_{2}$, s)]{\fbox{\includegraphics[height=2.25cm,width=3.56cm]{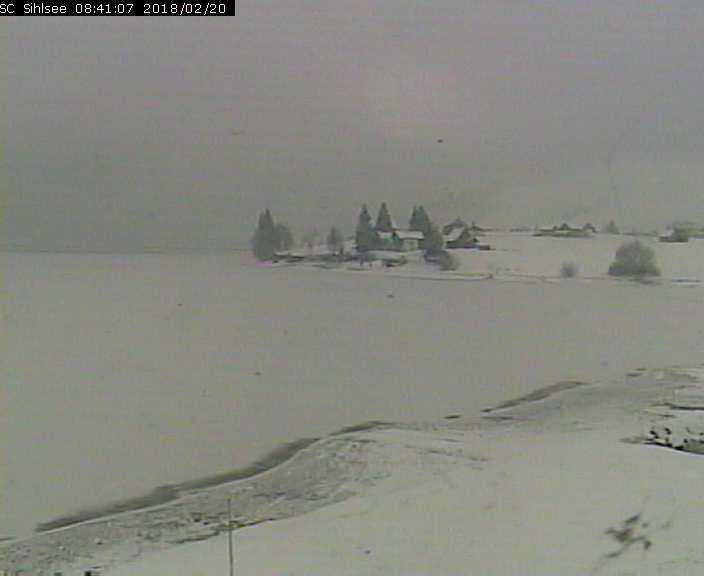}}}\hspace{0.001cm}
  \subfloat[Camera 2($R_{3}$, w)]{\fbox{\includegraphics[height=2.25cm,width=3.56cm]{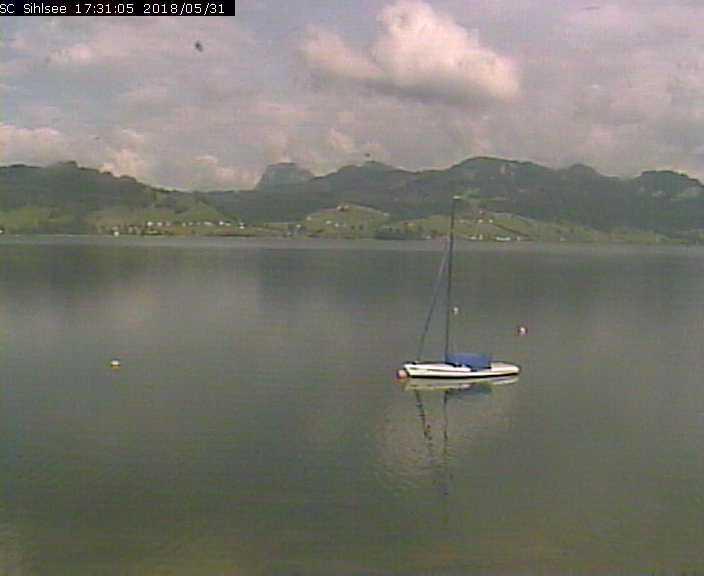}}}\hspace{0.001cm}
  \subfloat[Camera 2($R_{4}$, w)]{\fbox{\includegraphics[height=2.25cm,width=3.56cm]{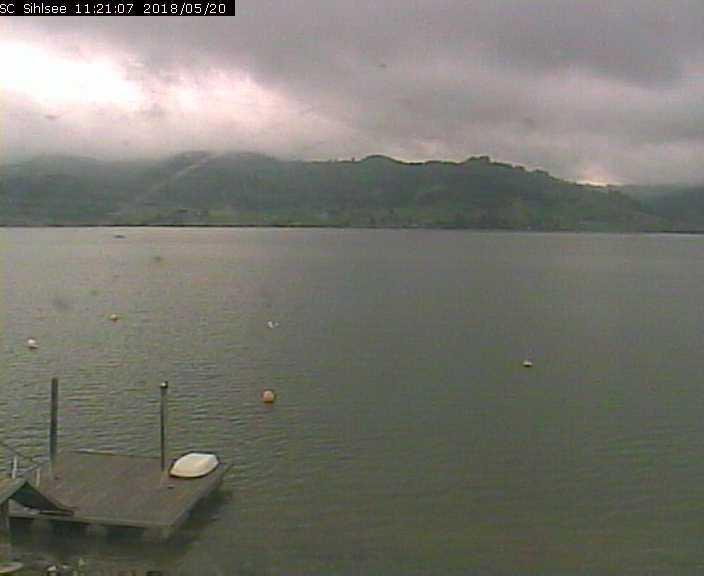}}}\\
  \caption{\textit{Photi-LakeIce} dataset. Rows 1 and 2 display sample images from cameras 0 and 1 (St.~Moritz) respectively. Row 3 shows example images of camera 2 (Sihl, non-stationary, some rotations [$R_{1},R_{2}, etc.$] are also displayed). State of the lake: water(w), ice(i), snow(s), clutter(c) is also displayed in brackets.}
  \label{fig:Photi-lakeice-dataset-snapshots}
\end{figure}
\begin{figure}[t!]
    \centering
        \includegraphics[width=0.7\textwidth]{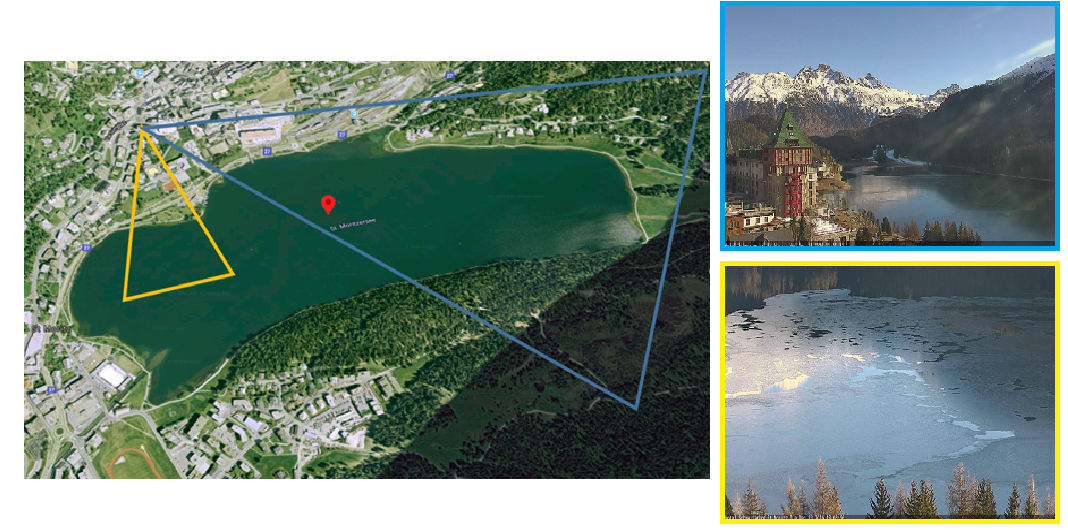} % first figure itself
        \caption{Two webcams monitoring lake St.~Moritz along with their approximate coverage. Image courtesy of Google.}
        \label{fig:2_webcams}
\end{figure}

We make available\footnote[1]{https://github.com/czarmanu/photi-lakeice-dataset} a new webcam dataset, termed \textit{Photi-LakeIce}, for lake ice monitoring and report our results on it. Sample images and details of the dataset are presented in Fig.~\ref{fig:Photi-lakeice-dataset-snapshots} and Table \ref{tab:datasetlakeice} respectively. RGB images (and the corresponding ground truth annotations) from two lakes (Sihl and St.~Moritz) and two winters (2016-17, 2017-18) are included in the dataset. Though the camera mounted at Hotel Schweizerhof in St.~Moritz is rotating, in our analysis we consider it as two different fixed cameras (camera 0 and camera 1, see Fig.~\ref{fig:2_webcams}). The major difference between these two streams is image scale: camera 0 captures images with larger GSD compared to camera 1. Another camera that monitors Sihl is non-stationary, but captures the lake at the same scale (refer Fig.~\ref{fig:Photi-lakeice-dataset-snapshots} row 3). Hence, we consider it as a single rotating camera (camera 2). Our dataset is not limited to but includes images with different lighting conditions (due to sun's angle, time of the year, presence of clouds etc.), shadows (from both clouds and nearby mountains), fog conditions (we remove the extreme cases but keep the images from slightly foggy days), wind scenarios etc.
\begin{table}[h!]
\centering
\small
    \vspace{-0.5em}
    \begin{tabular}{cccccc}
        \hline
     &  &  &  &  &  \\
        \textbf{Name} &\textbf{Lake (Lat, Long)} & \textbf{Camera model} & \textbf{\#images 16-17} & \textbf{\#images 17-18}& \textbf{Res (H$\times$W)}\\
     &  &  &  &  & \\
    Camera 0 & St.\ Moritz (46.50, 9.84)& AXIS Q6128-E & 820 & 474 & 324$\times$1209\\
    Camera 1 & St.\ Moritz (46.50, 9.84)& AXIS Q6128-E & 1180 & 443 & 324$\times$1209\\
    Camera 2 & Sihl (47.13, 8.74)&  unknown & 500 & 600 & 344$\times$420\\
     &  &  &  &  & \\
\hline
  \end{tabular}
  \caption[]{Details of the \textit{Photi-LakeIce} dataset. Lat and Long respectively denote latitude ($^{\circ}$North) and longitude ($^{\circ}$East) of the approximate camera location. Res stands for resolution and H and W represent height and width of the image in pixels (after cropping).}
  \label{tab:datasetlakeice}
\end{table}
\normalsize
\normalsize
\par
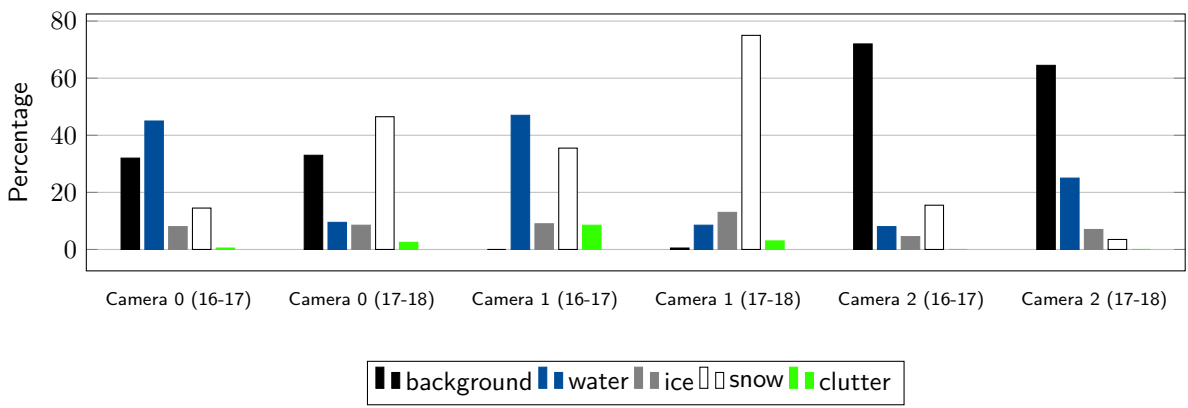
\begin{figure}[h!]
    \begin{tikzpicture}
        \centering
        \begin{axis}
        [
            width  = 1.0\linewidth,
            height = 5cm,
            major x tick style = transparent,
            ybar,
            bar width=7pt,
            ymajorgrids = true,
            ylabel = {Percentage},
            symbolic x coords={Camera 0 (16-17), Camera 0 (17-18), Camera 1 (16-17), Camera 1 (17-18), Camera 2 (16-17), Camera 2 (17-18)},
            xtick = data,
            scaled y ticks = false,
            x tick label style={font=\scriptsize,text width=2.5cm,align=center},
            enlarge x limits=0.1,
            legend style={at={(0.5,-0.350)}, anchor=north,legend columns=-1},
        ]
            \addplot[style={BLACK,fill=BLACK,mark=none}]
                coordinates {(Camera 0 (16-17),32) (Camera 0 (17-18),33) (Camera 1 (16-17),0) (Camera 1 (17-18),0.5) (Camera 2 (16-17),72) (Camera 2 (17-18),64.5) };%  2
            \addplot[style={BLUE2,fill=BLUE2,mark=none}]
                coordinates {(Camera 0 (16-17),45) (Camera 0 (17-18),9.5) (Camera 1 (16-17),47) (Camera 1 (17-18),8.5) (Camera 2 (16-17),8) (Camera 2 (17-18),25) };
            \addplot[style={DARKSILVER,fill=DARKSILVER,mark=none}]
                coordinates {(Camera 0 (16-17),8) (Camera 0 (17-18),8.5) (Camera 1 (16-17),9) (Camera 1 (17-18),13) (Camera 2 (16-17),4.5) (Camera 2 (17-18),7) };
            \addplot[style={BLACK,fill=WHITE,mark=none}]
                coordinates {(Camera 0 (16-17),14.5) (Camera 0 (17-18),46.5) (Camera 1 (16-17),35.5) (Camera 1 (17-18),75) (Camera 2 (16-17),15.5) (Camera 2 (17-18),3.5) };
            \addplot[style={GREEN2,fill=GREEN2,mark=none}]
                coordinates {(Camera 0 (16-17),0.5) (Camera 0 (17-18),2.5) (Camera 1 (16-17),8.5) (Camera 1 (17-18),3) (Camera 2 (16-17),0.0000001) (Camera 2 (17-18),0.0000001) };
            \legend{background, water, ice, snow, clutter}
        \end{axis}
    \end{tikzpicture}
    \caption{Bar graphs displaying class imbalance (including the class \textit{background}) in our dataset. Ice and clutter are the under-represented classes.}
    \label{fig:classimbalance}
\end{figure}

To study the class imbalance in our dataset, we plot the class distribution, individually for each camera and winter, see Fig.~\ref{fig:classimbalance}. It can be inferred that the classes are highly imbalanced in most of the sub-datasets, where \textit{ice} and \textit{clutter} classes suffer the most. In Fig.~\ref{fig:classimbalance}, we show the percentage of class \textit{background} in addition to the four main classes. Note that the percentage of clutter in camera 0 is less compared to camera 1. Note also, camera 1 has almost zero background while the lake area (foreground) to background ratio for Sihl is too low making it a very challenging case. Additionally, the number of ice pixels is consistently low in all the cameras across all years. It will not be surprising if the performance of classes clutter and ice are not good in a relative sense. Note that the background class frequencies differ from one year to another even for the same camera, since in each year the foreground-background separation was done by different human experts. The difference is even more for camera 2 (Sihl), since it is rotating.
\subsubsection{Ground truth generation for webcam analysis}
\begin{figure}[h!]
\footnotesize
  \subfloat[s]{\fbox{\includegraphics[height=1.6cm,width=2.5cm]{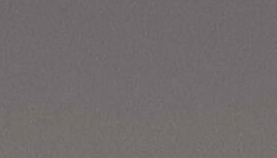}}}\hspace{0.2cm}
  \subfloat[i]{\fbox{\includegraphics[height=1.6cm,width=2.5cm]{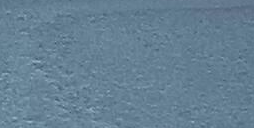}}}\hspace{0.2cm}
  \subfloat[i + w]{\fbox{\includegraphics[height=1.6cm,width=2.5cm]{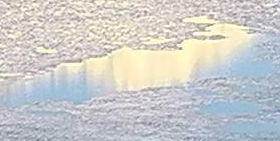}}}\hspace{0.2cm} 
  \subfloat[w]{\fbox{\includegraphics[height=1.6cm,width=2.5cm]{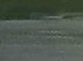}}}\hspace{0.2cm} 
  \subfloat[w]{\fbox{\includegraphics[height=1.6cm,width=2.5cm]{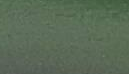}}}
   \caption{Inter-class similarities and intra-class differences of states \textit{snow} (s), \textit{ice} (i), and \textit{water} (w) in our webcam data.}
  \label{fig:2_webcams_challenge}
\end{figure}

The main difficulty in designing a machine (deep) learning system is the requirement for accurately labelled data. However, to generate pixel-wise labels, the interpretation of webcam images is challenging for several reasons. Image quality is limited and off-the-shelf webcams only offer poor radiometric and spectral resolution and are subject to adverse lighting conditions, fog etc.; which makes the image interpretation process difficult even for humans (see Fig.~\ref{fig:2_webcams_challenge}). Besides the limitations of the sensor itself, the cameras are mounted with rather horizontal viewing angle such that large parts of the water body can be observed. As a result, large differences in GSD within a single image are present. Significant intra-class appearance differences exist throughout the image sequences. This is caused by different ice structures, partly frozen water surfaces, waves, varying illumination conditions, reflections and shadows, etc. Furthermore, inter-class appearance similarities exist which impedes automatic interpretation. In fact, even manual interpretation for some examples is impossible without using additional temporal cues.  Pixel-wise ground truth annotations are produced by human operators by labelling polygons within the input images using the \textit{LabelMe} tool \cite{labelme2016}. For the lake detection task, each pixel is either labelled as foreground (lake) or background. Foreground pixels are further annotated as \textit{water, ice, snow} and \textit{clutter} for lake ice segmentation. 
\subsubsection{Ground truth generation for MODIS and VIIRS analysis}
To generate the ground truth for our satellite image analysis, for each day, a human expert visually interpreted the state of a lake (\textit{completely frozen, partly frozen, completely non-frozen, partly non-frozen}) using webcam images of the same. Note also that most of the freely available webcams are not optimally installed to monitor lakes. Hence, besides webcam images, we interpret cloud-free Sentinel-2 images whenever available. Additionally, we attempted to use online media reports to enrich the generated ground truth, which however provided only limited information. In our analysis, webcams have ground truth at a better granularity level (hourly, per pixel label) compared to satellite images (daily, global label). Accurately registering webcam pixels with satellite image pixels is beyond the scope of this work, hence we did not transfer the webcam-based per-pixel ground truth to the satellite images.
\subsection{Methodology}\label{sec:MainBody}
\subsubsection{Satellite image analysis}
Pre-processing of MODIS data (re-sampling to \emph{UTM32N} coordinate system, re-projection) is done using MRTSWATH \cite{MRTSWATH} software. Similarly, VIIRS imagery data is pre-processed (assembling of data granules, re-sampling to \emph{UTM32N}, and mapping) with SatPy \cite{Satpy} package. VIIRS cloud masks are extracted with H5py \cite{H5py} and re-sampled using Pyresample~\cite{PyResample} and GDAL \cite{GDAL} libraries. Among the 12 selected MODIS bands (refer section \ref{sec:optical_sat_data}), the lower resolution bands (500$\,$m and 1000$\,$m GSD) are upsampled to 250$\,$m resolution using bilinear interpolation. This step is not necessary for VIIRS analysis, since we use only the I-bands ($\approx$375$\,$m GSD). For both VIIRS and MODIS, we only analyse the images with at least 30\% cloud-free \textit{clean} pixels. In MODIS images there are also some pixels marked as invalid, which we mask out. For MODIS, we merge the \textit{cloudy} and \textit{uncertain clear} bits to construct a binary cloud-mask from the standard cloud-mask product. Similarly, a VIIRS pixel is considered non-cloudy only if it is either  \textit{confidently clear} or \textit{probably clear}. After Douglas-Peucker generalisation \cite{DouglasPeucker73}, the outlines are further backprojected from the ground coordinate system onto the satellite images to steer the estimation of lake ice. In addition, just the \textit{clean pixels} are analysed, after rectifying the outlines for absolute geolocation shifts, and backprojecting onto VIIRS band $I_{2}$ ($\approx$375$\,$m GSD), respectively MODIS band $B_{2}$ (250$\,$m GSD) as in Tom et al. \cite{tom_lakeice_2018}. For MODIS, the mean offsets in $x$ and $y$ direction were 0.75 and 0.85 pixels, respectively. For VIIRS, the mean offsets were 0 and 0.3 pixels in $x$, respectively $y$ direction.
\par
\begin{figure}[h!]
\centering
	\includegraphics[width=0.55\linewidth]{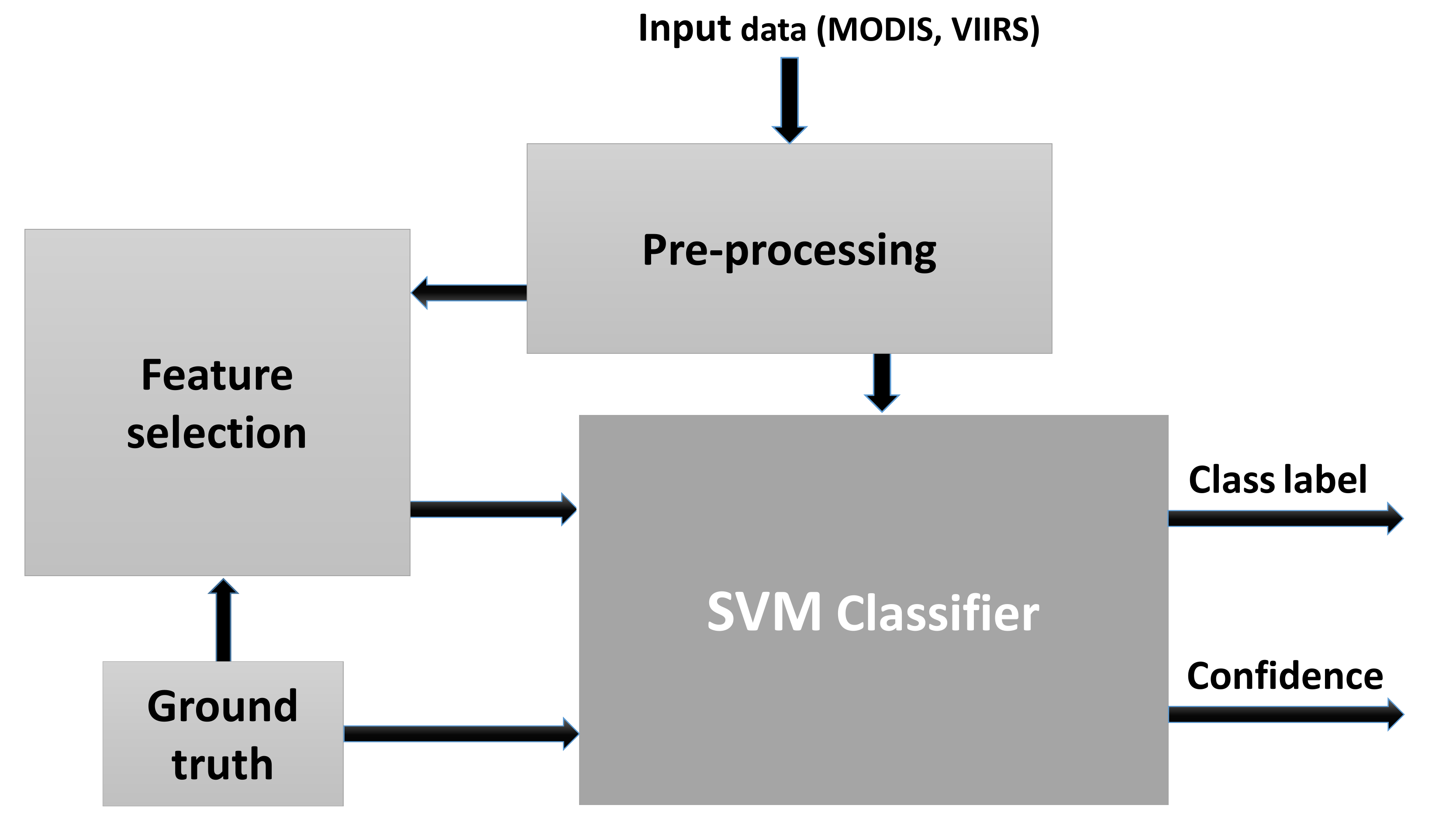}
	\caption{Block diagram of the proposed lake ice detection approach using satellite data.}
	\label{fig:satellites_methodology}
\end{figure}

Fig.~\ref{fig:satellites_methodology} displays the block diagram of the proposed lake ice monitoring system using satellite images. Our semantic segmentation methodology is generic and is applicable to both VIIRS and MODIS imagery. Here, we formulate ice detection as a supervised pixel-wise classification problem (two classes: \textit{frozen} and \textit{non-frozen}). To assess the inter-class separability of different bands, we carry out a supervised variable importance analysis using \textit{XGBoost} feature learning system \cite{xgboost}. The training of that method, a gradient boosting approach based on ensembles of decision trees, makes explicit variable importance conditioned on the class labels. The outcome (F-score) indicates how valuable each feature is in the formation of the boosted (shallow) decision trees within the model. The more a feature (in our case a band) is used to make correct predictions with the decision trees, the higher its relative importance. Though XGBoost is also a classifier by default, we only use the built-in variable selection to automatically determine the most informative bands. For the actual classification based on the selected channels we employ a support vector machine (SVM, \cite{SVM}) classifier, mainly because with SVM it is straight-forward to compare a linear and a non-linear variant. 
\par
\begin{figure}
\centering
\includegraphics[scale=1]{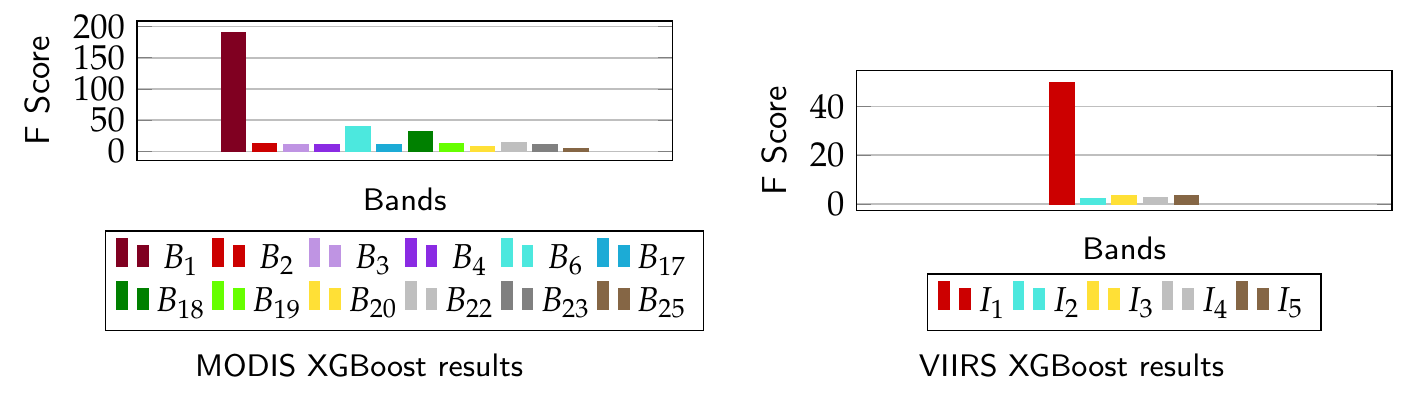}
\caption{Bar graphs for MODIS (left) and VIIRS (right) showing the significance of each of the selected bands (12 for MODIS, 5 for VIIRS) for \textit{frozen} vs. \textit{non-frozen} pixel separation using the \textit{XGBoost} algorithm~\cite{xgboost}. All non-transition days from winter 16-17 are included in the analysis.}
\label{fig:xgboost}
\end{figure}
\begin{figure}
\centering
	\includegraphics[width=0.8\linewidth]{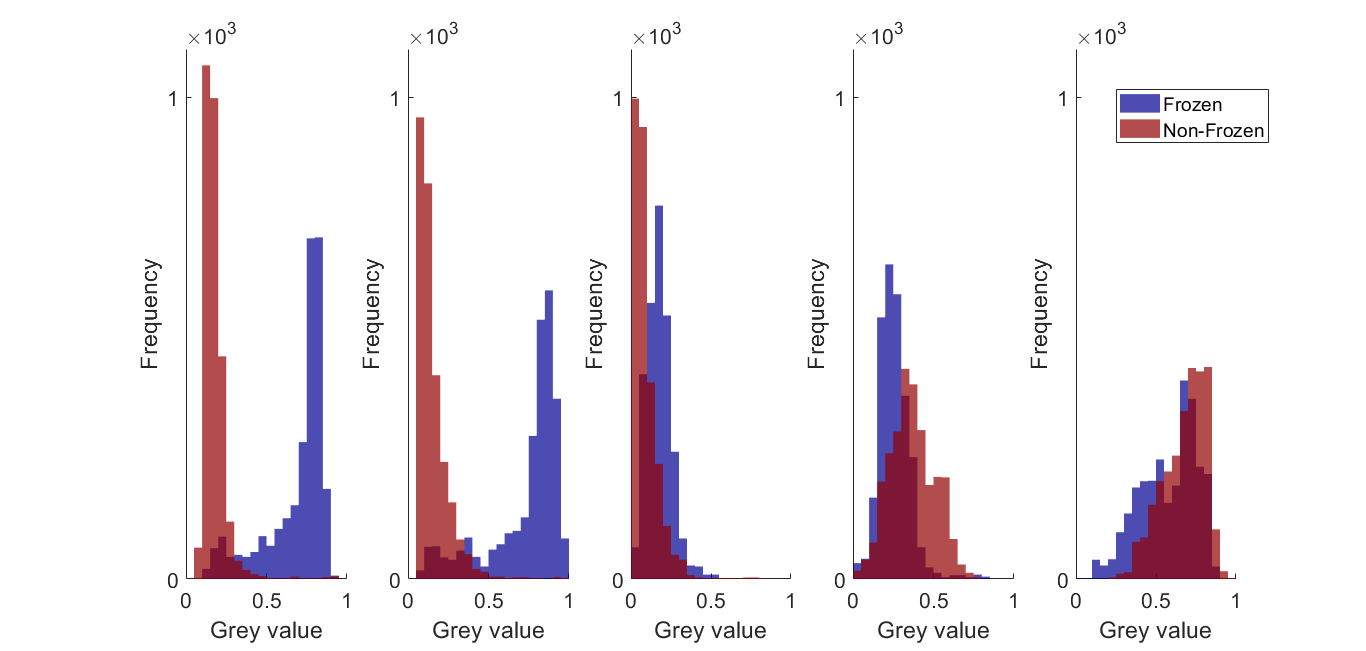}
    \caption{VIIRS grey-value histograms for sanity check (Bands $I_{1}$, $I_{2}$, $I_{3}$, $I_{4}$, $I_{5}$ are respectively shown from left to right).}
	\label{fig:FrovsNotFro_VIIRS}
\end{figure}

The 12 usable MODIS bands and 5 $I$-bands of VIIRS are independently analysed with \textit{XGBoost}. All the data from winter $16$-$17$ (see Table \ref{table:sat_data_stats}) is used to perform this analysis and the results for both MODIS and VIIRS are shown in Fig.~\ref{fig:xgboost} (left and right respectively). Bands $I_{1}$ and $B_{1}$ attain the best scores among the analysed MODIS and VIIRS bands respectively. Furthermore, we plot the grey-value histograms (see Fig.~\ref{fig:FrovsNotFro_VIIRS}) in order to double-check the results generated by \textit{XGBoost}. Due to space limitations, we show only the histograms for VIIRS. Similar histograms for MODIS can be found in Tom et al. \cite{TomACRS2017}. It can be judged from Fig.~\ref{fig:FrovsNotFro_VIIRS} that the two classes are almost similarly separable in the two near-infrared bands $I_{1}$ and $I_{2}$. Since those two bands have similar spectrum and are highly correlated, XGBoost picks only one among them. The same holds for the two near-infrared MODIS bands $B_{1}$ and $B_{2}$.
\par
It is very likely that the physical state of a lake pixel is the same on subsequent days (except during the highly dynamic freezing and thawing periods). Hence, as a post-processing step, multi-temporal analysis (MTA) is applied. For each pixel, a moving average of the SVM class scores is computed along the time dimension. The average is computed within a fixed window length (smoothing parameter) that is determined empirically. Choosing the smoothing parameter is critical, as too high values can easily wash out the critical dynamics during the transition days. Since the pixels from each MODIS (or VIIRS) acquisition are predicted independently by the trained SVM model, MTA is expected to improve the SVM results by leveraging on the temporal relationships. We test three different averaging schemes: \textit{mean, median} and \textit{Gaussian}. 
\subsubsection{Webcam image analysis}
\begin{figure}[ht]
  \centering
  \subfloat{\includegraphics[height=5cm]{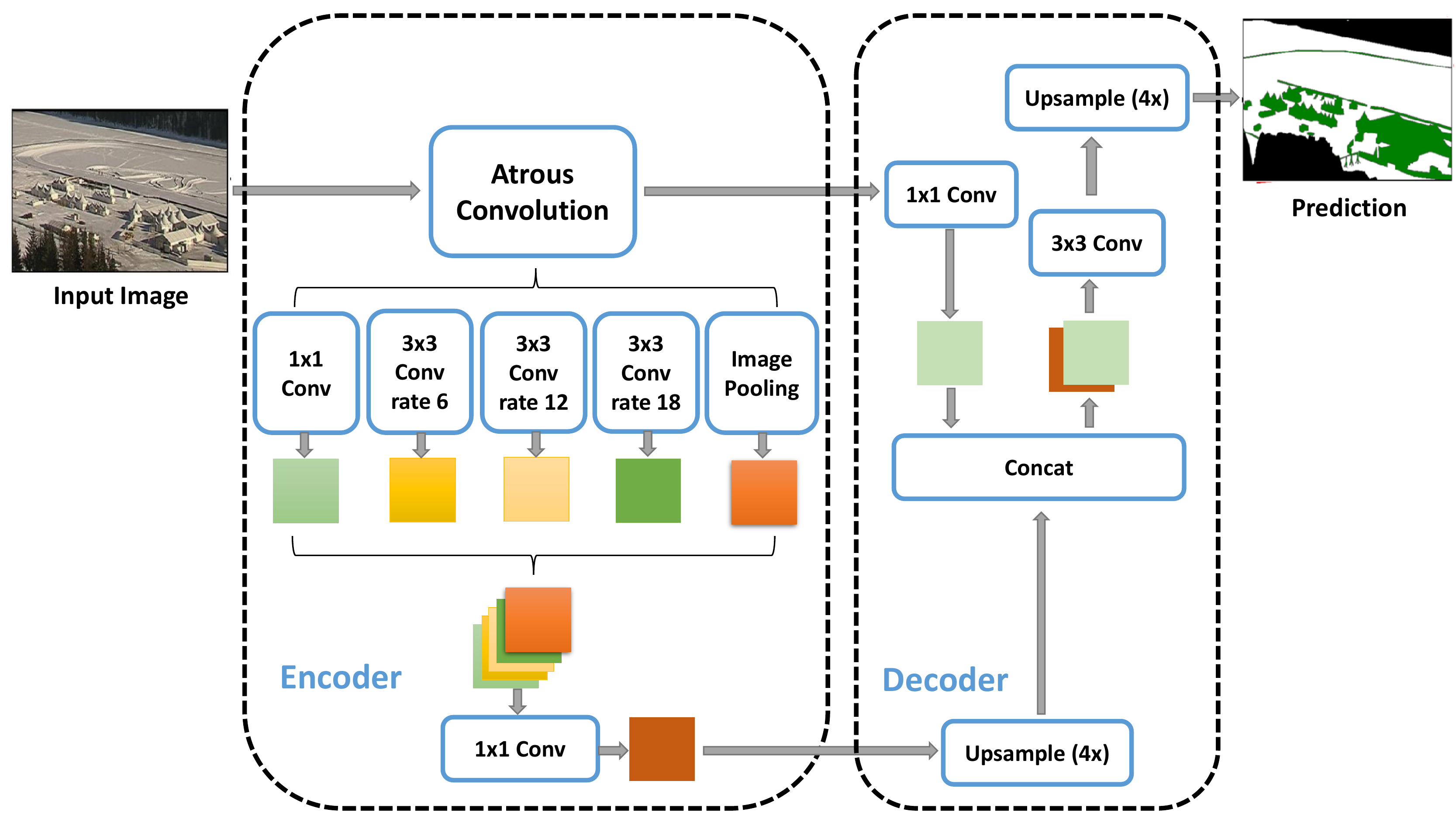}}\hspace{2em}
  \subfloat{\includegraphics[height=5cm]{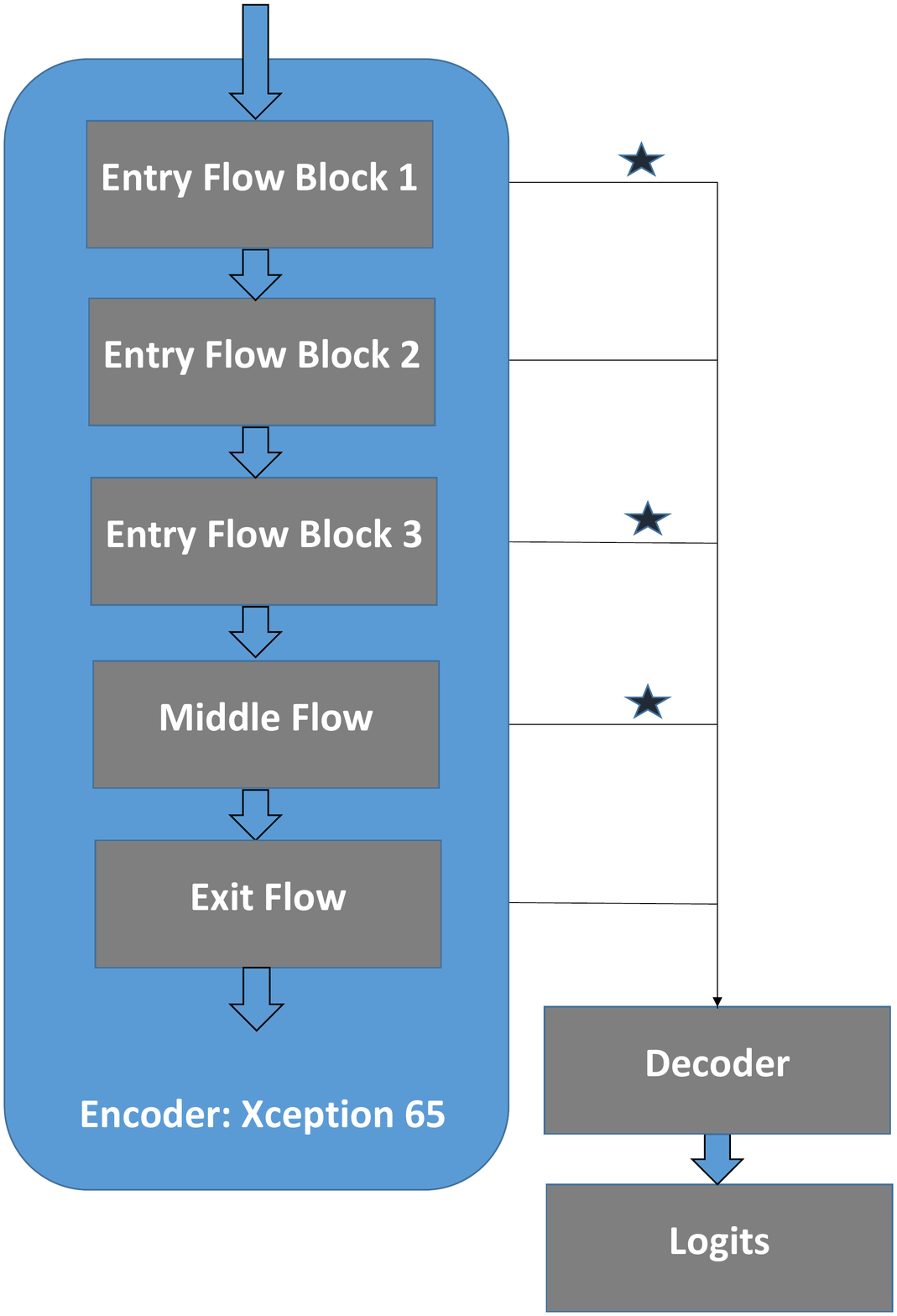}}
  \caption{\textit{Deeplab v3+} (left) and \textit{Deep-U-Lab} (right) architectures. The "$\star$" symbol indicates the additional skip connections for \textit{Deep-U-Lab}.}
  \label{fig:deeplab}
\end{figure}

Similar to satellite analysis, we formulate our webcam approach as supervised semantic segmentation problem. Here, we make use of the prominent high-performance deep learning architecture, \textit{Deeplab v3+}~\cite{deeplabv3plus2018}, with
the \textit{Xception65} encoder branch, see Fig.~\ref{fig:deeplab} (left). That method has a proven track record on other semantic segmentation benchmarks such as PASCAL VOC \citep{Everingham15} and Cityscapes \citep{Cordts_2016_CVPR}. Our network classifies each pixel in the RGB webcam image as \emph{water}, \emph{ice}, \emph{snow} or \emph{clutter}. The clutter class incorporates man-made objects on the lake, eg., structures built for sport events (such as tents in St.~ Moritz), boats, etc. Note that, as for satellite images, lake ice segmentation is done only for foreground (lake) pixels. 
\par
By integrating the spatial pyramid pooling technique as well as atrous convolution into the standard encoder-decoder architecture, the \textit{Deeplab v3+} network encodes rich contextual information at arbitrary scales and  retrieves segment boundaries more precisely. Spatial convolution was applied independently to each channel, followed by 1$\times$1 (point-wise) convolution to combine the per-channel outputs. This markedly reduces the computational complexity without any noticeable performance drop. Where needed, these depthwise separable convolutions employ stride 2 in the spatial component, making separate pooling operations obsolete. Note that the atrous (dilated) convolution effectively increases the receptive field without blurring the signal. Using multiple atrous rates makes sure that features are extracted at various spatial scales.
\par
Inspired from \textit{U-net} \cite{RonnebergerFB15}, and with an aim to sharpen the class boundaries, in addition to the single skip connection used in \textit{Deeplab v3+} per default, we introduce three more from different flow blocks (entry and mid-level flow) of the \textit{Xception65} encoder to the decoder. We call this new variant \textit{Deep-U-Lab}, see Fig.~\ref{fig:deeplab} (right). The new feature maps thus generated are concatenated along with the existing ones, for better preservation of high frequency information at the class boundaries. 
\par
In order to deal with the shortage (in deep learning terms) of labelled data, we make use of transfer-learning that allows to re-use knowledge gained from other similar tasks. To do so, we employ a model pre-trained on the PASCAL VOC benchmark dataset (for both lake detection and ice segmentation tasks) rather than starting from scratch. We fine-tune the weights in all layers since tuning just the final layer did not work, emphasising the fact that even low-level texture characteristics differ between our webcam images and the PASCAL dataset and must be adapted.%
\par
In previous works \cite{muyan_lakeice_2018,LIP1_final_report_2019}, the lake pixels were manually delineated before inferring their class. Locating the lake pixels  makes the job easier for the classifier as it does not have to deal with the spectral appearance outside of the lake. We propose to automate this step, in order to make lake ice observation more practical in operational scenarios. Automated lake detection is very useful especially when scaling up the webcam network to also include non-stationary cameras. Hence, we formulate lake detection as a two-class (\textit{background, foreground}) segmentation problem and train yet another instance of our \textit{Deep-U-Lab} model. Then, a fine-grained classifier predicts the state (\textit{water, ice, snow, clutter}) of the lake pixels.
\par
With the intention to minimise overfitting of the model, we perform data augmentation, i.e., we add more, synthetically generated variations to the training dataset. We do this by applying random rotations, zooming, up-down and left-right flipping.
\section{Results}\label{sec:results}
For our evaluation we use the following error metrics: recall, precision, overall accuracy, Intersection over Union (IoU, a.k.a.~Jaccard index), and mean IoU across classes (mIoU).
\subsection{Experiments on satellite images}
\label{experiments:satellites}
Unless explicitly specified, the experimental settings are the same (but independent) for both MODIS and VIIRS. We divide the time series into two parts: \textit{transition} and \textit{non-transition} dates. During the partially frozen transition days, ground truth annotation is very challenging, as one has to discriminate thin transparent ice from water. Hence, we report the quantitative results only on the non-transition dates. However, qualitative analysis is done on all the available dates. Furthermore, with VIIRS we process only 3 lakes (Sihl, Sils, Silvaplana) since there exists no clean pixel for lake St.~Moritz (see Table \ref{table:dataset_stats}).
\subsubsection{4-fold cross validation}
As a first experiment, we combine the data of all the lakes from two winters (16-17 and 17-18), and perform \textit{4-fold cross validation} in order to figure out the optimum SVM parameter settings for detailed experimentation. We did a grid-search for the two main hyperparamaters of SVM (the \textit{cost} of a misclassification and the kernel width \textit{gamma}) and found that, for both sensors, the best results with Radial Basis Function (RBF) kernel are obtained with values 10 and 1 for cost and gamma, respectively. With linear kernel, value 0.1 as cost works best for both MODIS and VIIRS. We notice that classification of our dataset using the RBF kernel is fairly sensitive to the selection of hyperparameters, while the linear kernel provides consistent results. Note also, optimum hyperparameters might vary from one dataset to another. Quantitative results of 4-fold cross validation experiments with the optimum parameters are displayed in Table \ref{table:quant_results_VIIRS_MODIS}. For MODIS we obtain the best results when all 12 bands are used as feature vector (and for VIIRS with all 5 bands). In addition, our results show that the performance of RBF kernel is a bit better compared to the linear counterpart. Additionally, we test variants that use fewer bands, down to a single band with highest F-score as selected by XGBoost ($B_{1}$ for MODIS and $I_{1}$ for VIIRS, refer Fig.~\ref{fig:xgboost}). Since it does not make sense to run an RBF kernel with a single input band, we try only the linear kernel for this experiment. Even with a single band and a simple linear kernel the results are fairly decent. The results are even better when using the top-5 bands of MODIS, but slightly worse than the full 12-band feature vector. Multi-temporal analysis (MTA) improves the results by a very small margin. For both MODIS and VIIRS, MTA with Gaussian kernel (smoothing parameter 3) gives the best results and is therefore used in all further experiments. For the best setting, we show the results in more detail in Table \ref{table:VIIRS_CV_2years_bestcase}.
\begin{table}[ht!]
\small
	    \centering
	    \vspace{-0.5em}
	    \begin{tabular}{ |cccccc| }
		\hline
        \textbf{Sensor}    &\textbf{Feature vector}     &  \textbf{SVM kernel} & \textbf{with MTA} & \textbf{Overall accuracy} & \textbf{mIoU}\\ 
		 &  &  &  &  & \\ 
		MODIS & $B_{1}$ & Linear & No & 0.91 & 0.78\\ 
		MODIS & Top 5 bands & Linear & No & 0.93 & 0.83\\ 
		MODIS & All 12 bands & Linear & No & 0.93 & 0.84\\ 
		MODIS & All 12 bands & Linear & Yes & 0.93 & 0.84 \\ 
		MODIS & Top 5 bands & RBF & No & 0.96 & 0.90\\ 
		MODIS & All 12 bands & RBF & No & \textbf{0.99}	& 0.98\\ 
		MODIS & All 12 bands  & RBF & Yes & \textbf{0.99} & \textbf{0.99} \\
		 &  &  &  &  & \\ 
		VIIRS & $I_{1}$ & Linear & No & 0.93 & 0.84 \\ 
		VIIRS & All 5 I-bands & Linear & No & 0.95 & 0.88 \\ 
		VIIRS & All 5 I-bands & Linear & Yes & 0.95 & 0.88  \\ 
		VIIRS & All 5 I-bands & RBF & No & \textbf{0.97} & \textbf{0.93}\\ 
		VIIRS & All 5 I-bands & RBF & Yes & \textbf{0.97} & \textbf{0.93}\\ 
		\hline
	    \end{tabular}
	    \caption{4-fold cross validation results on MODIS and VIIRS data from two winters (16-17 and 17-18). For the same SVM setup, results without and with multi-temporal analysis (MTA) are shown.}
	    \label{table:quant_results_VIIRS_MODIS}
	\end{table}
\normalsize
\begin{table}[!ht]
    \centering
    \small
    \vspace{-0.5em}
    \begin{minipage}{.49\linewidth}
      \centering
        \begin{tabular}{|cccc|}\hline
             & \textbf{Recall }  & \textbf{Precision } & \textbf{IoU}\\ 
            Frozen & 0.99 & 0.99 & 0.98\\
            Non-Frozen & 0.99 & 0.99 & 0.99\\
            \textbf{Accuracy / mIoU} & & 0.99~~/~~0.99 & \\ \hline
        \end{tabular}
    \end{minipage} 
    \begin{minipage}{.49\linewidth}
      \centering
        \begin{tabular}{|cccc|}\hline
             & \textbf{Recall }  & \textbf{Precision } & \textbf{IoU}\\ 
            Frozen & 0.93 & 0.97 & 0.90\\
            Non-Frozen & 0.99 & 0.97 & 0.96\\
            \textbf{Accuracy / mIoU} & & 0.97~~/~~0.93 & \\ \hline
        \end{tabular}
    \end{minipage}
    \caption{Detailed results on MODIS (left) and VIIRS (right) data for the best cases of 4-fold cross validation on combined data from two winters.}
    \label{table:VIIRS_CV_2years_bestcase}
\end{table}
\normalsize
\par
We note that feature selection may be beneficial especially with very small training sets. Ideally, SVM automatically prioritises the more important dimensions in the feature vector, but when only few examples are available, the danger of spurious correlations in less discriminative bands increases. For lake ice detection, where few channels carry most of the information, we recommend to use automatic feature selection in case the SVM overfits.
\par
For a practically useful and efficient learning-based monitoring framework, a model should be trained using annotated data from a handful of lakes as well as a few winters, but should be able to predict for lakes and winters not seen during training. To test the performance of our approach in such scenarios, we perform the \textit{leave one lake out}, respectively \textit{leave one winter out} cross validation experiments. In all the following  experiments, we use all the 12 (5) bands of MODIS (VIIRS), optimum hyperparameters chosen by grid-search (cost 10 and gamma 1 for RBF kernel, cost 0.1 for linear kernel), and MTA with Gaussian kernel (smoothing parameter 3).
\subsubsection{Leave one lake out cross validation}
This experiment evaluates the across-lake generalisation capability of the classifier. We use the SVM model trained on pixels of all but one lakes (from both winters) and test on the pixels from the remaining lake, in round-robin mode. MODIS and VIIRS results are presented in Tables~\ref{table:MODIS_LOLO} and \ref{table:VIIRS_LOLO}, respectively. As per the results, our models fair well even when trained using only pixels from different lakes. Using both RBF and linear kernels, our algorithm gives excellent results on lakes Sils and Silvaplana consistently with both VIIRS and MODIS data. Table \ref{table:target_lakes} shows that both these lakes are similar in many aspects. It is expected for a learning-based system that predictions are better if the test data is more similar to the training data. The performance of RBF kernel on lakes St.~Moritz and Sihl is also good, but not as good as Sils and Silvaplana. Recall that St.~Moritz has just 4 clean pixels per MODIS acquisition (see Table \ref{table:dataset_stats}) and that the absolute geolocation accuracy could be critical for such a small lake. It appears that for St.~Moritz the linear kernel does not generalise unlike other lakes, but we do not draw any firm conclusions based on these results, as the lake is too small. Lake Sihl is slightly different compared to the other lakes (altitude, area etc., refer Table \ref{table:target_lakes}) and hence the SVM encounters more generalisation loss. Still a mean IoU of 0.78 (corresponding to 93\% correctly classified pixels) for MODIS, respectively 0.85 (95\%) for VIIRS is a rather good result. For both sensors, the performance of the linear kernel on Sihl is better compared to RBF. Given the fact that Sihl has a somewhat different geography than the other lakes, it appears that the linear kernel generalises better. 
\begin{table}[!ht]
    \centering
    \small
    \vspace{-0.5em}
    \begin{minipage}{.49\linewidth}
      \centering
        \begin{tabular}{|cccc|}\hline
        \multicolumn{4}{|c|}{\textbf{Lake Sihl}}\\
        \hdashline
              & \textbf{Recall }  & \textbf{Precision } & \textbf{IoU}\\ 
            Frozen & 0.82/\textcolor{gray}{0.79} & \textcolor{gray}{0.63}/0.78 & \textcolor{gray}{0.55}/0.65\\
            Non-Frozen & \textcolor{gray}{0.90}/0.95 & 0.96/0.96 & \textcolor{gray}{0.87}/0.92\\
            \textbf{Accuracy } & & \textcolor{gray}{0.89}/0.93 & \\
            \textbf{mIoU} & & \textcolor{gray}{0.71}/0.78 & \\ \hline
        \end{tabular}
    \end{minipage} 
    \begin{minipage}{.49\linewidth}
      \centering
        \begin{tabular}{|cccc|}\hline
        \multicolumn{4}{|c|}{\textbf{Lake Sils}}\\
        \hdashline
                 & \textbf{Recall }  & \textbf{Precision } & \textbf{IoU}\\ 
            Frozen & 0.89/\textcolor{gray}{0.88} & 0.97/\textcolor{gray}{0.95} & 0.86/\textcolor{gray}{0.85}\\
            Non-Frozen & 0.98/\textcolor{gray}{0.97} & 0.92/0.92 & 0.90/\textcolor{gray}{0.89}\\
            \textbf{Accuracy } & & 0.94/\textcolor{gray}{0.93} & \\ 
            \textbf{mIoU} & & 0.88/\textcolor{gray}{0.87} & \\ \hline
        \end{tabular}
    \end{minipage} \\
    \vspace{1em}
    \begin{minipage}{.49\linewidth}
      \centering
        \begin{tabular}{|cccc|}\hline
        \multicolumn{4}{|c|}{\textbf{Lake Silvaplana}}\\
        \hdashline
             & \textbf{Recall }  & \textbf{Precision } & \textbf{IoU}\\ 
            Frozen & 0.91/\textcolor{gray}{0.81} & \textcolor{gray}{0.96}/0.97 & 0.88/\textcolor{gray}{0.79}\\
            Non-Frozen & \textcolor{gray}{0.97}/0.98 & 0.93/\textcolor{gray}{0.86} & 0.90/\textcolor{gray}{0.85}\\
            \textbf{Accuracy } & & 0.94/\textcolor{gray}{0.91} & \\
            \textbf{mIoU} & & 0.89/\textcolor{gray}{0.82} & \\ \hline
        \end{tabular}
    \end{minipage}
    \begin{minipage}{.49\linewidth}
      \centering
        \begin{tabular}{|cccc|}\hline
        \multicolumn{4}{|c|}{\textbf{Lake St.~Moritz}}\\
        \hdashline
             & \textbf{Recall }  & \textbf{Precision } & \textbf{IoU}\\ 
            Frozen & 0.85/\textcolor{gray}{0.64} & \textcolor{gray}{0.93}/0.96 & 0.80/\textcolor{gray}{0.63}\\
            Non-Frozen & \textcolor{gray}{0.95}/0.98 & 0.88/\textcolor{gray}{0.76} & 0.84/\textcolor{gray}{0.75}\\
            \textbf{Accuracy} & & 0.90/\textcolor{gray}{0.83} & \\
            \textbf{mIoU} & & 0.82/\textcolor{gray}{0.69} & \\ \hline
        \end{tabular}
    \end{minipage} 
    \caption{MODIS \textit{leave one lake out} results. Numbers are in A/B format where A and B represent the results using RBF and linear kernels, respectively. Better kernel for a given experiment is shown in black, inferior kernel in \textcolor{gray}{grey}.}
    \label{table:MODIS_LOLO}
\end{table}
\normalsize
\begin{table}[!ht]
    \centering
    \small
    \vspace{-0.5em}
    \begin{minipage}{.49\linewidth}
      \centering
        \begin{tabular}{|cccc|}\hline
        \multicolumn{4}{|c|}{\textbf{Lake Sihl}}\\
        \hdashline
         & \textbf{Recall }  & \textbf{Precision } & \textbf{IoU}\\ 
            Frozen & 0.87/0.87 & \textcolor{gray}{0.73}/0.85 & \textcolor{gray}{0.66}/0.76\\
            Non-Frozen & \textcolor{gray}{0.92}/0.97 & 0.97/0.97 & \textcolor{gray}{0.90}/0.94\\
            \textbf{Accuracy } & & \textcolor{gray}{0.91}/0.95 & \\ 
            \textbf{mIoU} & & \textcolor{gray}{0.78}/0.85 & \\ \hline
        \end{tabular}
    \end{minipage} 
    \begin{minipage}{.49\linewidth}
      \centering
        \begin{tabular}{|cccc|}\hline
        \multicolumn{4}{|c|}{\textbf{Lake Sils}}\\
        \hdashline
         & \textbf{Recall }  & \textbf{Precision } & \textbf{IoU}\\ 
            Frozen & 0.93/\textcolor{gray}{0.89} & \textcolor{gray}{0.97}/0.99 & 0.90/\textcolor{gray}{0.88}\\
            Non-Frozen & \textcolor{gray}{0.98}/0.99 & 0.94/\textcolor{gray}{0.91} & 0.92/\textcolor{gray}{0.90}\\
            \textbf{Accuracy } & & 0.95/\textcolor{gray}{0.94} & \\
            \textbf{mIoU} & & 0.91/\textcolor{gray}{0.89} & \\ \hline
        \end{tabular}
    \end{minipage}\\
    \vspace{1em}
    \begin{minipage}{.49\linewidth}
      \centering
        \begin{tabular}{|cccc|}\hline
        \multicolumn{4}{|c|}{\textbf{Lake Silvaplana}}\\
        \hdashline
         & \textbf{Recall }  & \textbf{Precision } & \textbf{IoU}\\ 
            Frozen & 0.91/\textcolor{gray}{0.87} & \textcolor{gray}{0.97}/0.98 & 0.88/\textcolor{gray}{0.86}\\
            Non-Frozen & \textcolor{gray}{0.97}/0.98 & 0.92/\textcolor{gray}{0.89} & 0.90/\textcolor{gray}{0.88}\\
            \textbf{Accuracy } & & 0.94/\textcolor{gray}{0.93} & \\
            \textbf{mIoU} & & 0.90/\textcolor{gray}{0.87} & \\ \hline
        \end{tabular}
    \end{minipage}
    \caption{VIIRS \textit{leave one lake out} results. Numbers are in A/B format where A and B represent the results using RBF and linear kernels, respectively. Better kernel for a given experiment is shown in black, inferior kernel in \textcolor{gray}{grey}.}
    \label{table:VIIRS_LOLO}
\end{table}
\normalsize
\subsubsection{Leave one winter out cross validation}
To investigate the adaptability of a model to the potentially different conditions of an unseen winter, we train the classifier using pixels from one of the two available winters (from all lakes), and test on the data from the other winter. The results for MODIS and VIIRS are shown in Tables \ref{table:MODIS_LOWO} and \ref{table:VIIRS_LOWO}, respectively. Comparing these results with Table \ref{table:quant_results_VIIRS_MODIS}, it can be inferred that, across winters, the SVM does encounter a generalisation loss, especially with the RBF kernel. The loss with the linear kernel is minimal. Apparently, the RBF overfitted to the data characteristics of the specific winter and did not generalise as well as its linear counterpart. Note also, it is possible that freezing patterns could vary across winters even for the same lake, and learning-based systems might fail in case a pattern appears while testing that was not encountered during training. It is encouraging that the linear kernel does not seem to overfit much, owing to its relatively lower capacity. Still, the results hint that annotated data from more than one winter should be present in the training set when setting up an operational system.
\begin{table}[h]
    \centering
    \small
    \vspace{-0.5em}
    \begin{minipage}{.49\linewidth}
      \centering
        \begin{tabular}{|cccc|}\hline
             & \textbf{Recall }  & \textbf{Precision } & \textbf{IoU}\\ 
            Frozen & \textcolor{gray}{0.72}/0.77 & \textcolor{gray}{0.90}/0.91 & \textcolor{gray}{0.67}/0.72\\
            Non-Frozen & \textcolor{gray}{0.96}/0.97 & \textcolor{gray}{0.89}/0.91 & \textcolor{gray}{0.86}/0.88\\
            \textbf{Accuracy } & & \textcolor{gray}{0.89}/0.91 & \\
            \textbf{mIoU} & & \textcolor{gray}{0.76}/0.80 & \\ \hline
        \end{tabular}
    \end{minipage} 
    \begin{minipage}{.49\linewidth}
      \centering
        \begin{tabular}{|cccc|}\hline
             & \textbf{Recall }  & \textbf{Precision } & \textbf{IoU}\\ 
            Frozen & \textcolor{gray}{0.73}/0.84 & \textcolor{gray}{0.64}/0.85 & \textcolor{gray}{0.52}/0.73\\
            Non-Frozen & \textcolor{gray}{0.89}/0.96 & \textcolor{gray}{0.93}/0.96 & \textcolor{gray}{0.83}/0.92\\
            \textbf{Accuracy } & & \textcolor{gray}{0.86}/0.94 & \\
            \textbf{mIoU} & & \textcolor{gray}{0.68}/0.83 & \\ \hline
        \end{tabular}
    \end{minipage}
    \caption{MODIS \textit{leave one winter out} results. The numbers are shown in A/B format where A and B represent the outcomes using RBF and linear kernels, respectively. Better kernel for a given experiment is shown in black, inferior kernel in \textcolor{gray}{grey}. Left is winter 16-17, right is winter 17-18.}
    \label{table:MODIS_LOWO}
    \vspace{-0.75em}
\end{table}
\normalsize
\begin{table}[h]
    \centering
    \small
    \vspace{-0.5em}
    \begin{minipage}{.49\linewidth}
      \centering
        \begin{tabular}{|cccc|}\hline
             & \textbf{Recall }  & \textbf{Precision } & \textbf{IoU}\\ 
            Frozen & 0.83/\textcolor{gray}{0.79} & \textcolor{gray}{0.92}/0.99 & \textcolor{gray}{0.77}/0.78\\
            Non-Frozen & \textcolor{gray}{0.96}/0.99 & 0.91/\textcolor{gray}{0.89} & \textcolor{gray}{0.88}/0.89\\
            \textbf{Accuracy } & & \textcolor{gray}{0.91}/0.92 & \\ 
            \textbf{mIoU} & & \textcolor{gray}{0.82}/0.84 & \\ \hline
        \end{tabular}
    \end{minipage} 
    \begin{minipage}{.49\linewidth}
      \centering
        \begin{tabular}{|cccc|}\hline
             & \textbf{Recall }  & \textbf{Precision } & \textbf{IoU}\\ 
            Frozen & \textcolor{gray}{0.87}/0.90 & \textcolor{gray}{0.77}/0.79 & \textcolor{gray}{0.69}/0.72\\
            Non-Frozen & \textcolor{gray}{0.93}/0.94 & 0.97/0.97 & \textcolor{gray}{0.90}/0.91\\
            \textbf{Accuracy } & & \textcolor{gray}{0.92}/0.93 & \\ 
            \textbf{mIoU} & & \textcolor{gray}{0.79}/0.82 & \\ \hline
        \end{tabular}
    \end{minipage}
    \caption{VIIRS \textit{leave one winter out} results. The numbers are shown in A/B format where A and B represent the outcomes using RBF and linear kernels respectively. Better kernel for a given experiment is shown in black, inferior kernel in \textcolor{gray}{grey}. Left is winter 16-17, right is winter 17-18.}
    \label{table:VIIRS_LOWO}
\end{table}
\normalsize
\subsubsection{Timeline plots and qualitative results}
\begin{figure}[!h]
    \centering
    \begin{tabular}{@{}c@{}c@{}c@{}c@{}c@{}c@{}}%
        \includegraphics[width=0.99\linewidth]{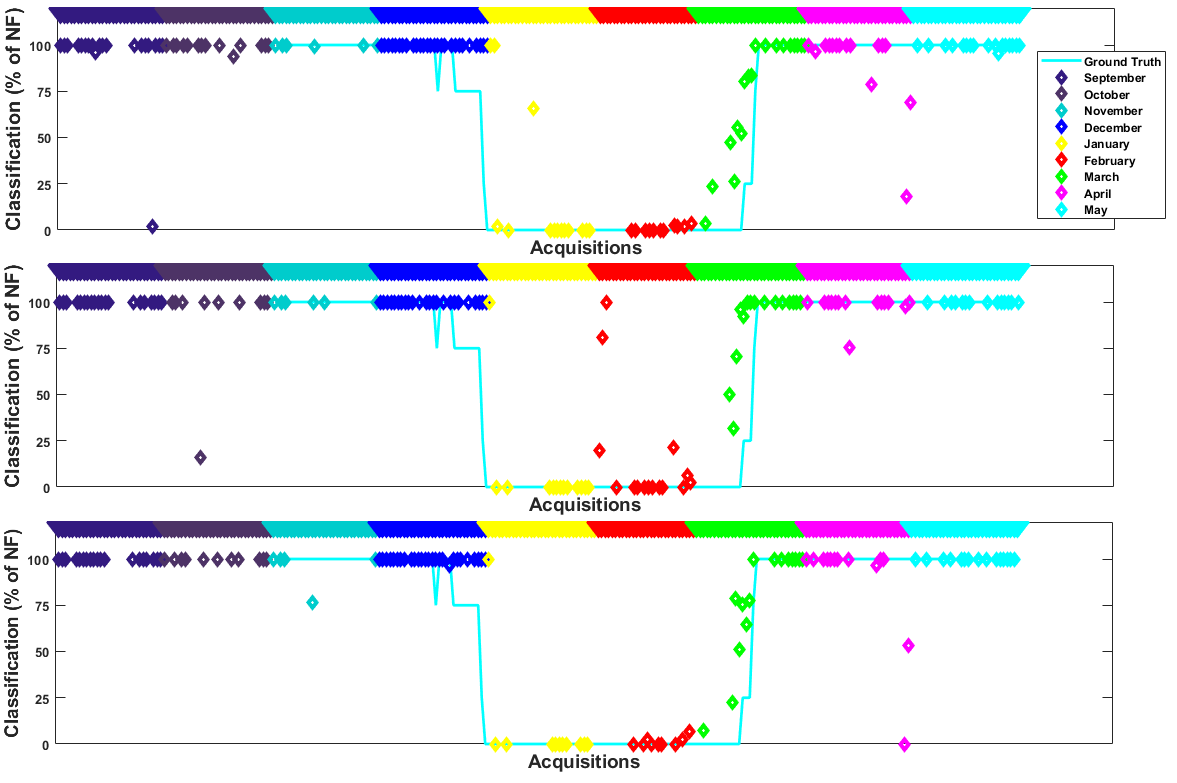}
    \end{tabular}
    \vspace{-1.5mm}%
    \caption{MODIS (top) and VIIRS (middle) timeline results of lake Sihl for full winter 16-17 using linear kernel. A timeline of the MODIS snow and ice product (bottom) is also plotted for comparison with our results and the webcam-based ground truth. In all the timelines, the $x$-axis shows all the dates that are at least 30\% cloud-free in chronological order and the respective results [\% of Non-Frozen (NF) pixels] are plotted on the $y$-axis.} 
    \label{fig:sihl_MODIS_VIIRS_timeline}
\end{figure}

Fig.~\ref{fig:sihl_MODIS_VIIRS_timeline} shows the results of lake Sihl from a full winter (September 2016 till May 2017), listed in chronological order on the $x$-axis. For each cloud-free day (at least 30\% of the lake pixels non-cloudy), the SVM result is shown on the $y$-axis (in the top and middle timelines) as the percentage of cloud-free clean pixels that are classified as non-frozen. In the bottom timeline, we display the MODIS snow product (100 means no snow and 0 means fully snow covered). The webcam-based ground truth is shown as a cyan colour line in all timeline plots, with four levels (100 for fully non-frozen, 75 for more snow or more ice days, 25 for more water and 0 for fully-frozen). For each sensor, the combined training data of all available lakes from two winters (except Sihl from 16-17) is used for these timeline plots. It can be seen from both MODIS and VIIRS timelines that \textit{thin ice} vs.\ \textit{water} confusion exists for both MODIS and VIIRS. This is because during the freeze-up period (late December) the model classifies a set of consecutive days as completely non-frozen, while the ground truth asserts more ice, probably thin ice floating on water.
\par
In this paper, we compare our results of lake Sihl from winter 2016-17 with the MODIS snow product \cite{snowmap}. It can be inferred from Fig.~\ref{fig:sihl_MODIS_VIIRS_timeline} that except for very few days, our MODIS results are in agreement with the MODIS snow product. Although the newly added MOD10A1F \cite{snowmap6.1} (collection 6.1) seems to be a better option with the \textit{'cloud gap filled'} feature, we use the MOD10A1 product \cite{snowmap} (collection 6, daily cloud-free snow cover derived from the Terra MODIS) since the former product is not yet available \cite{NSIDC} for winter 2016-17. Note that the MODIS product has a relatively coarse spatial resolution of 500$\,$m as opposed to our results at 250$\,$m resolution.
\par
\begin{figure}[h]
\centering
  \subfloat[03.01.2017]{%
      \includegraphics[width=0.24\textwidth]{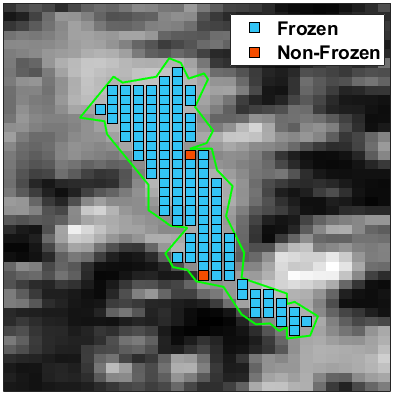}}\hspace{0.01em}
  \subfloat[28.03.2017]{%
      \includegraphics[width=0.24\textwidth]{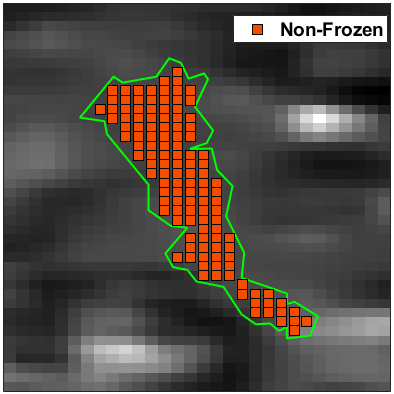}}\hspace{0.01em}
  \subfloat[14.03.2017]{%
      \includegraphics[width=0.24\textwidth]{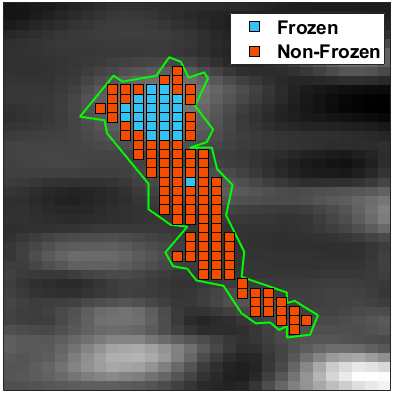}}\hspace{0.01em}
  \subfloat[27.09.2016]{%
      \includegraphics[width=0.24\textwidth]{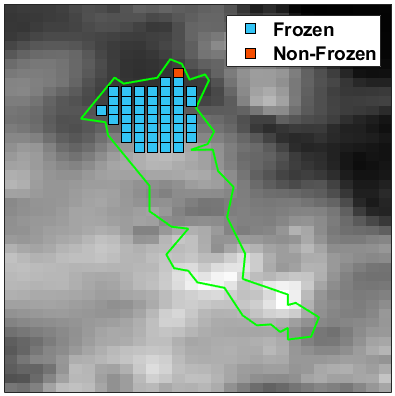}}\\
      \vspace{-0.75em}
  \subfloat[Frozen]{%
      \includegraphics[width=0.24\textwidth]{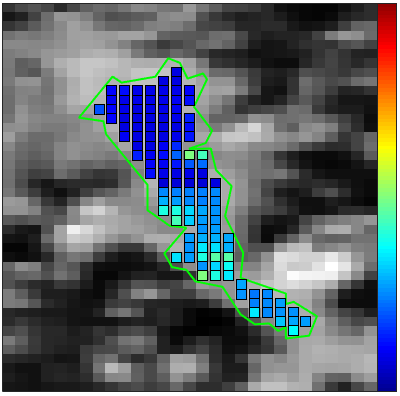}}\hspace{0.01em}
  \subfloat[Non-Frozen]{%
      \includegraphics[width=0.24\textwidth]{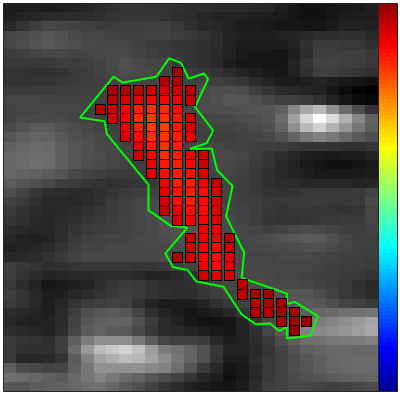}}\hspace{0.01em}
  \subfloat[Transition]{%
      \includegraphics[width=0.24\textwidth]{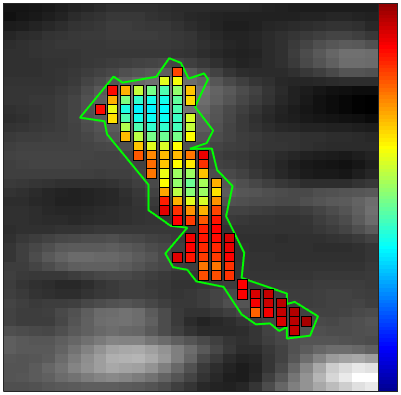}}\hspace{0.01em}
  \subfloat[Non-Frozen]{%
      \includegraphics[width=0.24\textwidth]{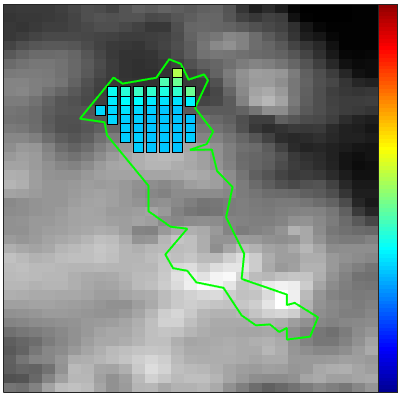}}\\
  \caption{MODIS qualitative results using the linear kernel. Top and bottom rows show classification results and corresponding confidence respectively. Results of cloudy pixels are not displayed. First, second, and third columns show success cases while the fourth column displays a failure case. In second row, the more red means more non-frozen and more blue means more frozen. The dates and ground truth labels are shown below each sub-figure in the first and second rows respectively.}
  \label{fig:VIIRS_success cases}
\end{figure}

Fig.~\ref{fig:VIIRS_success cases} displays exemplary qualitative results (lake Sihl, MODIS data, winter 16-17). Three non-transition dates (27.09.2016, 03.01.2017 and 28.03.2017) and a transition date (14.03.2017) are shown. The first and second rows portray the classification results and the confidence of the classifier (soft probability maps) respectively. In row 1, a clean pixel is shown as blue if the classifier estimates it as frozen, and red if non-frozen. In the second row, more blue / less red colour denotes a higher probability of being frozen. A pixel is not processed if it is cloudy. All except the fourth column show successfully classified days. In column 4, we present the results of an actually fully non-frozen day (27.09.2016) that was detected as almost fully frozen. Note the missing cloudy pixels in this image. This example shows that erroneous cloud masks (especially the false negatives) also induce errors in our predictions. Similar effects can be observed for end of April (MODIS) and early October (VIIRS). Confusion due to undetected clouds is also the reason why a few days were estimated as non-frozen during mid-winter (see VIIRS timeline, February). 
\subsection{Experiments on webcam images}
The neural network is implemented in \textit{Tensorflow}~\cite{Tflow}. We extract square patches (\textit{crop size}, see Table \ref{tab:parameters}) from the images and train the network by minimising the weighted cross-entropy loss, giving more attention to the under-represented classes to compensate imbalances in the training data. Testing is performed at full image resolution without cropping. More details about the hyper-parameters used are shown in Table \ref{tab:parameters}.%
\begin{table}[h]
  \caption[Parameters]{Hyperparameters for the \textit{Deep-U-Lab} model.}\label{tab:parameters}
  \centering
  \begin{tabular}{l l l}
    \toprule
      \textbf{Name} & \textbf{Lake detection} & \textbf{Lake ice segmentation} \\
    \midrule
      Crop size & 500, 500 & 321, 321\\
      Optimizer & stochastic gradient descent & stochastic gradient descent\\
      Atrous rates (dilation) & 6, 12, 18 & 6, 12, 18 \\
      Output stride & 16 & 16 (training), 8 (testing)\\
      Base learning rate & 1e-5 & 1e-5 \\
      Batch size & 4 & 8\\
      Epochs & 100 & 100\\
    \bottomrule
  \end{tabular}
\end{table}
\par
\begin{figure}[h!]
 \centering
  \subfloat[St.~Moritz (46.50$^{\circ}$N, 9.84$^{\circ}$E)]{\fbox{\includegraphics[width=0.33\textwidth,height =3cm]{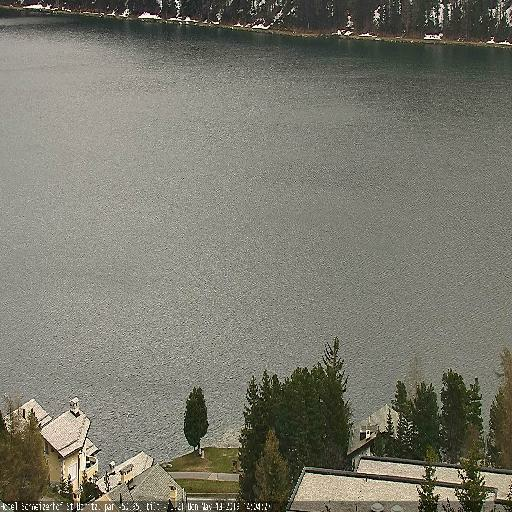}}}\hspace{0.01em}
  \subfloat[Our prediction]{\fbox{\includegraphics[width=0.3\textwidth,height =3cm]{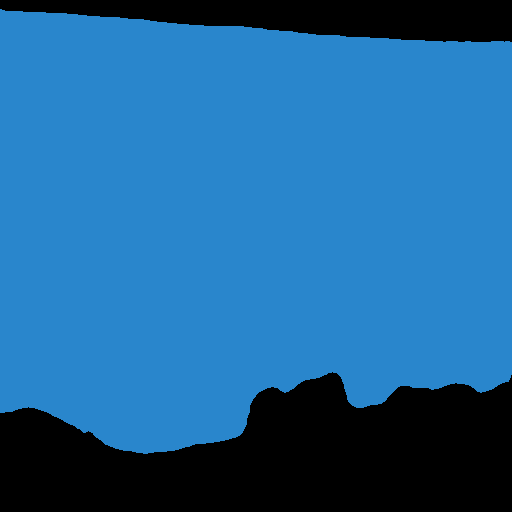}}}\hspace{0.01em}
  \subfloat[Ground truth]{\fbox{\includegraphics[width=0.3\textwidth,height =3cm]{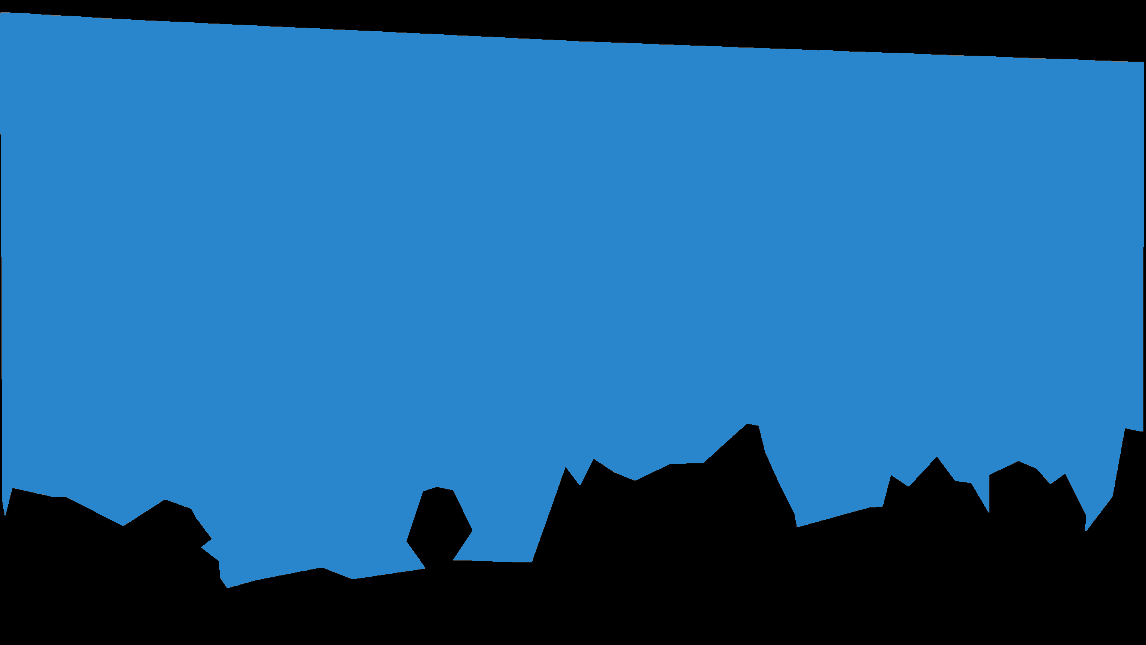}}}
  \vspace{-1.05em}
  \\
  \subfloat[Sihl (47.13$^{\circ}$N, 8.74$^{\circ}$E)]{\fbox{\includegraphics[width=0.33\textwidth,height =3cm]{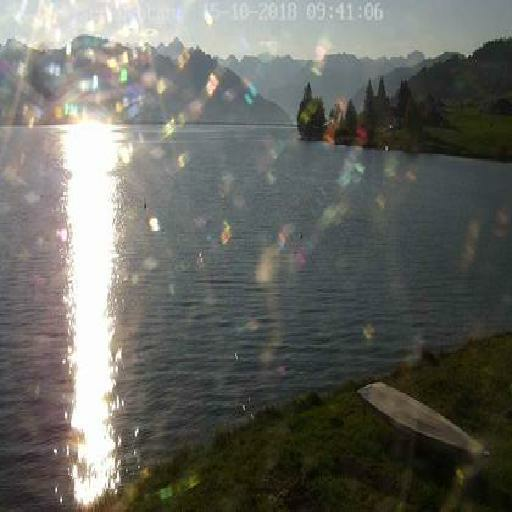}}}\hspace{0.01em}
  \subfloat[Our prediction]{\fbox{\includegraphics[width=0.3\textwidth,height =3cm]{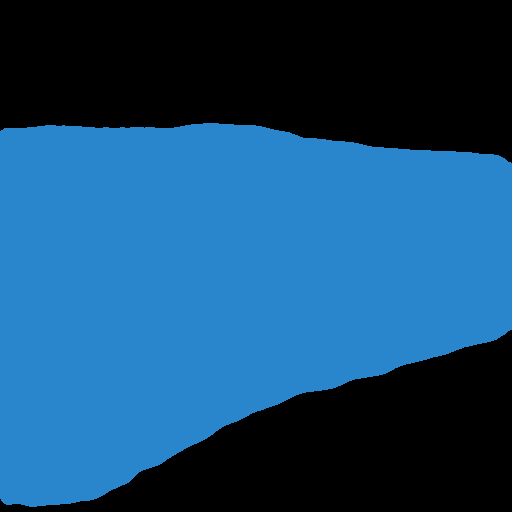}}}\hspace{0.01em}
  \subfloat[Ground truth]{\fbox{\includegraphics[width=0.3\textwidth,height =3cm]{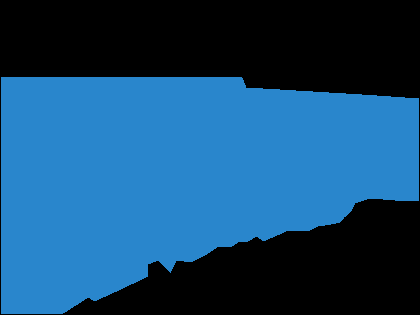}}}
  \vspace{-1.05em}
  \\
  \subfloat[Silvaplana (46.51$^{\circ}$N, 9.81$^{\circ}$E)]{\fbox{\includegraphics[width=0.33\textwidth,height =3cm]{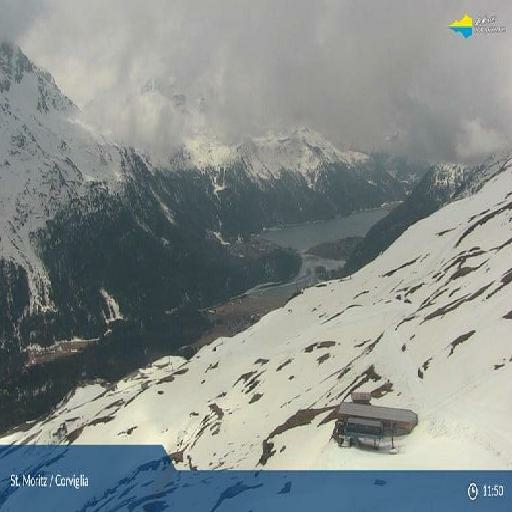}}}\hspace{0.01em}
  \subfloat[Our prediction]{\fbox{\includegraphics[width=0.3\textwidth,height =3cm]{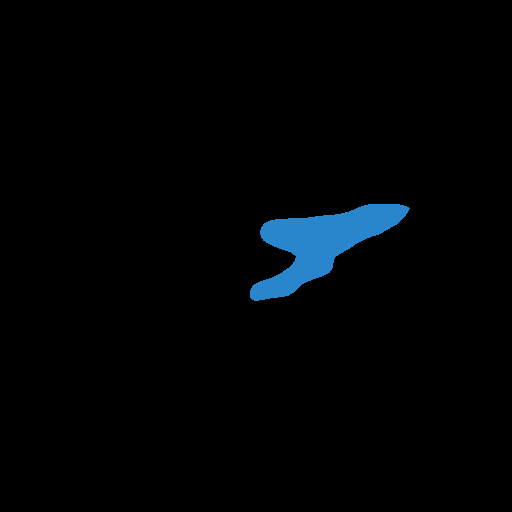}}}\hspace{0.01em}
  \subfloat[Ground truth]{\fbox{\includegraphics[width=0.3\textwidth,height =3cm]{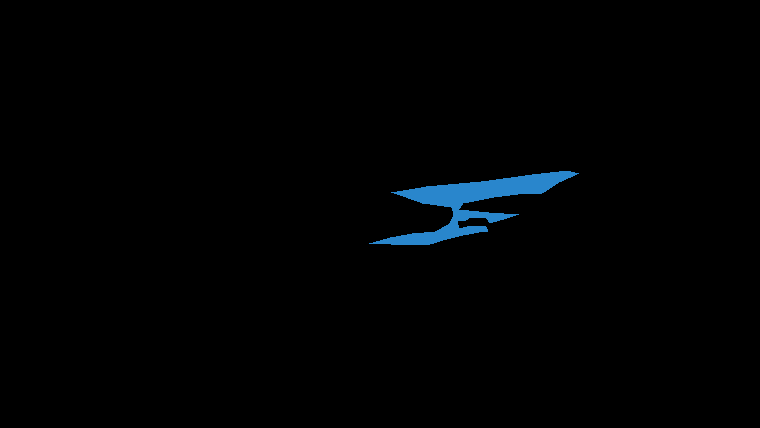}}}
  \vspace{-1.05em}
  \\
  \subfloat[Sihl (47.13$^{\circ}$N, 8.74$^{\circ}$E)]{\fbox{\includegraphics[width=0.33\textwidth,height =3cm]{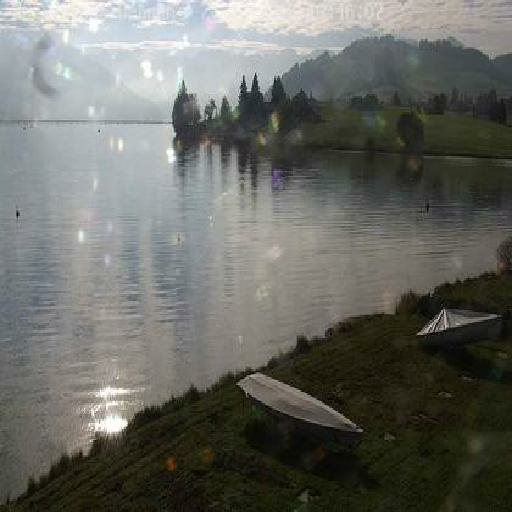}}}\hspace{0.01em}
  \subfloat[Our prediction]{\fbox{\includegraphics[width=0.3\textwidth,height =3cm]{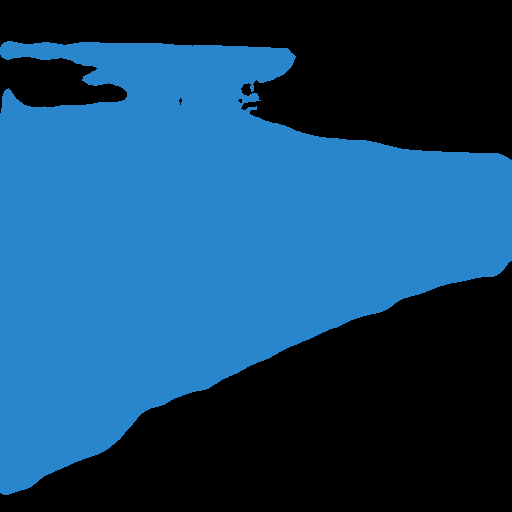}}}\hspace{0.01em}
  \subfloat[Ground truth]{\fbox{\includegraphics[width=0.3\textwidth,height =3cm]{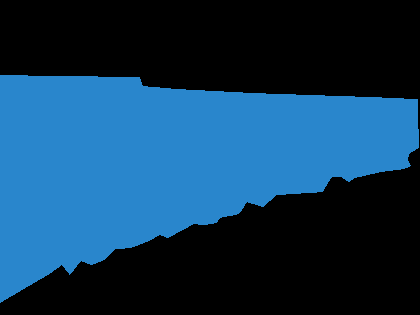}}}\\
  \subfloat[Colour code]{\includegraphics[width=0.2\columnwidth]{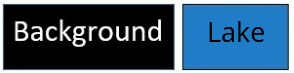}}
  \caption{Lake detection results. Both success (rows 1,2,3) and failure (row 4) cases are shown. The colour code used to visualise the results is also displayed. The first column shows the lakes being monitored, along with the approximate location (latitude, longitude) of the webcam.}
  \label{fig:lakedetection}
\end{figure}

For the task of lake detection, we have collected image streams from four different lakes, see Table~\ref{tab:lakedetection}. The cameras near lakes Sihl and St.~Moritz are rotating while the others are stationary. Performance of the network for lake detection is assessed only on summer images in order to sidestep the complications in winter due to the presence of snow in and around the lake. We achieve $\geq$0.9 mIoU score, see Table~\ref{tab:lakedetection}. However, the IoU for the lake (class foreground, FG) is somewhat lower because of the severe class imbalance. Note that IoU is a rather strict measure, e.g., detection with 80\% recall at 80\% precision results in an IoU of 60\%. Qualitative results are displayed in Fig.~\ref{fig:lakedetection}. The first three rows show successful cases while the last row displays a case with some misclassification. Note that on this low visibility day even humans find it difficult to spot the transition from lake to sky. Whereas our network detected the lake even in the presence of challenging sun reflections (row 2) and when the foreground lake area is very small (row 3).
\begin{table}[h]
\small
    \centering
    \vspace{-0.5em}
        \begin{tabular}{lllllll}\hline
        \multicolumn{2}{c}{\textbf{Training set}} & \multicolumn{2}{c}{\textbf{Test set}} & \multirow{2}{*}{\textbf{IoU (BG)}}& \multirow{2}{*}{\textbf{IoU (FG)}}& \multirow{2}{*}{\textbf{mIoU}}\\
    \textbf{Lakes}&\textbf{\#Images}&\textbf{Lake}&\textbf{\#Images}& \\
    \hline
    &  &  &  &  &  & \\
    $S_{0}$, $S_{1}$, $S_{2}$, $S_{3}$, Sils, St.~Moritz&9104 & Sihl & 448 & 0.95 & 0.60 & 0.93\\
    $S_{0}$, $S_{1}$, $S_{2}$, $S_{3}$, St.~Moritz, Sihl& 7477 & Sils & 2075 & 0.95 & 0.60 & 0.93\\
    $S_{1}$, $S_{2}$, $S_{3}$, Sils, St.~Moritz, Sihl&8041& $S_{0}$ & 1511 & 0.96 & 0.59 & 0.94\\
    $S_{0}$, $S_{2}$, $S_{3}$, Sils, St.~Moritz, Sihl&8676& $S_{1}$ & 876 & 0.92 & 0.58 & 0.90\\
    $S_{0}$, $S_{1}$, $S_{3}$, Sils, St.~Moritz, Sihl&7906& $S_{2}$ & 1646 & 0.98 & 0.44 & 0.95\\
    $S_{0}$, $S_{1}$, $S_{2}$, Sils, St.~Moritz, Sihl&7652& $S_{3}$ & 1900 & 0.98 & 0.55 & 0.95\\
    $S_{0}$, $S_{1}$, $S_{2}$, $S_{3}$, Sils, Sihl& 8456& St.~Moritz & 1096& 0.93 & 0.80 & 0.92\\
    &  &  &  &  &  & \\
    \hline
\end{tabular}
\caption{Lake detection results (mIoU). The four cameras that monitor lake Silvaplana are indicated as $S_{0}$, $S_{1}$, $S_{2}$, and $S_{3}$. BG and FG denote background and foreground (lake area) respectively.}
\label{tab:lakedetection}
\end{table}
\normalsize
\par
To assess lake ice segmentation, we experiment exhaustively for the two lakes (St.~Moritz, Sihl) and two winters (16-17, 17-18) annotated in the \textit{Photi-LakeIce} dataset. The evaluation includes experiments for segmentation within the \textit{same camera}, \textit{across cameras}, \textit{across winters}, and \textit{across lakes}. 
\par
For \textit{same camera} experiments, we employ a 75\%/25\% train-test split. Corresponding quantitative results are presented in Table \ref{tab:webcams_quant:same_camera}. Note that, in all comparable experiments, we surpass previous state-of-the-art~\cite{LIP1_final_report_2019} by a significant margin. They produced results only on two cameras monitoring St.~Moritz. We demonstrate our system also on a new lake (Sihl, camera 2) with images that are significantly harder to classify (see Fig.~\ref{fig:Photi-lakeice-dataset-snapshots} and Table \ref{tab:datasetlakeice}) due to poor spatial resolution, image compression artefacts, frequent unfavourable lighting, etc. Additionally, the foreground to background pixel ratios in Sihl images are very low, which poses an additional challenge, and magnifies the influence of very small misclassified areas on the quantitative error metrics. As a result, the performance on lake Sihl is not as good as for St.~Moritz. Nevertheless, the predictions have a mean IoU $>$74\%. The images with a mix of classes, like water with some ice or partially snow-covered ice are the most difficult ones to classify; in part due to the fact that especially the ice class is rare, and therefore under-represented in the training data, as snow that falls on the ice does not melt away for a long time. 
\begin{table}[ht!]
\small
\centering
\begin{tabular}{ccccccccc}\hline
   \multicolumn{2}{c}{\textbf{Training set}} & \multicolumn{2}{c}{\textbf{Test set}} & \multirow{2}{*}{\textbf{Water}} & \multirow{2}{*}{\textbf{Ice}} & \multirow{2}{*}{\textbf{Snow}} & \multirow{2}{*}{\textbf{Clutter}} & \multirow{2}{*}{\textbf{mIoU}}\\
   \textbf{Camera} & \textbf{Winter} & \textbf{Camera} & \textbf{Winter} & &  &  &  &  \\\hline
       &  &  &  &  &  &  &  &  \\
    Camera 0 & 16-17 & Camera 0 & 16-17 & 0.98 / \textcolor{gray}{0.70} & 0.95 / \textcolor{gray}{0.87} & 0.95 / \textcolor{gray}{0.89} & 0.97 / \textcolor{gray}{0.63} & 0.96 / \textcolor{gray}{0.77} \\
   Camera 0 & 17-18 & Camera 0 & 17-18 & 0.97 & 0.88 & 0.96 & 0.87 & 0.93 \\
   Camera 1 & 16-17 & Camera 1 & 16-17 & 0.99 / \textcolor{gray}{0.90} & 0.96 / \textcolor{gray}{0.92} & 0.95 / \textcolor{gray}{0.94} & 0.79 / \textcolor{gray}{0.62} & 0.92 / \textcolor{gray}{0.85} \\
   Camera 1 & 17-18 & Camera 1 & 17-18 & 0.93 & 0.84 & 0.92 & 0.84 & 0.89 \\
  Camera 2 & 16-17 & Camera 2 & 16-17 & 0.79 & 0.62 & 0.81 & --- & 0.74 \\
   Camera 2 & 17-18 & Camera 2 & 17-18 & 0.81 & 0.69 & 0.86 & --- & 0.79 \\
    &  &  &  &  &  &  &  &  \\
   \hline
   \end{tabular}
   \caption{Results (IoU) of \textit{same camera train test} experiments. We compare our results with \textit{Tiramisu Network}~\cite{LIP1_final_report_2019} (shown in \textcolor{gray}{grey}). Cameras 0 and 1 monitor lake St.~Moritz while camera 2 captures lake Sihl.}
\label{tab:webcams_quant:same_camera}
\end{table}
\normalsize
\par
\begin{figure}[h]
\centering
    \includegraphics[width=0.65\columnwidth]{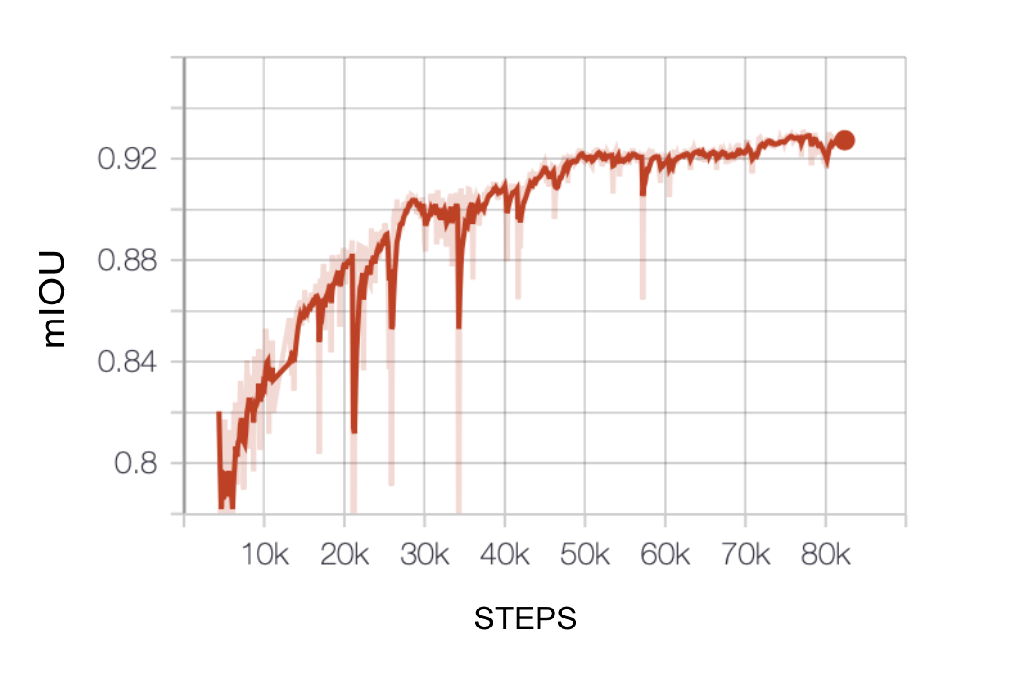}
   \caption{Evolution of mIoU against the number of training steps (camera 0, St.~Moritz, winter 2016-17).}
  \label{fig:accuracy_vs_steps}
\end{figure}

 
All results shown so far are for networks trained with data augmentation. To quantify the influence of this common practice, we also report results without augmentation for camera 0, which are 2 percent points lower, see Table \ref{tab:moritzaugment}.
\begin{table}[h]
  \centering
    \begin{tabular}{cccccc}
        \toprule
        \textbf{Experiment} & \textbf{Water} & \textbf{Ice} & \textbf{Snow} & \textbf{Clutter} & \textbf{mIoU} \\
        \midrule
   Without augmentation & 0.97 & 0.93 & 0.91 & 0.96 & 0.94 \\
   With augmentation & 0.98 & 0.95 & 0.95 & 0.97 & 0.96 \\
\bottomrule
\end{tabular}
\caption{Effect of data augmentation (IoU values) on same camera experiment (camera 0).}
\label{tab:moritzaugment}
\end{table}
Additionally, in order to study how quickly the network learns, we plot the mIoU on the training set against the number of training iterations. For that study we use the example of lake St.~Moritz (camera 0)  from winter 16-17. Results are shown in Fig.~\ref{fig:accuracy_vs_steps}. The (smoothed) learning curve is very steep initially (< 10k steps) but does not completely saturate, which indicates that more training data could probably improve the results further.
\normalsize
\par
\par
The generalisation performance (across cameras and winters) of the best webcam model reported in previous work \cite{LIP1_final_report_2019} is still unsatisfactory, especially for the cross-camera case. As can be seen from our cross-camera results (within St.~Moritz cameras, refer Table \ref{tab:webcams_quant:cross_camera}), the \textit{Deep-U-Lab} model trained using data from one camera works well on a different camera, meaning  that our method generalises well across cameras with totally different viewpoints, image scales and lighting conditions. Note that, we indeed improve over prior state-of-the-art \cite{LIP1_final_report_2019} significantly (gain of 35-40 percent points), which implies that \textit{Deep-U-Lab} has the capacity to learn generalisable class appearance, without overfitting to a specific camera geometry or viewpoint.
Our results for winter 17-18 are not as good as 16-17, primarily due to complicated lighting and ice patterns (e.g., black ice) which appeared only in that winter. In addition, the scores for the ice and clutter classes are low, primarily due to lower sample numbers. A comparison to prior work is not possible for winter 17-18, as that season has not been processed before. 
\begin{table}[ht!]
\small
\centering
\begin{tabular}{ccccccccc}\hline
   \multicolumn{2}{c}{\textbf{Training set}} & \multicolumn{2}{c}{\textbf{Test set}} & \multirow{2}{*}{\textbf{Water}} & \multirow{2}{*}{\textbf{Ice}} & \multirow{2}{*}{\textbf{Snow}} & \multirow{2}{*}{\textbf{Clutter}} & \multirow{2}{*}{\textbf{mIoU}}\\
   \textbf{Camera} & \textbf{Winter} & \textbf{Camera} & \textbf{Winter} & &  &  &  &  \\\hline
       &  &  &  &  &  &  &  &  \\
   Camera 0 & 16-17 & Camera 1 & 16-17 & 0.76 / \textcolor{gray}{0.36} & 0.75 / \textcolor{gray}{0.57} & 0.84 / \textcolor{gray}{0.37} & 0.61 / \textcolor{gray}{0.27} & 0.74 / \textcolor{gray}{0.39}  \\
   Camera 0 & 17-18 & Camera 1 & 17-18 & 0.62 & 0.66 & 0.89 & 0.42 & 0.64 \\
   Camera 1 & 16-17 & Camera 0 & 16-17 & 0.94 / \textcolor{gray}{0.32} & 0.75 / \textcolor{gray}{0.41} & 0.92 / \textcolor{gray}{0.33} & 0.48 / \textcolor{gray}{0.43} & 0.77 / \textcolor{gray}{0.37} \\
   Camera 1 & 17-18 & Camera 0 & 17-18 & 0.59 & 0.67 & 0.91 & 0.51 & 0.67 \\
       &  &  &  &  &  &  &  &  \\
   \hline
   \end{tabular}
   \caption{Results (IoU) of \textit{cross-camera} experiments. We compare our results with the \textit{Tiramisu Network}~\cite{LIP1_final_report_2019} (shown in \textcolor{gray}{grey}). Both cameras 0 and 1 monitor lake St.~Moritz.}
\label{tab:webcams_quant:cross_camera}
\end{table}
\normalsize
\par
\textit{Deep-U-Lab} performs superior to prior state-of-the-art in \textit{cross-winter} experiments, too (Table \ref{tab:webcams_quant:cross_winter}), outperforming~\cite{LIP1_final_report_2019} by about 14-20 percent points. However, it does not generalise across winters as well on lake Sihl. 
\begin{table}[h]
\small
\centering
\begin{tabular}{ccccccccc}\hline
   \multicolumn{2}{c}{\textbf{Training set}} & \multicolumn{2}{c}{\textbf{Test set}} & \multirow{2}{*}{\textbf{Water}} & \multirow{2}{*}{\textbf{Ice}} & \multirow{2}{*}{\textbf{Snow}} & \multirow{2}{*}{\textbf{Clutter}} & \multirow{2}{*}{\textbf{mIoU}}\\
   \textbf{Camera} & \textbf{Winter} & \textbf{Camera} & \textbf{Winter} & &  &  &  &  \\\hline
      &  &  &  &  &  &  &  &  \\
   Camera 0 & 16-17 & Camera 0 & 17-18 & 0.64 / \textcolor{gray}{0.45} & 0.58 / \textcolor{gray}{0.44} & 0.87 / \textcolor{gray}{0.83} & 0.59 / \textcolor{gray}{0.40} & 0.67 / \textcolor{gray}{0.53}  \\
   Camera 0 & 17-18 & Camera 0 & 16-17 & 0.98  &0.91 & 0.94 &0.58  &0.87 \\
   Camera 1 & 16-17 & Camera 1 & 17-18 & 0.86 / \textcolor{gray}{0.80} & 0.71 / \textcolor{gray}{0.58} & 0.93 / \textcolor{gray}{0.92} & 0.57 / \textcolor{gray}{0.33} & 0.77 / \textcolor{gray}{0.57}  \\
   Camera 1 & 17-18 & Camera 1 & 16-17 & 0.93 & 0.76 & 0.86 & 0.65 & 0.80  \\
   Camera 2 & 16-17 & Camera 2 & 17-18 & 0.61 & 0.14 & 0.35 & --- &0.51 \\
   Camera 2 & 17-18 & Camera 2 & 16-17 & 0.41 & 0.18 & 0.45 & --- & 0.50 \\
         &  &  &  &  &  &  &  &  \\
   \hline
   \end{tabular}
   \caption{Results (IoU) of \textit{cross-winter} experiments. We compare our results with \textit{Tiramisu Network}~\cite{LIP1_final_report_2019} (shown in \textcolor{gray}{grey}). Cameras 0 and 1 monitor lake St.~Moritz while camera 2 captures lake Sihl.}
\label{tab:webcams_quant:cross_winter}
\end{table}
\normalsize
\par
\begin{figure}[h!]
\centering
  \vspace{-2em}
  \subfloat[Tested on camera 0 (16-17), trained on camera 0 (17-18).]{\includegraphics[width=5.5cm]{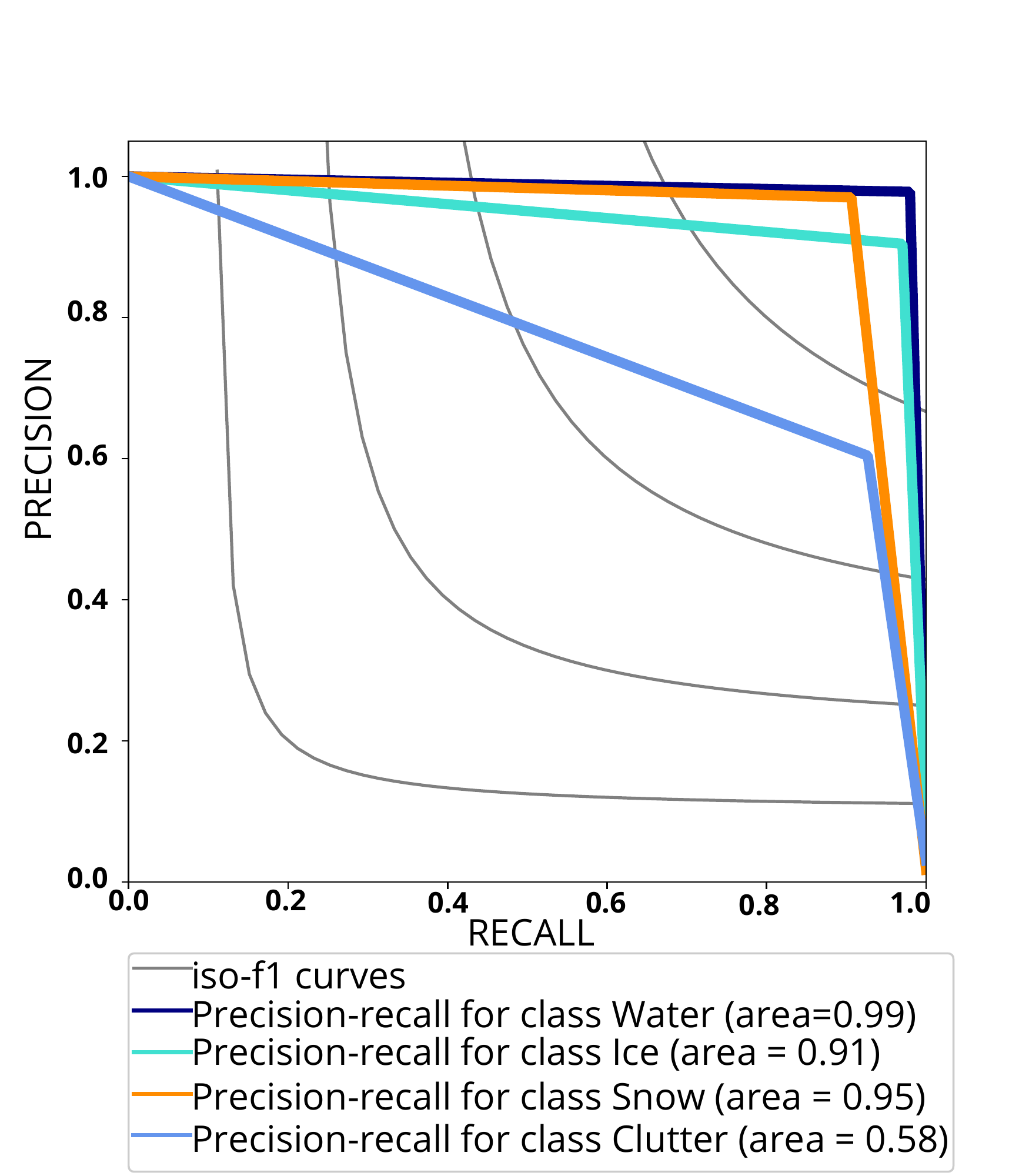}}
  \hspace{1em}
  \subfloat[Tested on camera 1 (16-17), trained on camera 1 (17-18).]{\includegraphics[width=5.5cm]{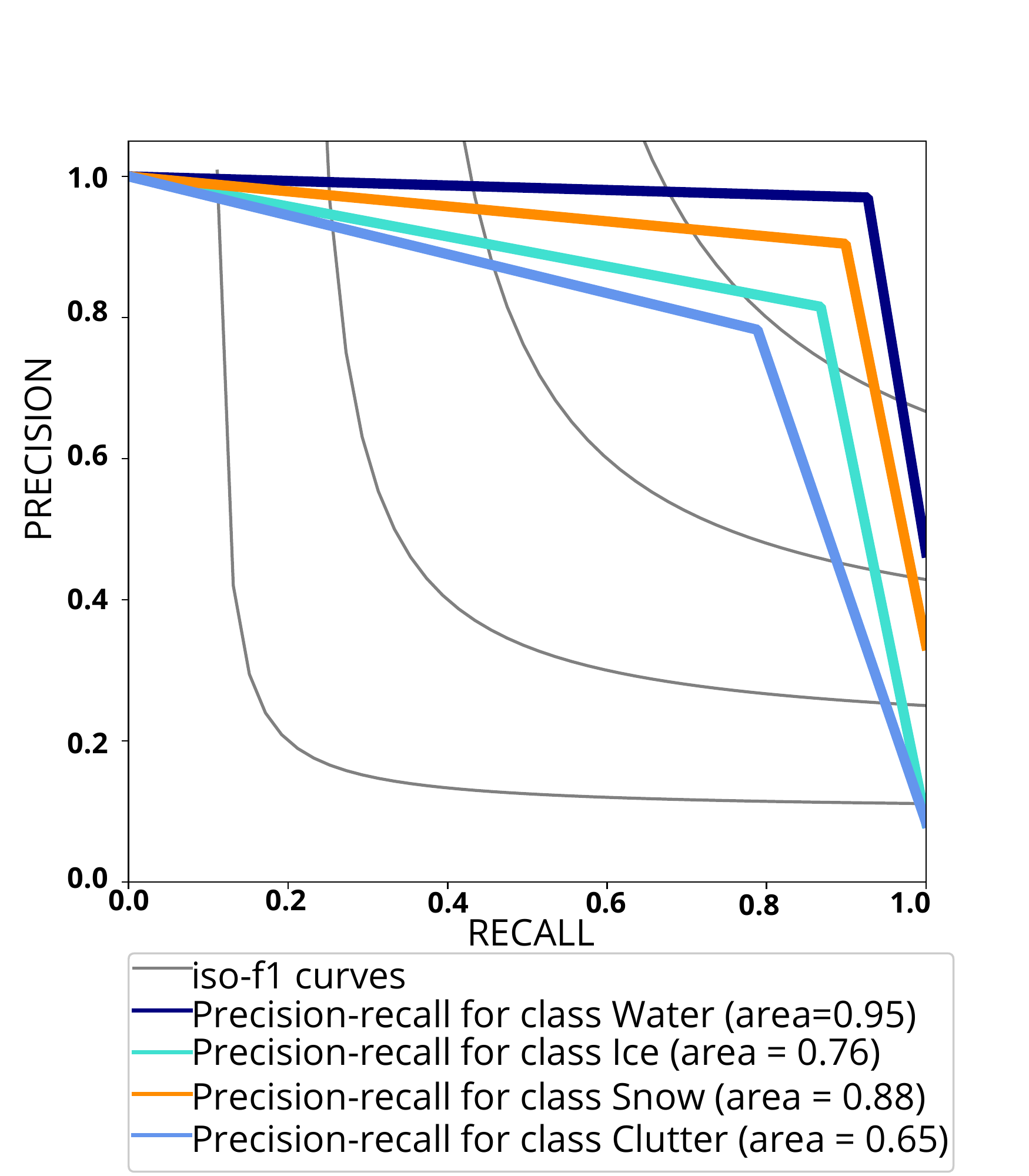}}
  \caption{Precision-recall plots (St.~Moritz) of \textit{cross-winter} experiments. Best if viewed on screen.} 
\label{fig:figure_placement}
\end{figure}

For a more complete picture of the cross-winter generalisation experiment, we also plot precision-recall curves (refer Fig.~\ref{fig:figure_placement}). Similar curves for same camera and cross-camera experiments can be found in Prabha et al. \cite{rajanie_tom_2020}. As expected,  segmentation of the two under-represented classes (\textit{clutter}, \textit{ice}) is less correct. Additionally, for the class \textit{clutter}, a considerable amount of the deviations from ground truth occur due to improper annotations rather than erroneous predictions. As drawing pixel-accurate ground truth boundaries around narrow man-made items placed on frozen lakes such as tents, poles etc. is time-consuming and tedious, the clutter objects are often annotated with rough summary masks that include a lot of snow/ice background. This greatly exaggerates the (relative) number of clutter pixels in the annotations, thus increasing the relative error.
According to Fig.~\ref{fig:figure_placement}, operating points around 85\% recall are a good trade-off for \textit{cross-winter} segmentation, if not every single pixel must be labelled.
%
%\par
A more extreme test is generalisation across lakes, with different spatial resolution, image quality, reflection and lighting patterns, shadows, etc. To our knowledge our work is the first one to try this. See Table \ref{tab:webcams_quant:cross_lake} for results. Before training the models, we remove the clutter pixels from camera 0, since camera 2 does not have any clutter that could serve as training data. 
Classifying images from a lake with different characteristics and acquired with a different type of camera proves challenging.
In one case the results are acceptable for the more frequent classes despite a noticeable drop, as for the case camera 2$\rightarrow$camera 0.
In the other case camera 0$\rightarrow$camera 2 the attempt largely fails.
The images of lake Sihl (camera 2) are of clearly lower quality and more difficult to classify, challenging even human annotators.
Consequently, training on St.~Moritz does not equip the classifier to deal with them.
\begin{table}[ht!]
\small
\centering
\begin{tabular}{ccccccc}\hline
      &  &  &   &  \\
   \textbf{Training set} & \textbf{Test set} & \textbf{Water} & \textbf{Ice} & \textbf{Snow} & \textbf{mIoU}\\
      &  &  &   &  \\
   Camera 0 (16-17) & Camera 2 (16-17) & 0.40 & 0.23 & 0.42 & 0.35 \\
   Camera 2 (16-17) & Camera 0 (16-17) & 0.85 & 0.25 & 0.68 & 0.60  \\
       &  &  &   &  \\
  \hline
   \end{tabular}
   \caption{Results (IoU) of \textit{cross-lake} experiments. Cameras 0 and 2 monitor lakes St.~Moritz and Sihl respectively.}
\label{tab:webcams_quant:cross_lake}
\end{table}
\normalsize
\par
In a further experiment, we divide the \textit{Photi-LakeIce} dataset into six folds, see Table \ref{tab:webcams_quant:LODO}. This makes it possible to perform experiments with a larger amount of training data, given that in previous experiments the loss had not fully saturated.
As expected from a high-capacity statistical model, more training data improves the results. I.e., it seems feasible to build a practical system if one is willing to undertake a bigger (but still reasonable and realistic) annotation effort.
An exception in this experiment is lake Sihl (camera 2), where the performance drops. This confirms the observation above that this camera is the most difficult one to segment in our dataset, and the domain gap from St.~Moritz to Sihl is too large to bridge without appropriate adaptation measures. One solution might be fine-tuning with at least a small set of cleverly picked samples from the target lake, but this is beyond the scope of the present paper.
\begin{table}[ht!]
\small
\centering
\begin{tabular}{ccccccc}\hline
      &  &  &  &    &  &  \\
   \textbf{Training set} & \textbf{Test set} & \textbf{Water} & \textbf{Ice} & \textbf{Snow} & \textbf{Clutter} & \textbf{mIoU}\\
      &  &  &  &    &  &  \\
   Camera 0 (17-18), Cameras 1 and 2 (2 winters) & Camera 0 (16-17) & 0.98  & 0.90 & 0.96 & 0.62 & 0.86 \\
   Camera 0 (16-17), Cameras 1 and 2 (2 winters) & Camera 0 (17-18) & 0.83 & 0.78 & 0.95 & 0.59 & 0.78  \\
   Camera 1 (17-18), Cameras 0 and 2 (2 winters) & Camera 1 (16-17) & 0.99 & 0.92 & 0.91 & 0.69 & 0.87 \\
   Camera 1 (16-17), Cameras 0 and 2 (2 winters) & Camera 1 (17-18) & 0.92 & 0.81 & 0.96 & 0.55 & 0.81 \\
   Camera 2 (16-17), Cameras 0 and 1 (2 winters) & Camera 2 (16-17) & 0.35 & 0.25 & 0.46 & --- & 0.35 \\
   Camera 2 (17-18), Cameras 0 and 1 (2 winters) & Camera 2 (17-18) & 0.66 & 0.30 & 0.36 & --- & 0.44 \\
         &  &  &  &  &  &  \\
   \hline
   \end{tabular}
   \caption{Results (IoU) of \textit{leave one dataset out} experiments. Cameras 0 and 1 monitor lake St.~Moritz while camera 2 captures lake Sihl.}
\label{tab:webcams_quant:LODO}
\end{table}
\normalsize
\par
\begin{figure}[h!]
  \centering
  \subfloat[Camera 0 (St.~Moritz)]{\fbox{\includegraphics[width=0.3\textwidth,height =3cm]{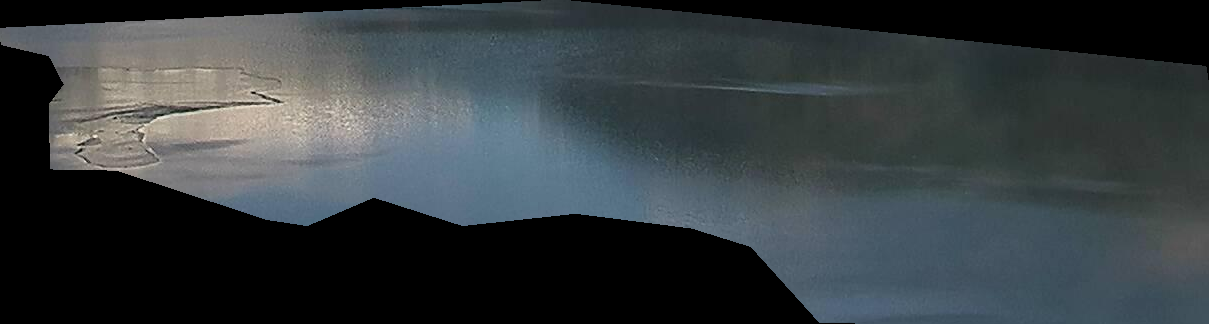}}}\hspace{0.01 em}
  \subfloat[Our prediction]{\fbox{\includegraphics[width=0.3\textwidth,height =3cm]{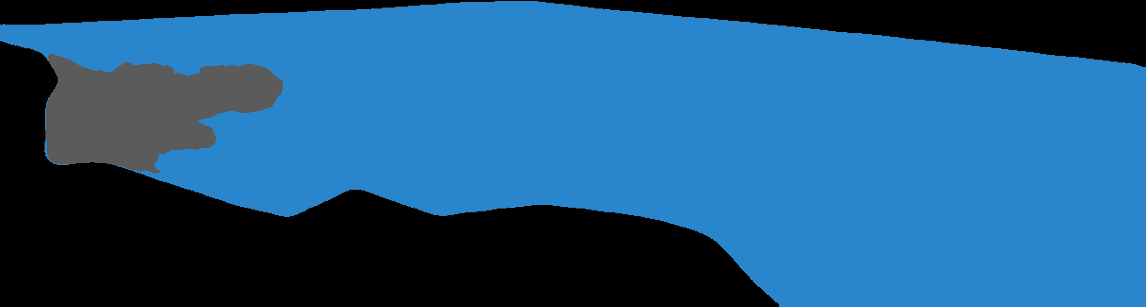}}}\hspace{0.01 em}
  \subfloat[Ground truth]{\fbox{\includegraphics[width=0.3\textwidth,height =3cm]{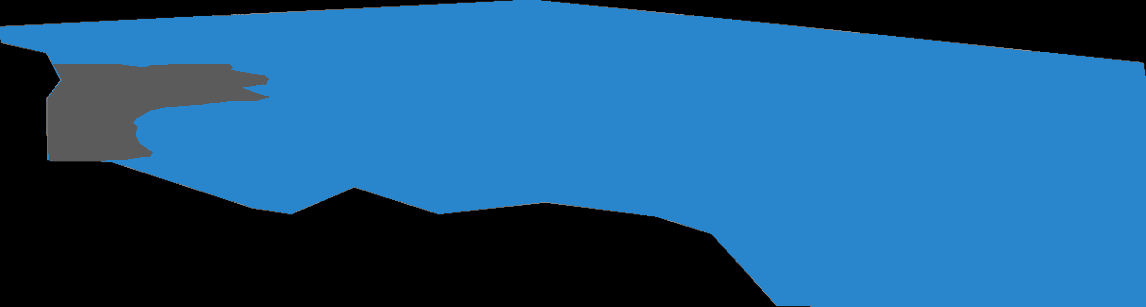}}}
  \vspace{-1.05em}
  \\
  \subfloat[Camera 1 (St.~Moritz)]{\fbox{\includegraphics[width=0.3\textwidth,height =3cm]{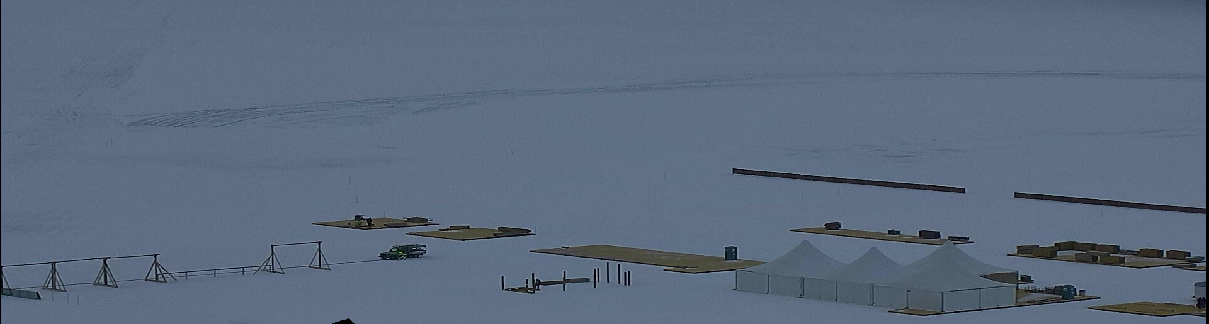}}}\hspace{0.01 em}
  \subfloat[Our prediction]{\fbox{\includegraphics[width=0.3\textwidth,height =3cm]{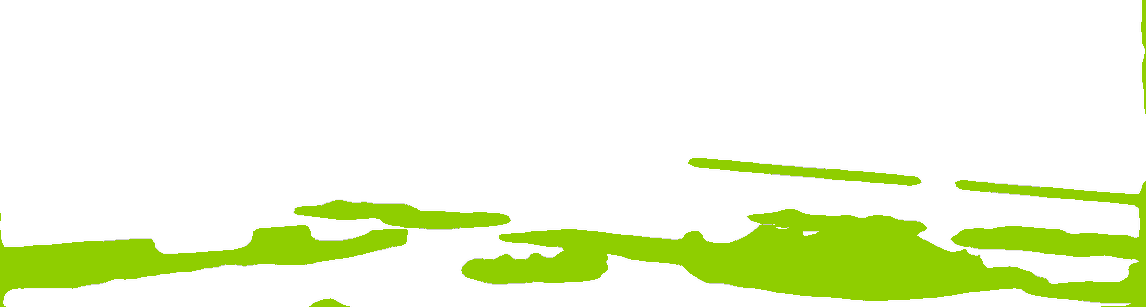}}}\hspace{0.01 em}
  \subfloat[Ground truth]{\fbox{\includegraphics[width=0.3\textwidth,height =3cm]{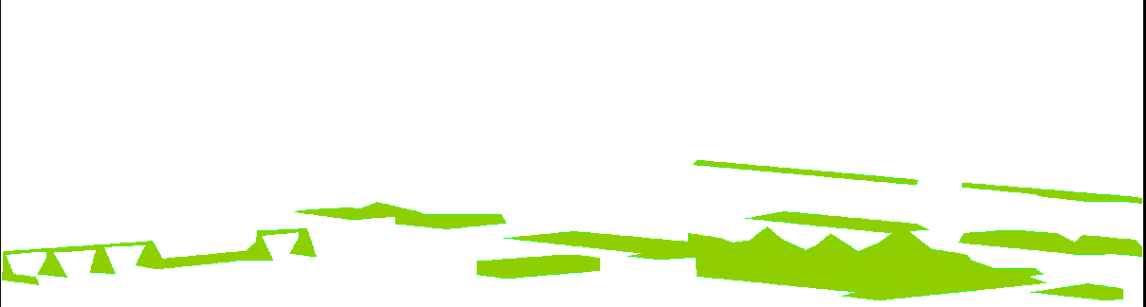}}}\\
  \subfloat[Colour code]{\includegraphics[width=0.35\columnwidth]{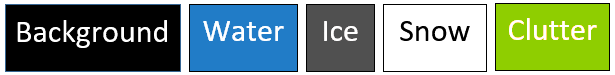}}
\caption{Qualitative lake ice segmentation results on webcam images. The colour code is also shown.} 
\label{fig:webcams_quali}
\end{figure}

To assess the lake ice segmentation visually, we depict qualitative webcam results for cameras 0 and 1 in Fig.~\ref{fig:webcams_quali}. \textit{Deep-U-Lab} successfully segments correctly in challenging scenarios. For instance, our network performed well even when shadows appeared on the lake either from clouds or nearby mountains (Fig.~\ref{fig:webcams_quali} row 1).
To determine how well the \textit{Deep-U-Lab} predictions follow the ground truth, especially during the freezing and thawing periods, we plot time series results which include all the transition as well as non-transition days from a full winter (17-18, see Fig.~\ref{fig:icecoverageon0}). 
Per image, we sum the numbers of ice and snow pixels and divide by the sum of all lake pixels. The resulting fractions of frozen pixels are aggregated into a daily value by taking the median.
Optionally, the daily values are further processed with another 3-day median to improve temporal coherence.
The daily fractions of frozen pixels ($y$-axis) are displayed in chronological order ($x$-axis), for the ground truth, daily prediction and smoothed prediction. Smoothing across time improves the final results by $\approx$3\%.
\subsection{Ice-on/off results}
\label{sec:iceonoff}
We go on to estimate the \textit{ice-on} and \textit{ice-off} dates using our satellite- and webcam-based approaches, results are shown in Table \ref{tab:iceonofftable}. A comparison with the ground truth dates (estimated by visual interpretation of webcams by a human operator) is also provided. Additionally, we compare our results with the in-situ temperature analysis results reported in Tom et al. \cite{LIP1_final_report_2019}. We can show the results only for one winter (16-17), since ground truth is not available for 17-18.
\par
Prior to the estimation of the two dates, we combine the time series results of both MODIS and VIIRS (Fig.~\ref{fig:sihl_MODIS_VIIRS_timeline}) in order to minimise  data gaps due to clouds, by simply filling in missing days in the VIIRS time series with MODIS results, whenever the latter are available. Even after merging the two time series, some gaps still exist during the critical transition periods. Note that the presence of gaps near the ice-on/off dates could affect the accuracy and confidence in the estimated dates. This is one of the risks when using optical satellite image analysis, and where webcams could constitute a valuable alternative, if sufficient coverage can be ensured for a lake of interest.
\begin{figure}[h!]
\centering
   \includegraphics[width=0.99\linewidth]{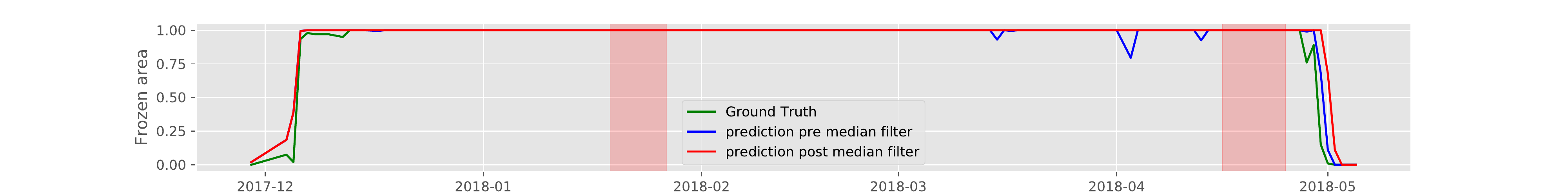}
   \caption{Cross-camera time-series results (winter 17-18) of lake St.~Moritz using \textit{Deep-U-Lab}. Results for camera 1 (when trained on camera 0 data) is displayed. All dates are shown in chronological order on the x-axis and the respective results (percentage of frozen pixels) are plotted on the y-axis. Data lost due to technical failures are shown as red bars.}
\label{fig:icecoverageon0}
\end{figure}
\begin{table}[h!]
  \centering
  \vspace{-0.5em}
    \begin{tabular}{ccccc}
        \toprule
        \textbf{Dates} & \textbf{Ground truth}& \textbf{Satellite approach} & \textbf{Webcam approach} & \textbf{In-situ (T)~\cite{LIP1_final_report_2019}} \\
        \midrule
   ice-on (Sihl) & 01.01.2017& 03.01.2017 & 04.01.2017 & 28-29.12.2016\\
   ice-off (Sihl) & 14.03.2017, 15.03.2017& 10.03.2017 & 14.02.2017 & 16.03.2017\\
   ice-on (Sils) & 02.01.2017, 05.01.2017& 06.01.2017 & - & 31.12.2016 \\
   ice-off (Sils) & 08.04.2017, 11.04.2017 & 31.03.2017 & - & 10.04.2017\\
   ice-on (Silvaplana) & 12.01.2017 & 15.01.2017 & - & 14.01.2017\\
   ice-off (Silvaplana) & 11.04.2017 & 30.03.2017 & - & 14.04.2017\\
   ice-on (Moritz) & 15-17.12.2016 & 01.01.2017 &  14.12.2016 & 17.12.2016\\
   ice-off (Moritz) & 30.03-06.04.2017 & 07.04.2017 & 18.03-26.04.2017 & 05-08.04.2017\\
\bottomrule
\end{tabular}
\caption{Ice-on/off dates (winter 16-17). Ground truth dates are shown in the order of confidence in case of more than one candidate.}
\label{tab:iceonofftable}
\end{table}
\par
We estimate the ice-on/off dates from the combined time series and show them as "satellite approach" in Table \ref{tab:iceonofftable}. The best results are obtained with an RBF kernel for St.~Moritz and with a linear kernel for rest of the lakes. In most cases the ice-on/off dates have an offset of 1-4 days from the ground truth. Exceptions are the ice-off dates of Sils and Silvaplana. Note that data from only one winter (of the lake being tested) is present in the corresponding training set. It appears that training data from more winters is critical to estimate accurate ice-on/off dates. We note that there could also be noise in the ground truth ice-on/off dates due to human interpretation errors. Using the satellite approach, significant errors in ice-on/off estimation exist also for St.~Moritz. Recall that the daily decision for lake St.~Moritz is taken based on just four pixels, and based only on cloud-free days in MODIS. This clearly points to the fact that MODIS (and even more, VIIRS) imagery is not the best choice for very small lakes. The results obtained with webcams show higher accuracy for lake St.~Moritz (see Table \ref{tab:iceonofftable} and Fig.~\ref{fig:icecoverageon0}). Here, we use camera 0 to estimate the two phenological dates, since it has a better coverage of the lake than camera 1. Note that no data is available between 18.03.2017 and 26.04.2017 due to a technical problem with the camera, and \textit{ice-off} unfortunately occurred during that period. While we obtain excellent results for St.~Moritz with webcams, the accuracy of ice-on/off for lake Sihl is not good, primarily because of the limited image quality with low spatial resolution (see Table \ref{tab:datasetlakeice}), compression artefacts, and acute view angles (see Fig.~\ref{fig:Photi-lakeice-dataset-snapshots}).
\section{Discussion}
\subsection{VIIRS and MODIS analysis}
The optical satellite sensors such as MODIS and VIIRS can clearly serve as a basis for routine monitoring of lake ice (especially for global coverage) and the results achieved show a high level of accuracy. One weakness is their inability to penetrate clouds, especially during lake freeze-up and break-up. The main advantage of MODIS is the availability of longer time series data. In addition, MODIS has useful bands in various areas of the electromagnetic spectrum. However, there are several disadvantages too. The radiometric quality is not that good, moreover, the sensor is very old and its absolute geolocation is less accurate than that of VIIRS (more important for small lakes). Furthermore, MODIS data is expected to eventually be discontinued, whereas VIIRS operation is guaranteed over a longer future period (JPSS-1/NOAA-20 until 2024; JPSS-2 with same suite of sensors will be launched in 2021 with designed life time of 7 years; JPSS-3 and -4 are in the planning phase). %
\par
Poor spatial resolution (particularly for VIIRS), makes it impossible to operate our satellite methodology on small-size lakes up to at least 2$\,$km$^\text{2}$. Another issue is the significant confusion between (thin) \emph{ice} and \emph{water} since the similar reflectance of these two classes can confuse the classifier. Unfortunately there exist very few non-transition dates with no clouds, snow-free ice, mixture of thin ice and water in both the MODIS and VIIRS datasets, such that training a reliable model for these situations still remains a challenge. Moreover, the presence of label noise in the ground truth impacts the training. Such noise occurs mainly because most of the webcams are not optimally configured and it is very difficult to capture the whole lake in a single webcam frame. This problem is even bigger for larger lakes. Integrating the visual interpretations from multiple cameras observing a lake is cumbersome as well as challenging. One possible solution for the noise problem could be to not train on dates near the transition period, for which label noise in the ground truth is more likely. It is, however, equally possible that this would even aggravate the problem, as the conditions seen in the training set would become even less representative of the transition periods. Large-scale in-situ measurements are an alternative to webcam-based ground truth, but are not realistic for wide-area coverage. Also, imperfections of the cloud-masks bring in more errors in the high-accuracy range where our method operates. 
\par
Regarding our SVM-based methodology, the RBF kernel tends to not generalise as well as its linear counterpart. However, this behaviour may depend on the available training data. Under our current experimental conditions, the linear kernel overall has the upper hand, but this assessment could still change when using data from multiple winters.
\subsection{Webcam analysis}
For webcam data featuring sufficient spatial resolution, we see a great potential for lake ice monitoring. 
We do note that webcam placement is restricted by practical considerations. Selecting and/or mounting webcams for lake ice monitoring will normally be a compromise between the ideal geometric configuration and finding a place where the device can be installed with reasonable effort. For the ideal placement, the usual perspective imaging rules apply, most importantly viewing directions from above are preferable over grazing angles, and viewing directions directly towards the sun should be avoided as much as possible.
\par
One question that still remains unanswered is: what is the reason that \textit{Deep-U-Lab} outperforms the \textit{Tiramisu} lake ice network~\cite{Jegou2016_CVPRW,muyan_lakeice_2018,LIP1_final_report_2019}? One possible explanation is that our model profits from the smarter dilated convolutions and multi-scale pyramid pooling at the feature extraction stage, effectively letting the network grasp a relatively broader context as opposed to the \textit{Tiramisu} network. Additionally, our model heavily benefits from the pre-trained weights to learn with still limited training data for the lake ice task. Our \textit{Deep-U-Lab} model did not converge when we tried to train it without transfer learning; whereas pre-trained weights for FC-DenseNet are not available, so that we can at the moment not quantify the influence of pre-training.
\par
Regarding the computational efficiency of the CNN approach, (off-line) training for 100k steps on camera 0 (820 images) takes $\approx$24 hours on a PC equipped with a NVIDIA GeForce GTX 1080 Ti graphics card (for cross-camera experiment, lake St.~Moritz, winter 16-17). Testing takes $\approx$10 minutes for the 1180 images of camera 1.
\section{Conclusions}\label{sec:conclusion}
We have investigated the potential of machine-learning based image analysis, in combination with various image sensors to retrieve lake ice. So far that approach has been rarely explored, especially with regard to the many small lakes on Earth (particularly in mountain regions), but can be a valuable source of information that is largely independent of in-situ observations as well as models of the freezing/thawing process.
We put forward an easy-to-use, SVM-based approach to detect lake ice in MODIS and VIIRS satellite images and show that it delivers conclusive results. Additionally, we have set a new state-of-the-art from webcam-based lake ice monitoring, using the \textit{Deep-U-Lab} network, and have in that context also automated the detection of lake outlines, as a further step towards operational monitoring with webcams. Finally, we have introduced a new, public webcam dataset with pixel-accurate annotations.
\par
To detect lake ice from MODIS and VIIRS optical satellite imagery, we proposed a simple, generic machine learning-based approach which achieves high accuracy for all tested lakes. Though we focused on Swiss Alpine lakes, the proposed approach is very straight-forward and hence the results could hopefully be directly applied to other lakes with similar conditions, in Switzerland and abroad, and possibly to other sensors with similar characteristics. We have demonstrated that our approach generalises well across winters and lakes (with similar geographical and meteorological conditions). In addition to the lake ice detection from space, we have proposed to use free data from terrestrial webcams for lake ice monitoring. Webcams are especially suitable	for	small lakes (ca. up	to 2$\,$km$^\text{2}$), which cannot be monitored by VIIRS-type sensors. Despite the limited image quality, we obtained promising results using deep learning. Webcams have good ice detection capability	with much higher spatial resolution compared to satellites. However, one has no control over the location, orientation, lake area coverage, and image quality (often poor) of public webcams. Also, there are no, or too few, webcams for some lakes. On the positive side, webcams are largely unaffected by cloud cover. For continental/global coverage, satellite-based monitoring is clearly the method of choice, again confirming the advantage of satellite observations for large-scale Earth observation. For focused, local campaigns and as a source of reference data at selected sites, webcam-based monitoring method may be an interesting alternative. Note also, it may in certain cases be warranted to install dedicated webcams (respectively, surveillance cameras) with pan, tilt and zoom functionalities optimised for lake ice monitoring.
\par
One way to circumvent the problem of clouds with optical satellite sensors is to use microwave data. In particular, Sentinel-1 SAR data (GSD 10$\,$m, freely available) looks very promising~\cite{tom_aguilar_2020}. Optical sensors such as Sentinel-2 and Landsat-8 are visually easier to interpret w.r.t.\ lake ice than Radar, and have better spatial resolution than MODIS and VIIRS. Although they are not suitable as stand-alone source for lake ice monitoring due to their low temporal resolution (under ideal conditions 5 days), still they may in certain cases be useful to fill gaps in VIIRS / MODIS results.
\par
We consider our satellite-based approach as a first step, and ultimately hope to produce a 20-year time series, using MODIS data since 2000. It will be interesting to correlate the longer-term lake freezing trends with other climate time series such as surface temperature or CO$_\text{2}$ levels. We expect that such a time series will be helpful to draw conclusions about the local and global climate change.
\par
One technical finding of our study is that the prior learning-based approaches ~\citep{muyan_lakeice_2018,LIP1_final_report_2019} did not fully leverage the power of deep CNNs to observe lake ice. At the methodological level, we have clearly demonstrated the potential of machine (deep) learning systems for lake ice monitoring, and hope that this research direction will be pursued further. Given the good cross-winter and cross-camera generalisation of the models and computational efficiency at inference time (on GPU for the CNN part), an operational deployment is within reach. Our results show that employing the state-of-the-art CNN frameworks is highly effective for ice analysis, especially during the transition periods. What still needs improvement is cross-lake generalisation. We do expect that a \textit{Deep-U-Lab} model trained using data from a couple of winters could consistently reach >80\% IoU on the four major classes. From the segmentation results we were in many cases able to determine the \textit{ice-on} and \textit{ice-off} dates to within 1-2 days -- for that task the relatively better quality webcams were particularly helpful, as satellite-based segmentation was less reliable during the transition periods. An interesting direction may be to reduce the one-time effort for ground truth labelling with techniques such as domain adaptation or active learning.
\par
For monitoring small lakes, integrating the webcam results with in-situ temperature measurements looks like a possible future direction. Additionally, for such lakes, usage of UAVs equipped with both thermal and RGB cameras could be a plausible option, but may be difficult to operationalise due to the need for accurate georeferencing, lack of robustness in cold weather, as well as legal flight restrictions. An intriguing extension of the webcam-based approach could be to use crowd-sourced imagery for lake ice detection. We published some preliminary results in one of our recent works \cite{rajanie_tom_2020}. A large, and exponentially growing, number of images are available on the internet and social media. With the advance of smartphones equipped with cameras and the habit of selfies, many personal images show a lake in background. Still regular coverage of a given site is hard to ensure, and accurate geo-referencing of such images is challenging.
%%%%%%%%%%%%%%%%%%%%%%%%%%%%%%%%%%%%%%%%%%

%%%%%%%%%%%%%%%%%%%%%%%%%%%%%%%%%%%%%%%%%%

%%%%%%%%%%%%%%%%%%%%%%%%%%%%%%%%%%%%%%%%%%

%%%%%%%%%%%%%%%%%%%%%%%%%%%%%%%%%%%%%%%%%%
%\section{Patents}
%This section is not mandatory, but may be added if there are patents resulting from the work reported in this manuscript.

%%%%%%%%%%%%%%%%%%%%%%%%%%%%%%%%%%%%%%%%%%
\vspace{6pt} 

%%%%%%%%%%%%%%%%%%%%%%%%%%%%%%%%%%%%%%%%%%
%% optional
%\supplementary{The following are available online at \linksupplementary{s1}, Figure S1: title, Table S1: title, Video S1: title.}


%%%%%%%%%%%%%%%%%%%%%%%%%%%%%%%%%%%%%%%%%%
%\authorcontributions{Conceptualization, M.T., R.P., E.B., L.L., K.S.; Data curation, M.T., R.P and T.W.; Methodology, M.T. and R.P.; Software, M.T. and R.P.; Supervision, M.T., E.B., L.L. and K.S.; Validation, M.T., R.P and T.W.; Writing--original draft, M.T.; Writing--review and editing, E.B., K.S.; All authors have read and agreed to the submitted version of the manuscript.}

%%%%%%%%%%%%%%%%%%%%%%%%%%%%%%%%%%%%%%%%%%

%%%%%%%%%%%%%%%%%%%%%%%%%%%%%%%%%%%%%%%%%%
\acknowledgments{This work was funded mainly by the Swiss Federal Office of Meteorology and Climatology (MeteoSwiss) and partially by the Sofja Kovalevskaja Award of the Humboldt Foundation. We express our gratitude to all the partners in our MeteoSwiss projects for their support. Additionally, we acknowledge Mathias Rothermel, Muyan Xiao, and Konstantinos Fokeas for their contributions in collecting and annotating the images of Photi-LakeIce dataset. We also thank Hotel Schweizerhof for providing the webcam data of lake St.~Moritz.}

%%%%%%%%%%%%%%%%%%%%%%%%%%%%%%%%%%%%%%%%%%
%\conflictsofinterest{The authors declare no conflict of interest.} 

%%%%%%%%%%%%%%%%%%%%%%%%%%%%%%%%%%%%%%%%%%
%% optional
%\abbreviations{The following abbreviations are used in this manuscript:\\

%\noindent 
%\begin{tabular}{@{}ll}
%MODIS & Moderate Resolution Imaging Spectroradiometer\\
%VIIRS & Visible Infrared Imaging Radiometer Suite\\
%SVM & Support Vector Machine\\
%RBF & Radial Basis Function\\
%CNN & Convolutional Neural Network\\
%GSD & Ground Sampling Distance\\
%LSWT & Lake Surface Water Temperature
%\end{tabular}}

%%%%%%%%%%%%%%%%%%%%%%%%%%%%%%%%%%%%%%%%%%
%% optional
\appendixtitles{no} % Leave argument "no" if all appendix headings stay EMPTY (then no dot is printed after "Appendix A"). If the appendix sections contain a heading then change the argument to "yes".
%\appendix
%\section{}
%\unskip
%\subsection{}
%The appendix is an optional section that can contain details and data supplemental to the main text. For example, explanations of experimental details that would disrupt the flow of the main text, but nonetheless remain crucial to understanding and reproducing the research shown; figures of replicates for experiments of which representative data is shown in the main text can be added here if brief, or as Supplementary data. Mathematical proofs of results not central to the paper can be added as an appendix.

%\section{}
%All appendix sections must be cited in the main text. In the appendixes, Figures, Tables, etc. should be labeled starting with `A', e.g., Figure A1, Figure A2, etc. 

%%%%%%%%%%%%%%%%%%%%%%%%%%%%%%%%%%%%%%%%%%
\reftitle{References}



\externalbibliography{yes}
%\bibliography{references}

%=====================================
% References, variant B: internal bibliography
%=====================================
%\begin{thebibliography}{999}
% Reference 1
%\bibitem[Author1(year)]{ref-journal}
%Author1, T. The title of the cited article. {\em Journal Abbreviation} {\bf 2008}, {\em 10}, 142--149.
% Reference 2
%\bibitem[Author2(year)]{ref-book}
%Author2, L. The title of the cited contribution. In {\em The Book Title}; Editor1, F., Editor2, A., Eds.; Publishing House: City, Country, 2007; pp. 32--58.
%\end{thebibliography}

% The following MDPI journals use author-date citation: Arts, Econometrics, Economies, Genealogy, Humanities, IJFS, JRFM, Laws, Religions, Risks, Social Sciences. For those journals, please follow the formatting guidelines on http://www.mdpi.com/authors/references
% To cite two works by the same author: \citeauthor{ref-journal-1a} (\citeyear{ref-journal-1a}, \citeyear{ref-journal-1b}). This produces: Whittaker (1967, 1975)
% To cite two works by the same author with specific pages: \citeauthor{ref-journal-3a} (\citeyear{ref-journal-3a}, p. 328; \citeyear{ref-journal-3b}, p.475). This produces: Wong (1999, p. 328; 2000, p. 475)

%%%%%%%%%%%%%%%%%%%%%%%%%%%%%%%%%%%%%%%%%%
%% optional
%\sampleavailability{Samples of the compounds ...... are available from the authors.}

%% for journal Sci
%\reviewreports{\\
%Reviewer 1 comments and authors’ response\\
%Reviewer 2 comments and authors’ response\\
%Reviewer 3 comments and authors’ response
%}

%%%%%%%%%%%%%%%%%%%%%%%%%%%%%%%%%%%%%%%%%%
\end{document}